%% file: LLM_sourcing.tex
\documentclass[acmsmall,screen]{acmart}

\usepackage{mydef}
\usepackage{lineno}
\usepackage{multirow}
\usepackage{amsmath,amsfonts}
\usepackage{algorithmic}
\usepackage{algorithm}
\usepackage{array}
\usepackage[caption=false,font=normalsize,labelfont=sf,textfont=sf]{subfig}
\usepackage{textcomp}
\usepackage{stfloats}

\usepackage{url}
\usepackage{verbatim}
\usepackage{graphicx}

\usepackage{amsmath}      % Optional, for math support
\usepackage{hyperref}     % Optional, for hyperlinks
\usepackage{balance}
\usepackage{enumitem}

\def\BibTeX{{\rm B\kern-.05em{\sc i\kern-.025em b}\kern-.08em
    T\kern-.1667em\lower.7ex\hbox{E}\kern-.125emX}}

\usepackage{tabularray}
\usepackage{booktabs}       % professional-quality tables

\usepackage{tikz}
\usetikzlibrary{shapes,arrows.meta,positioning,fit,matrix}

\usepackage[edges]{forest}
\usepackage{xcolor}

\usepackage{svg}
\usepackage{float}

\usepackage{adjustbox} % 可选，用于微调
\usepackage{makecell}
\AtBeginDocument{%
  \providecommand\BibTeX{{%
    Bib\TeX}}}

\setcopyright{acmlicensed}
\copyrightyear{2025}
\acmYear{2025}
\acmDOI{XXXXXXX.XXXXXXX}

\acmJournal{JACM}
\acmVolume{37}
\acmNumber{4}
\acmArticle{111}
\acmMonth{8}

\begin{document}

\title{Large Language Model Sourcing: A Survey}
\author{Liang~Pang}
\email{pangliang@ict.ac.cn}
\affiliation{%
  \institution{Institute of Computing Technology, Chinese Academy of Sciences and University of Chinese Academy of Sciences}
  \city{Beijing}
  \country{China}
}

\author{Jia~Gu}
\authornote{These authors contributed equally to this research.}
\affiliation{%
  \institution{Institute of Computing Technology, Chinese Academy of Sciences and University of Chinese Academy of Sciences}
  \city{Beijing}
  \country{China}
  }
  
\author{Sunhao~Dai}
\authornotemark[1]
\affiliation{%
  \institution{Gaoling School of Artificial Intelligence, Renmin University of China}
  \city{Beijing}
  \country{China}
}

\author{Zihao~Wei}
\authornotemark[1]
\author{Zenghao~Duan}
\authornotemark[1]
\author{Kangxi~Wu}
\authornotemark[1]
\author{Zhiyi~Yin}
\affiliation{%
  \institution{Institute of Computing Technology, Chinese Academy of Sciences and University of Chinese Academy of Sciences}
  \city{Beijing}
  \country{China}
  }

\author{Jun~Xu}
\affiliation{%
  \institution{Gaoling School of Artificial Intelligence, Renmin University of China}
  \city{Beijing}
  \country{China}
}

\author{Huawei~Shen}
\author{Xueqi~Cheng}
\affiliation{%
  \institution{Institute of Computing Technology, Chinese Academy of Sciences and University of Chinese Academy of Sciences}
  \city{Beijing}
  \country{China}
}

%%
%% By default, the full list of authors will be used in the page
%% headers. Often, this list is too long, and will overlap
%% other information printed in the page headers. This command allows
%% the author to define a more concise list
%% of authors' names for this purpose.
\renewcommand{\shortauthors}{Pang et al.}

\input{sections/0_abstract}

%%
%% The code below is generated by the tool at http://dl.acm.org/ccs.cfm.
%% Please copy and paste the code instead of the example below.
%%

%todo
\begin{CCSXML}
<ccs2012>
   <concept>
       <concept_id>10003752.10010070.10010111.10003623</concept_id>
       <concept_desc>Theory of computation~Data provenance</concept_desc>
       <concept_significance>500</concept_significance>
       </concept>
   <concept>
       <concept_id>10002951</concept_id>
       <concept_desc>Information systems</concept_desc>
       <concept_significance>300</concept_significance>
       </concept>
   <concept>
       <concept_id>10002978.10003029</concept_id>
       <concept_desc>Security and privacy~Human and societal aspects of security and privacy</concept_desc>
       <concept_significance>500</concept_significance>
       </concept>
   <concept>
       <concept_id>10002978.10003018</concept_id>
       <concept_desc>Security and privacy~Database and storage security</concept_desc>
       <concept_significance>500</concept_significance>
       </concept>
   <concept>
       <concept_id>10010147.10010178.10010179</concept_id>
       <concept_desc>Computing methodologies~Natural language processing</concept_desc>
       <concept_significance>500</concept_significance>
       </concept>
 </ccs2012>
\end{CCSXML}

\ccsdesc[500]{Theory of computation~Data provenance}
\ccsdesc[300]{Information systems}
\ccsdesc[500]{Security and privacy~Human and societal aspects of security and privacy}
\ccsdesc[500]{Security and privacy~Database and storage security}
\ccsdesc[500]{Computing methodologies~Natural language processing}

\keywords{AI Safety, Large Language Models, Wartermarking, AIGC Detection, Influence Function, Retrieval-Augmented Generation, Model Sourcing, Structure Sourcing, Training Data Sourcing, External Data Sourcing.}

\received{20 February 2025}
\received[revised]{12 March 2009}
\received[accepted]{5 June 2009}

%%
%% This command processes the author and affiliation and title
%% information and builds the first part of the formatted document.
\maketitle

\input{sections/1_introduction}

\input{sections/_tree}

\input{sections/2_background}

% gujia & lixiang
\input{sections/3_model_sourcing}

% weizihao & duanzenghao
\input{sections/4_structure_sourcing}

% wukangxi
\input{sections/5_data_sourcing}

% daisunhao
\input{sections/6_citation_sourcing}

\input{sections/7_challenges}

\input{sections/8_conclusion}

\newpage

\bibliographystyle{ACM-Reference-Format}
\bibliography{references}

\end{document}

%% file: sections/0_abstract.tex
\begin{abstract}

Due to the black-box nature of large language models (LLMs) and the realism of their generated content, issues such as hallucinations, bias, unfairness, and copyright infringement have become significant. In this context, sourcing information from multiple perspectives is essential. This survey presents a systematic investigation organized around four interrelated dimensions: Model Sourcing, Model Structure Sourcing, Training Data Sourcing, and External Data Sourcing. Moreover, a unified dual-paradigm taxonomy is proposed that classifies existing sourcing methods into prior-based (proactive traceability embedding) and posterior-based (retrospective inference) approaches. Traceability across these dimensions enhances the transparency, accountability, and trustworthiness of LLMs deployment in real-world applications.

\end{abstract}

%% file: sections/1_introduction.tex
\section{Introduction}

The launch of ChatGPT\footnote{\url{https://openai.com/blog/chatgpt}} marked a watershed moment for large language models (LLMs), catalyzing rapid advancements that have propelled natural language processing into a transformative new era. Models such as DeepSeek~\cite{guo2025deepseek}, GPT-4~\cite{openai2023gpt4}, and LLaMA3~\cite{grattafiori2024llama} exemplify this explosive growth. Collectively, they signal a paradigm shift in artificial intelligence -- from supporting objective tasks such as classification and prediction to enabling more subjective, decision-oriented reasoning~\cite{zhao2023survey}. These models now exhibit impressive versatility across a range of domains, including programming~\cite{roziere2023code}, healthcare~\cite{xu2025reversephysicianairelationshipfullprocess}, and legal services~\cite{lai2023large}. As LLMs increasingly influence high-stakes decisions and enable naturalistic human-AI interactions, they bring us closer to the frontier of general intelligence and emergent reasoning capabilities~\cite{wei2022emergent}.

\begin{figure}
\centering
\includegraphics[width=0.7\textwidth,trim=0 0 0 0,]{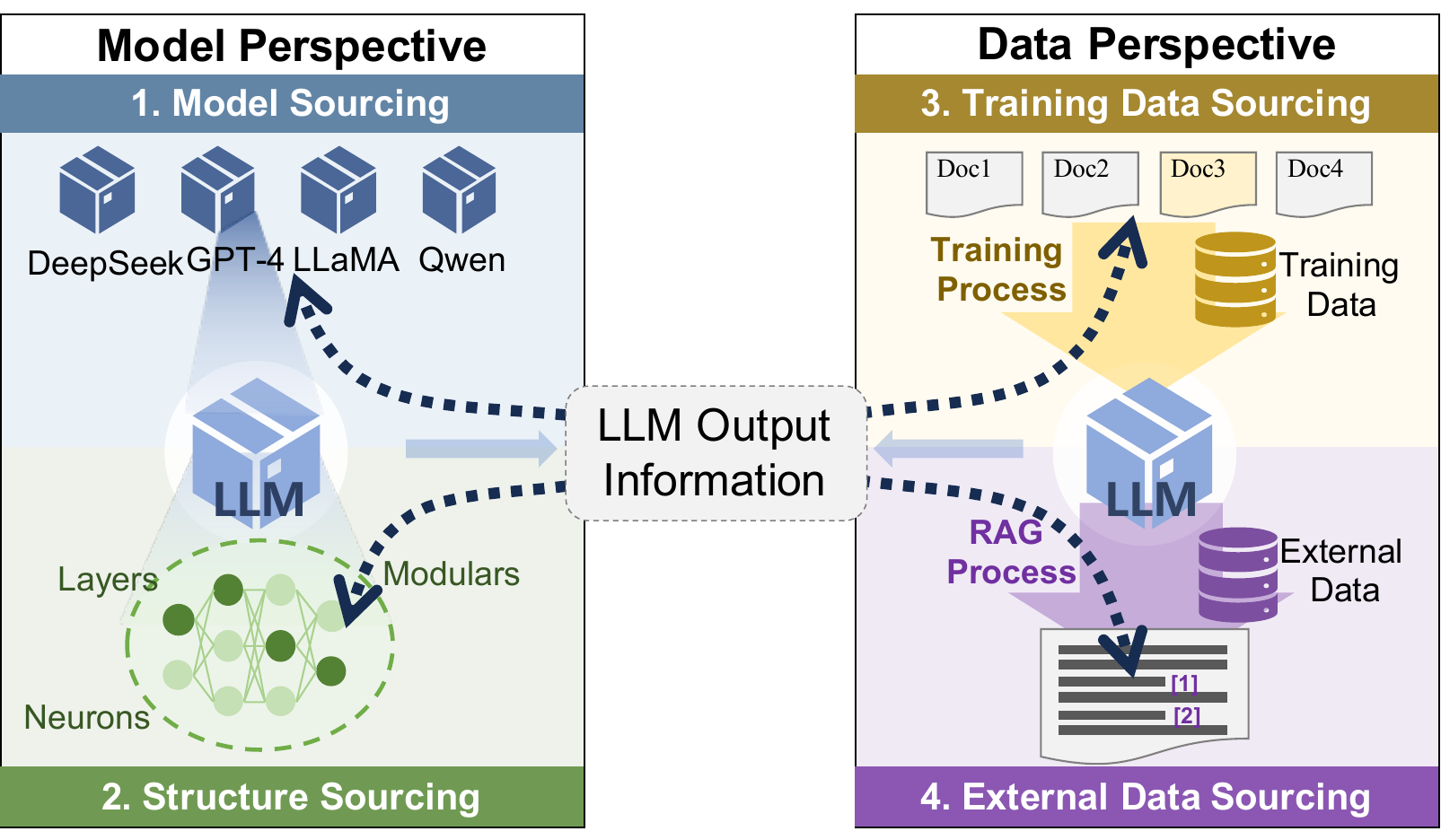}
\caption{Four sourcing dimensions in LLMs: from the model side, outputs derive from the specific model (model sourcing) or its internal architecture and mechanisms (structure sourcing); from the data side, outputs trace to training samples (training data sourcing) or to external corpora (external data sourcing).}

\label{fig:framework}
\end{figure}
However, this broad applicability also amplifies risks rooted in model architectures and consumed data. In practice, inherent biases in training and retrieved data~\cite{dai2024bias} (e.g., underrepresentation of marginalized groups) propagate inequities by reinforcing stereotypes and producing discriminatory outputs~\cite{gallegos2024bias}. Hallucinations and factual inaccuracies, especially in medicine and law~\cite{huang2025survey, ding2024retrieve}, carry real-world consequences. LLMs opacity hinders accountability, while security vulnerabilities (e.g., prompt injection, data poisoning) enable malicious misuse~\cite{li2025security}, including harmful content generation or sensitive-information leakage. Their capacity for manipulative and coercive content further raises ethical concerns around autonomy, consent, and trust~\cite{methuku2025digital}. As LLMs enter critical applications, confronting these multifaceted risks is essential.

Among these challenges, two stand out as particularly foundational: the indistinguishability of generated content and the structural opacity of the models themselves. Outputs derived from latent linguistic patterns learned during training increasingly blur boundaries with human-authored text and can mask harmful or non-compliant content~\cite{ parshakov2025usersfavorllmgeneratedcontent, jones2024liesdamnedliesdistributional}. At the same time, LLMs' black-box nature makes error diagnosis and causal attribution prohibitively difficult, while their scale renders traditional debugging and mitigation approaches infeasible~\cite{ singh2024rethinking, zhao2023explainability}. These twin issues highlight the urgent need for robust output sourcing mechanisms.

Sourcing enables systematic risk mitigation by establishing verifiable accountability across the LLM lifecycle. \textbf{From the model perspective}, provenance sourcing pinpoints harmful or non-compliant outputs to specific models or architectural components, enabling targeted refinement and oversight. \textbf{From the data perspective}, sourcing links outputs to training samples or retrieved sources, supporting attribution, fair compensation, bias and noise diagnosis, and hallucination reduction~\cite{wu2024enhancing, abdelrahman2023knowledge}. By bridging internal mechanisms (model structure and parameters) and external dependencies (training corpora and retrieval), sourcing shifts risk management from reactive to proactive and auditable, addressing structural opacity and lack of accountability.

To support this vision, our survey proposes a unified sourcing framework integrating model and data perspectives, categorizing methods into prior-based and posterior-based paradigms. Prior-based approaches embed traceable markers during training or data preparation for explicit attribution, while posterior-based methods infer provenance from outputs, activations, or gradients without architectural changes. This framework grounds attribution across the full LLM lifecycle. As shown in Fig.~\ref{fig:framework}, our survey spans four dimensions: \textbf{Model}, \textbf{Model Structure}, \textbf{Training Data}, and \textbf{External Data}, offering a comprehensive lens on provenance, responsibility, and traceability in modern LLM ecosystems.

$\bullet$ \textbf{Model Sourcing} attributes generated content to specific LLMs or human authors, a crucial task as AI-generated text becomes increasingly indistinguishable from human writing.
This dimension plays a key role in curbing misinformation by tracing the origins of synthetic content.

$\bullet$ \textbf{Model Structure Sourcing} examines how architectural components (e.g., parameter configurations, attention mechanisms, and activation pathways) influence model behavior, thereby shedding light on otherwise opaque decision-making processes. 

$\bullet$ \textbf{Training Data Sourcing} enables causal attribution of generated outputs to specific training samples, addressing risks related to biased, sensitive, or noisy data. This dimension formalizes the relationship between data provenance and model behavior, advancing solutions to key challenges in fairness, privacy, and compliance.

$\bullet$ \textbf{External Data Sourcing} attributes output content to user inputs or retrieved external knowledge, helping to detect adversarial prompts, manipulation, or contextual ambiguities. 

These four dimensions span the ``supply side'' (model, structure) and ``demand side'' (training data, external data) of LLM-generated content, ensuring comprehensive provenance analysis.
This survey proposes a unified full-lifecycle provenance framework for LLMs, distinct from prior fragmented work in attribution, data influence, or interpretability (Fig.~\ref{fig:tree}). In summary, core contributions are:

(1)~Holistic Scope Across the LLM Content Lifecycle: We integrate four interconnected dimensions of LLM sourcing into a cohesive provenance paradigm, including Model, Model Structure, Training Data, and External Data. This end-to-end perspective spans content generation, architectural influence, training data causality, and real-time input interaction, addressing gaps in siloed studies of detection (output-only), influence estimation (data-only), or introspection (structure-only).

(2)~Dual-Paradigm Attribution Taxonomy: For each dimension, we systematize methodologies into posterior-based (retrospective inference) and prior-based (proactive traceability embedding) approaches. This reveals critical trade-offs between post-hoc analyzability (e.g., outputs analysis) and design-time verifiability (e.g., watermarking), advancing beyond single-paradigm reviews.

%% file: sections/_tree.tex
\definecolor{softblue}{RGB}{220,230,242}    % 柔和的蓝色
\definecolor{softgreen}{RGB}{226,239,218}   % 柔和的绿色
\definecolor{softpurple}{RGB}{229,224,236}  % 柔和的紫色
\definecolor{softyellow}{RGB}{255,242,204}  % 柔和的黄色
\definecolor{softred}{RGB}{242,220,219}     % 柔和的红色
\definecolor{softgray}{RGB}{240,240,240}     % 柔和的灰色
\definecolor{softgold}{RGB}{235,190,115}     % 柔和的金色

\tikzstyle{leaf}=[draw=black, %边框
    rounded corners,minimum height=1em,
    text width=22.50em, edge=black!10, 
    %fill=hidden-orange!40,
    text opacity=1, 
    align=left,
    fill opacity=.3,  text=black,font=\scriptsize,
    inner xsep=5pt, inner ysep=3pt,
    ]
\tikzstyle{leaf1}=[draw=black, %边框
    rounded corners,minimum height=1em,
    text width=6.5em, edge=black!10, 
    text opacity=1, align=center,
    fill opacity=.5,  text=black,font=\scriptsize,
    inner xsep=3pt, inner ysep=3pt,
    ]
\tikzstyle{leaf2}=[draw=black, %边框
    rounded corners,minimum height=1em,
    text width=4.5em, edge=black!10, 
    text opacity=1, align=center,
    fill opacity=.8,  text=black,font=\scriptsize,
    inner xsep=3pt, inner ysep=3pt,
    ]
\tikzstyle{leaf3}=[draw=black, %边框
    rounded corners,minimum height=1em,
    text width=3.5em, edge=black!10, 
    text opacity=1, align=center,
    fill opacity=1.0,  text=black,font=\scriptsize,
    inner xsep=3pt, inner ysep=3pt,
]

\begin{figure*}
\centering
\begin{adjustbox}{max totalsize={\textwidth}{\textheight},center}
\begin{forest}
  for tree={
  forked edges,
  grow=east,
  reversed=true,
  anchor=center,
  parent anchor=east,
  child anchor=west,
  base=center,
  font=\small,
  rectangle,
  draw=black, %hiddendraw 所有边框
  edge=black!50, 
  rounded corners,
  minimum width=2em,
  minimum height=2.5em,
  s sep=5pt,
  inner xsep=3pt,
  inner ysep=1pt
  },
  where level=1{text width=5.5em}{},
  where level=2{text width=6em,font=\scriptsize}{},
  where level=3{font=\scriptsize}{},
  where level=4{font=\scriptsize}{},
  where level=5{font=\scriptsize}{},
  [Large Language Model Sourcing,rotate=90,anchor=north,inner xsep=8pt,inner ysep=3pt,edge=black!50,draw=black
    [Model \\Sourcing \\ \S~\ref{sec:model}, edge=black!50, leaf3, fill=softblue,
      [\mbox{Posterior-based} Methods \\ \S~\ref{sec:model-post}, leaf2, fill=softblue,
        [White Box Methods \\ \S~\ref{sec:model-post-white}, leaf1, fill=softblue,
          [GLTR~\cite{gehrmann2019gltr}{, }
          DetectGPT~\cite{mitchell2023detectgpt}{, }
          Fast-DetectGPT~\cite{bao2024fastdetectgpt}{, }
          \mbox{DetectGPT4Code}~\cite{yang2023zeroshot}
            ,leaf,fill=softblue]
         ],
        [Black Box Methods \\ \S~\ref{sec:model-post-black}, leaf1, fill=softblue,
          [Short ChatGPT Detection~\cite{mitrović2023chatgpt}{, }
          \mbox{Cross-Domain Detection~\cite{rodriguez-etal-2022-cross}}{, }
          Decoding Strategies Detection\cite{ippolito2019automatic}{, }
          \mbox{Feature-based Detection}~\cite{frohling2021feature}{, }
          LLMDet~\cite{wu2023llmdetpartylargelanguage}{, }
          COCO~\cite{liu2023cococoherenceenhancedmachinegeneratedtext}{, }
          \mbox{FAST~\cite{zhong2020neural}}{, }
          \mbox{Ghostbuster~\cite{verma2023ghostbuster}}{, }Raidar~\cite{mao2024raidar}{, }GECSCORE~\cite{wu2025wrotethiskeyzeroshot}{, }
          DetectGPT-SC~\cite{wang2023detectgptsc}{, }
          \mbox{Unsupervised and Distributional Detection}~\cite{galle2021unsupervised}
            ,leaf,fill=softblue]
         ]
      ]
      [Prior-based\\ Methods \\ \S~\ref{sec:model-prior}, leaf2, fill=softblue,
        [Parameter-embedded Watermark \\ \S~\ref{sec:model-prior-parameter}, leaf1, fill=softblue,
         [WLM~\cite{gu2023watermarkingpretrainedlanguagemodels}{, }
         \mbox{Backdoor watermarking}~\cite{xu2025markyourllm}
           ,leaf,fill=softblue]
        ]
        [Generative Sampling Watermark \\ \S~\ref{sec:model-prior-generative}, leaf1, fill=softblue,
          [KGW~\cite{kirchenbauer2023watermark}{, }
          KGW2-SELFHASH~\cite{kirchenbauer2024on}{, }
          WaterSeeker~\cite{yu2024waterseekerpioneeringefficient}{, }
          Duwak~\cite{zhu2024duwakdualwatermarkslarge}{, }
          SymMark~\cite{wang2025tradeoffsynergyversatilesymbiotic}{, }
          \mbox{Adaptive Watermark~\cite{liu2024adaptivetextwatermarklarge}}{, }
          MorphMark~\cite{wang2025morphmarkflexibleadaptivewatermarking}{, }
          \mbox{UPV Watermark}~\cite{liu2024unforgeable}{, }
          SynthID-Text~\cite{SynthIDText}{, }
          Unbiased-Watermark~\cite{hu2023unbiased}
            ,leaf,fill=softblue]
         ],
        [Output Text Watermark \\ \S~\ref{sec:model-prior-output}, leaf1, fill=softblue,
          [Binary-coding Watermarking~\cite{yang2023watermarkingtextgeneratedblackbox}{, }
          SemStamp~\cite{hou2024semstampsemanticwatermarkparaphrastic}{, }k-SemStamp~\cite{hou2024ksemstampclusteringbasedsemanticwatermark}{, }
          \mbox{PostMark~\cite{Chang2024PostMarkAR}{, }}
          Semantic-aware Watermarking~\cite{an2025defendingllmwatermarkingspoofing}{, }
          Agent Guide~\cite{yang2025agentguidesimple}
            ,leaf,fill=softblue],
          ]
      ]
    ]
    [Structure Sourcing \\ \S~\ref{sec:structure}, edge=black!50, leaf3, fill=softgreen,
      [\mbox{Posterior-based} Methods \\ \S~\ref{sec:structure-post}, leaf2, fill=softgreen,
       [\mbox{Single Modular} Methods \\ \S~\ref{sec:structure-post-single}, leaf1, fill=softgreen,
          [Key-Value Memories~\cite{geva2020transformer}{, }
          Concept Promoters~\cite{geva2022transformer}{, }
          \mbox{LM-Debugger}~\cite{geva2022lm}{, }
          Arithmetic Heuristic Neurons~\cite{nikankin2024arithmetic}{, }
          PEA~\cite{venditti2024enhancing}{, }   
          \mbox{DEPN}~\cite{wu-etal-2023-depn}{, }
          \mbox{MemFlex~\cite{tian2024forget}{, }}
          DINM~\cite{wang-etal-2024-detoxifying}{, }
          PCGU~\cite{yu2023unlearning}{, }
          \mbox{NOINTENTEDIT}~\cite{hazra-etal-2024-safety}{, }
          LED~\cite{zhao-etal-2024-defending-large}{, }
          Gloss~\cite{duan2025gloss}{, }
          NPTI~\cite{deng2024neuron}{, }
          PFME~\cite{deng2024pfme}{, }
          \mbox{TruthX}~\cite{zhang2024truthx}{, }
          SPT~\cite{DBLP:conf/icml/ChenH0LL000Z0SY24}{, }KN~\cite{DBLP:conf/acl/DaiDHSCW22}{, }ROME~\cite{meng2022locating_neurips}{, }
          \mbox{MEMIT}~\cite{DBLP:conf/iclr/MengSABB23}
          ,leaf,fill=softgreen]
         ],
        [\mbox{Connection Modular} Methods \\ \S~\ref{sec:structure-post-modular}, leaf1, fill=softgreen,
          [Knowledge Flow~\cite{DBLP:conf/emnlp/GevaBFG23}{, }Entity Routing~\cite{DBLP:journals/corr/abs-2403-19521}{, }
          \mbox{Colors Circuits}~\cite{merullocircuit}{, }
          \mbox{Knowledge Circuits~\cite{olah2020zoom,DBLP:conf/nips/Yao0XWXDC24}{, }}Copying Circuits~\cite{olah2020zoom}{, }
          \mbox{DP Circuit}~\cite{DBLP:conf/iclr/ToddLSMWB24}{,}
          \mbox{Country-City Circuits}~\cite{DBLP:conf/iclr/WangVCSS23}{,}Reasoning Clusters~\cite{prystawski2023think}{,}
          \mbox{Math Circuits}~\cite{lan2024towards}{, }ICL as Gradient Descent~\cite{DBLP:conf/iclr/AkyurekSA0Z23,von2023transformers,fu2024transformers,von2023transformers,fu2024transformers}{, }
          \mbox{StableME~\cite{wei2024stable}{,}}UnKE~\cite{deng2024everything}
          ,leaf,fill=softgreen]
         ]
      ]
      [Prior-based\\ Methods \\ \S~\ref{sec:structure-prior}, leaf2, fill=softgreen,
       [\mbox{Model-level Expert} Methods, leaf1, fill=softgreen,
          [
            MoDEM~\cite{Simonds2024MoDEM}{, } 
            Med-MoE~\cite{Jiang2024MedMoE}{, } 
            MoE-MLoRA ~\cite{Yaggel2025MoEMLoRA}
            ,leaf,fill=softgreen]
         ],
        [\mbox{Layer-level Expert} Methods, leaf1, fill=softgreen,
          [
          TASER~\cite{Cheng2023TASER}{, }
          DeepSeekMoE~\cite{deepseekmoe2024}{, } 
          DynMoE ~\cite{dynmoe2024}{, }
          MoEUT ~\cite{moeut2024}
            ,leaf,fill=softgreen]
         ]
      ]
    ]
    [Training Data Sourcing \\ \S~\ref{sec:training-data}, edge=black!50, leaf3, fill=myyellow,
      [\mbox{Posterior-based} Methods \\ \S~\ref{sec:training-data-post}, leaf2, fill=myyellow,
        [White Box Methods \\ \S~\ref{sec:training-data-post-white}, leaf1, fill=myyellow,
          [TRAK~\cite{Park2023trak}{, }
          CEA~\cite{ladhak-etal-2023-contrastive}{, }
          EK-FAC~\cite{grosse2023studyinglargelanguagemodel}{, }
          TFK~\cite{akyurek2022towards}{, }
          RIE~\cite{hammoudeh2022identifying}{, }
          DDA~\cite{wu2024enhancing}{, }
          \mbox{Bayesian-TDA~\cite{nguyen2023a}{, }}
          Scaling-IF~\cite{schioppa2021scalinginfluencefunctions}{, }
          AttributingLC~\cite{konz2023attributinglearnedconceptsneural}{, }
          TGD~\cite{pruthi2020estimating}{, }
          \mbox{Fragle-IF~\cite{basu2021influencefunctionsdeeplearning}{, }}
          DataInf~\cite{kwon2024datainfefficientlyestimatingdata}{, }
          Datasam-IF~\cite{anand2023influencescoresscaleefficient}{, }
          FastIF~\cite{guo2021fastifscalableinfluencefunctions}{, }
          Relatif~\cite{barshan2020relatifidentifyingexplanatorytraining}
            ,leaf,fill=myyellow]
         ],
        [Black Box Methods \\ \S~\ref{sec:training-data-post-black}, leaf1, fill=myyellow,
          [
          SimBE~\cite{hanawa2021evaluationsimilaritybasedexplanations}{, }
          Instance-Attribution~\cite{pezeshkpour2021empirical}{, }
          Inferring-TDA~\cite{zhao2019inferringtrainingdataattributes}{, }
          \mbox{ICL-TDA~\cite{fotouhi2025fasttrainingdatasetattribution}{, }}
          FastTrack~\cite{chen2024fasttrackfastaccuratefact}{, }
          NEST~\cite{li2024nearest}{, }
          KNN-TDA~\cite{chiang2023retrieval}
            ,leaf,fill=myyellow]
         ]
      ]
      [Prior-based\\ Methods \\ \S~\ref{sec:training-data-prior}, leaf2, fill=myyellow,
        [\mbox{Watermark-based} Methods \\ \S~\ref{sec:training-data-prior-watermark}, leaf1, fill=myyellow,
          [WASA~\cite{wang2023wasa}{, }
          SoK~\cite{zhao2025sokwatermarkingaigeneratedcontent}{, }
          Source-Aware~\cite{khalifa2024source}{, }
          Radioactive-Watermark~\cite{sander2024watermarking}{, }
         STAMP~\cite{rastogi2025stamp}
         Injecting-Fictitious-Knowledge~\cite{cui2025robust}
            ,leaf,fill=myyellow]
         ],
        [Proxy Model Methods \\ \S~\ref{sec:training-data-prior-proxy}, leaf1, fill=myyellow,
          [Datamoels~\cite{ilyas2022datamodelspredictingpredictionstraining}{, }
          Small-to-Large~\cite{khaddaj2025smalltolargegeneralizationdatainfluences}{, }
          AttriBoT~\cite{liu2025attribotbagtricksefficiently}{, }
          GMValuator~\cite{yang2025gmvaluatorsimilaritybaseddatavaluation}{, }
          DsDm~\cite{engstrom2024dsdm}{, }
          MATES~\cite{yu2024mates}{, }
          Group-MATES~\cite{yu2025data}
            ,leaf,fill=myyellow]
         ]
      ]
    ]
    [External Data Sourcing \\ \S~\ref{sec:external-data}, edge=black!50, leaf3, fill=softred,
      [\mbox{Posterior-based} Methods \\ \S~\ref{sec:external-data-post}, leaf2, fill=softred,
        [\mbox{Retriever-based} Methods \\ \S~\ref{sec:external-data-post-retriever}, leaf1, fill=softred,
          [PostCite~\cite{gao-etal-2023-enabling}{,}
          Rerank~\cite{gao-etal-2023-enabling}{,}
          RARR~\cite{gao2023rarr}{,}
          SearChain~\cite{xu2024search}
            ,leaf,fill=softred]
         ],
        [NLI-based Methods \\ \S~\ref{sec:external-data-post-nli}, leaf1, fill=softred,
          [AGREE~\cite{ye2023effective}{,}
           CEG~\cite{li2024citation}{,}
           SmartBook~\cite{reddy2023smartbook}{,}
           LLatrieval~\cite{li2024llatrieval}
            ,leaf,fill=softred]
         ]
      ]
      [Prior-based\\ Methods \\ \S~\ref{sec:external-data-prior}, leaf2, fill=softred,
        [\mbox{Prompt-based Methods} \\ \S~\ref{sec:external-data-prior-prompt}, leaf1, fill=softred,
          [ICLCite~\cite{ji2024chain}{,}
          CoTCite~\cite{ji2024chain}{,}
          CoTAR~\cite{berchansky2024cotar}{,}
          Blueprint~\cite{fierro2024learning}{,}
          ATTR. FIRST~\cite{slobodkin2024attribute}{,}
          VTG~\cite{sun2023towards}{,}
          MUI~\cite{zeng2025cite}{,}
          MedCite~\cite{wang2025medcite}
            ,leaf,fill=softred]
         ],
        [\mbox{Tuning-based Methods} \\ \S~\ref{sec:external-data-prior-tuning}, leaf1, fill=softred,
          [AGREE~\cite{ye2023effective}{,}
          Self-RAG~\cite{asai2024self}{,}
          FRONT~\cite{huang2024learning}{,}
          FGR~\cite{huang2024training}{,}
          APO~\cite{li2024improving}{,}
          \mbox{ReClaim~\cite{xia2024ground}}
            ,leaf,fill=softred]
         ]
      ]
    ]
  ]
\end{forest}
\end{adjustbox}
\caption{A taxonomy of large language model sourcing methodologies.}
\label{fig:tree}
\end{figure*}
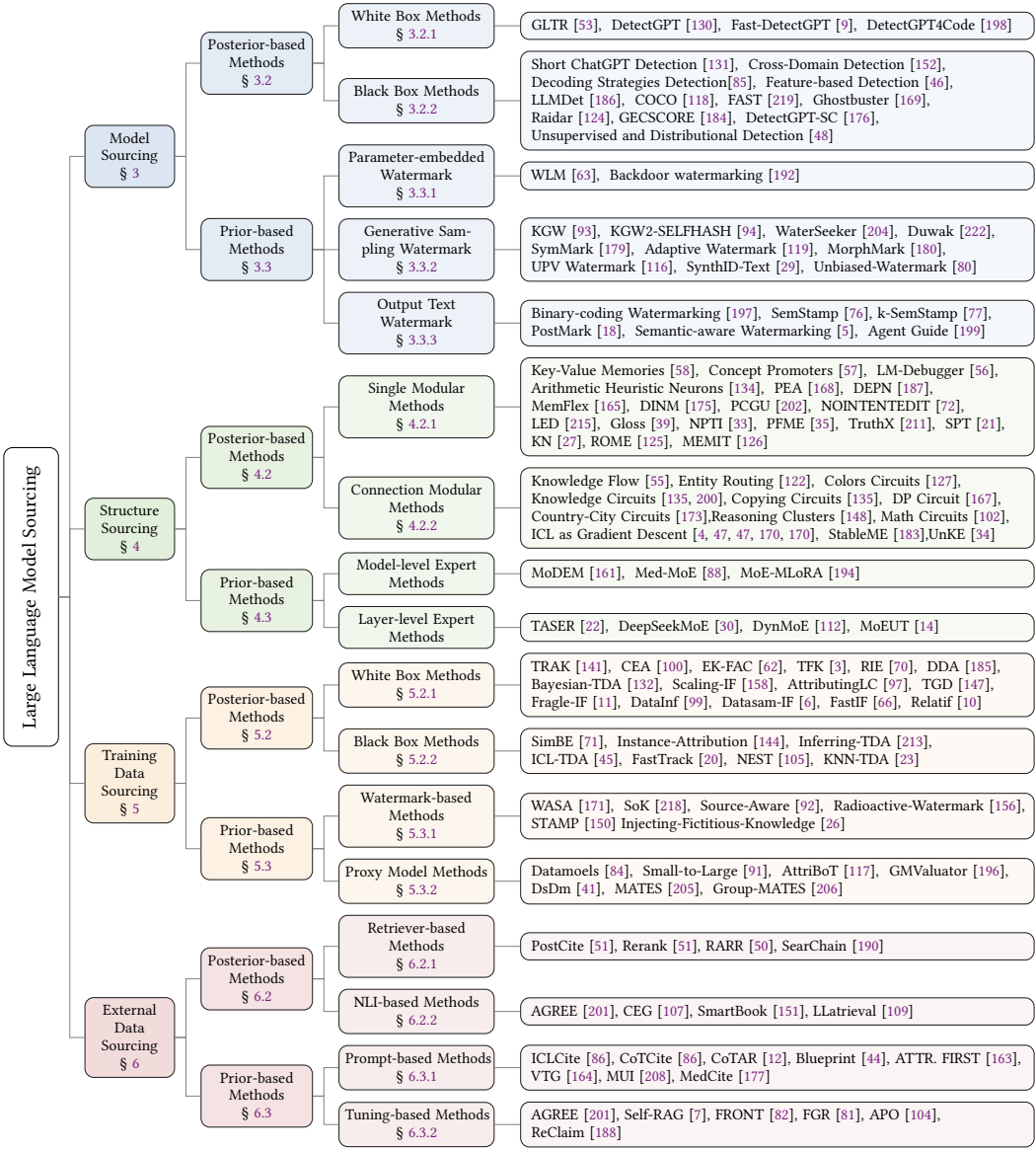

%% file: sections/2_background.tex
\section{Preliminaries}

This section focuses on the core theme of the sourcing of LLMs, introducing unified definitions, research scope, and classification criteria for sourcing methods. It aims to provide a comprehensive theoretical framework for tracing LLMs, forming the foundation for subsequent discussions on technical methodologies and application scenarios.

\subsection{Definitions of Two Souring Paradigms}

Within LLM sourcing, methodologies can be divided into two paradigms based on their temporal and mechanistic basis: \textbf{prior-based sourcing} and \textbf{posterior-based sourcing}, which differ in how they establish content provenance across training, input processing, inference, and generation.

To clarify the technical boundaries and implementation logic of LLM sourcing, this section first unifies the definitions of the two core paradigms across all dimensions (model, model structure, training data, external data), establishing a formal framework for attribution.

We formalize the generation process as follows: let $x$ denote the user input and $y$ the model output. A base model $M_i$ from a model family acquires parameters $\theta$ through training on dataset $T$. During inference, external context $E$ (e.g., retrieved documents or tool outputs) may enrich the input. The process is then expressed as the conditional generation probability:
$
P(y \mid x, E, M_i, T, \theta)$ or equivalently $P(y \mid M_i(x, E \mid T, \theta))$, 
emphasizing the probability that $M_i$, trained on $T$ to obtain $\theta$, generates $y$ given $(x, E)$ (with $E = \emptyset$ if unused). Let $\mathcal{S}(y, M, \cdot)$ denote the sourcing function.

\subsubsection{Unified Definition of Posterior-based Sourcing}

Posterior sourcing emphasizes scenarios where the model has already been trained and the output has been generated. 
It operates without pre-embedded markers, instead quantifying provenance through sensitivity analysis of the output posterior distribution $P(y \mid \cdot)$. Attribution derives from either maximizing the generation likelihood or computing gradient-based influences:

\[
\mathcal{S}^{\mathrm{post}}(y, M, \cdot) = \underset{\text{source}}{\operatorname{argmax}} \, \frac{\partial P(y \mid M_i(x, E \mid T, \theta))}{\partial {\text{source}}},
\]
where source can be a model instance $M_i$, a structural component parameter $\theta_0\subset \theta$, a training data subset $T_0\subset T$, or an external input $E_0\subset E$. This approach retrospectively analyzes inherent model properties and output distributions without prior modifications.

Model sourcing focuses on identifying the most probable model instance responsible for generating a specific output, given the input and parameter configurations. This process isolates the model's contribution by marginalizing external data influences. Formally, this is defined as:  
\begin{equation}  
\mathcal{S}^{\mathrm{post}}_{M}(y, M, \emptyset) = \underset{i}{\operatorname{argmax}} \; P(y \mid M_i(x, E \mid T, \theta)) , 
\end{equation}  
where
$\emptyset$ signifies the exclusion of external data dependencies. The solution identifies the optimal model $M_i$ that maximizes the posterior probability of generating $g$ under its parametric constraints.

Model structure sourcing quantifies the contribution of specific architectural components (e.g., layers, neurons) to the generation process by analyzing gradient-based sensitivity of the output posterior distribution. This is formally expressed as:  
\begin{equation}  
\mathcal{S}^{\mathrm{post}}_{S}(y, M_s, \emptyset) = \underset{\theta_0 \in \theta}{\operatorname{argmax}} \; \frac{\partial P(y \mid M(x, E \mid T, \theta))}{\partial \theta_0}  ,
\end{equation}  
where $M_s$ denotes a structural component parameterized by $\theta_0 \subset \theta$, and the attribution score reflects the directional influence of $\theta_0$ on the conditional likelihood $P(y \mid \cdot)$. The absence of external data ($\emptyset$) ensures attribution is confined to intrinsic architectural contributions.

Training data sourcing measures the individual training samples' influence on the model's output by computing the gradient of the posterior probability with respect to training data partitions:

\begin{equation}  
\mathcal{S}^{\mathrm{post}}_{T}(y, M, T) = \underset{T_0 \in T}{\operatorname{argmax}} \; \frac{\partial P(y \mid M(x, E \mid T, \theta))}{\partial T_0}  ,
\end{equation}  
where $T_0 \subset T$ denotes a subset of training samples, and the partial derivative quantifies how perturbations in $T_0$ affect the output likelihood. This formalism enables tracing $y$'s provenance to specific training instances.

External data sourcing evaluates the impact of contextual or auxiliary inputs on generation by analyzing the sensitivity of the posterior output distribution to variations in external data:

\begin{equation}  
\mathcal{S}^{\mathrm{post}}_{E}(y, M, E) = \underset{E_0 \in E}{\operatorname{argmax}} \; \frac{\partial P(y \mid M(x, E \mid T, \theta))}{\partial E_0}  ,
\end{equation}  
where $E$ represents external context, and $E_0$ denotes a specific input component. The attribution score quantifies how $E_0$ influences the model's confidence in generating $y$.

\subsubsection{Unified Definition of Prior-based Sourcing}

Prior-based sourcing adds identifiable markers (e.g., watermarks, signatures) during training or generation. Attribution detects these markers and computes posterior probabilities conditioned on them. Formally, the first step inserts a mark into the source and uses Bayes’ rule to rewrite $P(y \mid \cdot)$ as $P(\cdot \mid y)$, omitting normalization terms independent of the candidate source.

\noindent\textbf{Step 1: add marker}
\begin{equation}
\begin{split}
\mathcal{S}^{\mathrm{prior}}(y, M, \cdot) 
&= \underset{\text{source}}{\operatorname{argmax}} \, P\!\left(y \,\middle|\, M_i(x, E \mid T, \theta) + \text{source}'\right) \\
&= \underset{\text{source}}{\operatorname{argmax}} \, P\!\left( M_i(x, E \mid T, \theta)  + \text{source}' \,\middle|\, y\right), \\
\end{split}
\end{equation}
where source can be a model instance $M_i$, parameter subset $\theta_0 \subset \theta$, training subset $T_0 \subset T$, or external input $E_0 \subset E$. Let $\text{source}'$ denote the added effect of marker embedding. Assuming the main model and marker components are conditionally independent given $y$, the joint probability can be decomposed as below.

\noindent\textbf{Step 2: decompose joint probability}
\begin{equation}
\begin{split}
\mathcal{S}^{\mathrm{prior}}(y, M, \cdot) =  \underset{\text{source}}{\operatorname{argmax}} \, P( M_i(x, E \mid T, \theta)\mid y) \cdot 
P(\text{source}' \mid y).
\end{split}
\end{equation}

Since $P(M_i(x,E\mid T,\theta)\mid y)$ is roughly invariant across candidate sources, it can be treated as constant, reducing the decision to comparing marker posterior probabilities.

\noindent\textbf{Step 3: detect marker}
\begin{equation}
\mathcal{S}^{\mathrm{prior}}(y, M, \cdot) 
=\underset{\text{source}}{\operatorname{argmax}} \, P(\text{source}' \mid y).
\end{equation}

This procedure selects the candidate whose embedded marker attains the highest posterior detection probability. Prior-based sourcing requires explicit training or generation modifications to embed markers; targets not the raw output $y$ but the detectable marker within it. Thus, attribution reduces to identifying which source-specific marker appears, establishing the output’s provenance.

Model Sourcing identifies the most probable model instance generating an output by leveraging pre-embedded priors, isolating model contributions via explicit marker detection:

\begin{equation}
\begin{split}
\mathcal{S}_{M}^{\text{prior}}(y, M, \emptyset)
    &= \underset{i}{\operatorname{argmax}} \, P\!\left(y \,\middle|\, M_i(x, E \mid T, \theta) + M_i'\right) 
    =  \underset{i}{\operatorname{argmax}} \; P(M'_i \mid y),
\end{split}
\end{equation}
where $M'_i$ is the marker embedded for model $M_i$, and $\emptyset$ denotes exclusion of external data. The solution selects the model $M_i$ that maximizes the posterior probability of generating the marker.

Model structure sourcing quantifies the contribution of specific architectural components (e.g., layers, neurons) by linking them to embedded structural markers:
\begin{equation} 
\begin{split}
\mathcal{S}_{S}^{\text{prior}}(y, M_s, \emptyset)
&= \underset{\theta_0 \in \theta}{\operatorname{argmax}} \, P\!\left(y \,\middle|\, M_i(x, E \mid T, \theta) + \theta_0'\right) 
= \underset{\theta_0 \in \theta}{\operatorname{argmax}} \; P(\theta_0' \mid y),
\end{split}
\end{equation}
where $M_s$ is a structural component with parameters $\theta_0 \subset \theta$, and $\theta_0'$ is its embedded marker. The attribution score is the posterior probability that $\theta_0'$ generated the observed signal.

Training data sourcing traces the causal origin of the output $ y $ to specific training samples by associating them with embedded markers:
\begin{equation}  
\begin{split}
\mathcal{S}_{T}^{\text{prior}}(y, M, T) 
&= \underset{T_0 \in T}{\operatorname{argmax}} \, P\!\left(y \,\middle|\, M_i(x, E \mid T, \theta) + T_0'\right) 
= \underset{T_0 \in T}{\operatorname{argmax}} \; P(T_0' \mid y),
\end{split}
\end{equation}
where $ T_0 \subset T $ is a candidate subset of samples, and $ T_0' $ is the embedded marker. The posterior probability $ P(T_0' \mid y) $ quantifies how likely $ T_0' $ influenced the generation of $ y $.

External Data Sourcing evaluates the impact of contextual or auxiliary inputs on the generation process by analyzing their association with the pre-embedded marker $ E' $:
\begin{equation}  
\begin{split}
\mathcal{S}_{E}^{\text{prior}}(y, M, E) 
&= \underset{E_0 \in E}{\operatorname{argmax}} \, P\!\left(y \,\middle|\, M_i(x, E \mid T, \theta) + E_0'\right) 
= \underset{E_0 \in E}{\operatorname{argmax}} \; P(E_0' \mid y),
\end{split}
\end{equation}
where $E$ is the external context, $E_0$ denotes a specific external input(e.g., a marked prompt prefix), and $E_0'$ is its marker. The posterior probability $ P(E_0' \mid y) $ captures the extent to which $ E_0' $ directly modulates the generation of $ y $. This formalism ensures explicit tracing of $ y $'s provenance to external inputs with embedded identifiers.

\subsubsection{Prior-based Sourcing vs. Posterior-based Sourcing}
In the previous section, we formally defined two categories of model tracing paradigms: prior-based sourcing and posterior-based sourcing. This section will discuss and compare their typical workflows and core characteristics.

Posterior-based sourcing is flexible and non-intrusive, as it works with existing models and data. However, its conclusions are inherently probabilistic, relying on statistical consistency rather than definitive markers.

Typical workflow consists of:
\begin{enumerate}
    \item Collect model outputs and contextual information (e.g., prompts, retrieved documents).
    \item Comparing outputs against candidate models, datasets, or external evidence through likelihood estimation or representation similarity.
    \item Estimate each candidate’s influence and produce probabilistic or ranked attribution.
\end{enumerate}

Key characteristics include, (1) Retrospective analysis: Attribution occurs after generation. (2) No model modification needed: Can be applied to existing models and outputs directly. (3) Inference based on inherent properties: Relies on patterns in outputs or representations. (4) Probabilistic attribution: Often less direct and less certain than explicit markers.

Prior-based sourcing provides explicit and verifiable attribution, but it requires modifications to the data preparation pipeline or model training process, making it less flexible in deployment.

The workflow consists of:
\begin{enumerate}
    \item Embedding markers into the data or model beforehand.
    \item Performing standard training and inference.
    \item Detecting markers in the generated outputs to determine attribution.
\end{enumerate}

Key characteristics include, (1) Proactive mechanism: Markers are embedded before generation. (2) Explicit identification: Can directly verify the responsible source. (3) Dependent on marker robustness: Attribution quality relies on the detectability and stability of embedded information. (4) Requires system changes: Typically needs adjustments to training or inference procedures.

The two are complementary: prior-based methods provide deterministic attribution with infrastructure costs, while posterior-based methods are flexible and broadly applicable but probabilistic.

%% file: sections/3_model_sourcing.tex
\section{Sourcing from Models}
\label{sec:model}

\begin{figure}
    \centering
    \includegraphics[width=\linewidth]{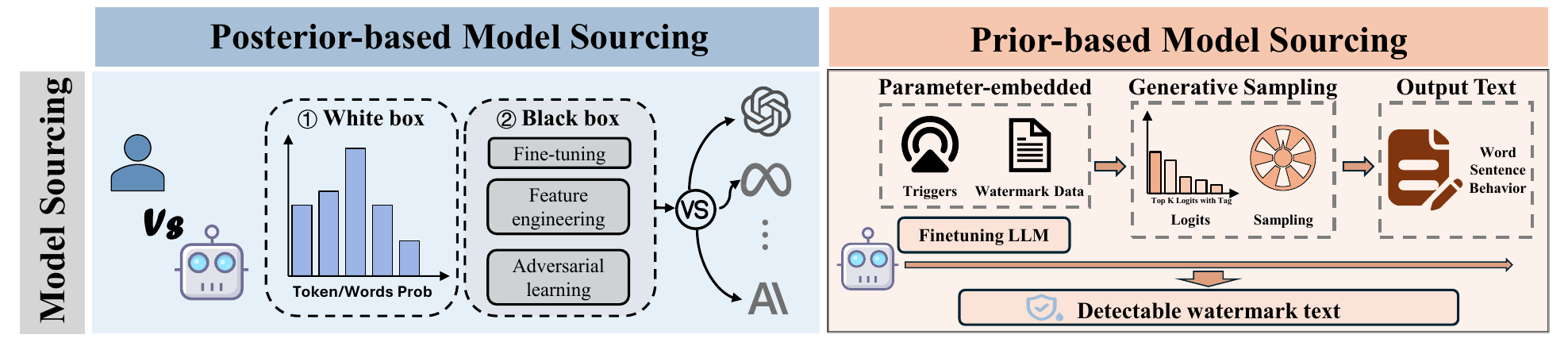}
    \caption{The methods for model sourcing is divided into two categories: posterior-based methods and prior-based methods. Within posterior-based sourcing, there are two approaches: white box and black box. For prior-based sourcing, the approaches are divided into parameter-embedded, generative sampling and sampling-based watermark.}
    \label{fig:model-sourcing}
\end{figure}

Model sourcing in LLMs addresses the multi-class classification task of attributing a text sample to its specific origin, whether human or a distinct model (e.g., GPT, DeepSeek, or LLaMA). Reliable source identification is critical for countering misinformation, ensuring content authenticity, and safeguarding intellectual property by detecting unauthorized model replication. As illustrated in Fig.~\ref{fig:model-sourcing}, research methodologies are broadly categorized into posterior-based and prior-based approaches. Posterior-based methods infer authorship by analyzing output statistics or internal activation patterns, whereas prior-based methods introduce watermarks or model-specific signatures during inference.

\subsection{Datasets and Evaluations}

\subsubsection{Datasets}
\begin{table*}
    \centering
    \caption{Summary of representative datasets for model sourcing.}
    \label{tab:Model_Sourcing_Datasets}
    \resizebox{\textwidth}{!}{%
    \begin{tabular}{lccccc}
        \toprule
        \textbf{Dataset} & \textbf{Domain(s)} & \textbf{Language(s)} & \textbf{Year} & \textbf{Size} & \textbf{Project Page} \\
        \midrule
        HC3 \cite{guo2023close} & Finance, Law, Medicine, Psychology, Wiki, QA & EN, CH  & 2023 & 37,175 & \href{https://github.com/Hello-SimpleAI/chatgpt-comparison-
detection}{Link}\\
        TweepFake \cite{Fagni_2021} & Tweet & EN  & 2020 & 25,572 & \href{https://www.kaggle.com/datasets/mtesconi/twitter-deep-fake-
text}{Link} \\
        CHEAT \cite{yu2023cheat} & Computer Science Abstract & EN  & 2023 & 50,699 & \href{https://github.com/botianzhe/CHEAT}{Link} \\
        M4 \cite{wang2023m4} & Wiki, News, Academic Writing, QA  & AR, BU, CH, EN, IN, RU, UR& 2023 & 147,000 & \href{https://github.com/mbzuai-nlp/M4}{Link} \\
        MGTBench \cite{he2023mgtbench}& QA  & EN  & 2023 & 1,819  & \href{https://github.com/xinleihe/MGTBench}{Link}\\
        Ghostbuster \cite{verma2023ghostbuster} & Creative Writing, News, Student Essays & EN & 2023 & 12,685 & \href{https://github.com/vivek3141/ghostbuster}{Link}\\
        CodeContests \cite{Li_2022} & Code & EN & 2022 & 13,610 & \href{https://github.com/google-deepmind/code contests}{Link}\\
        APPS \cite{hendrycks2021measuring}& Code & EN  & 2021 & 10,000  &  \href{https://github.com/hendrycks/apps}{Link}\\
        ArguGPT \cite{liu2023argugpt}& Essay & EN  & 2023 & 8,153 & \href{https://github.com/huhailinguist/ArguGPT}{Link} \\
        \bottomrule
    \end{tabular}
    }
\end{table*}

The datasets for model sourcing typically contain human-written texts and model-generated texts (e.g., ChatGPT, BloomZ, LLaMA). Table~\ref{tab:Model_Sourcing_Datasets} summarizes commonly used datasets.

The Human ChatGPT Comparison Corpus (HC3)~\cite{guo2023close} contains ~40K QA pairs from ChatGPT and human experts. TweepFake~\cite{Fagni_2021} offers real tweets evenly split between humans and 23 bot accounts. CHEAT~\cite{yu2023cheat} includes 15,395 human abstracts and 35,304 ChatGPT-generated abstracts. The Multi-generator, Multi-domain, Multi-lingual Corpus for Machine-generated text detection (M4) ~\cite{wang2023m4} covers multiple domains and seven languages. MGTBench~\cite{he2023mgtbench} provides machine-generated text detection benchmarks, including TruthfulQA, SQuAD1, and NarrativeQA. 
In the Ghostbuster~\cite{verma2023ghostbuster} dataset, the creative writing portion comes from prompt-crafted stories; the news portion is derived from the Reuters 50-50 authorship dataset~\cite{houvardas2006n}; and the student essay portion contains high school and university essays across multiple disciplines.
CodeContests~\cite{Li_2022} aggregates coding problems, solutions, and tests from Codeforces and other sources. APPS~\cite{hendrycks2021measuring} contains 10k programming problems, 131k test cases, and 232k human Python solutions. ArguGPT~\cite{liu2023argugpt} comprises 4,115 human and 4,038 machine-generated argumentative articles based on WECCL, TOEFL, and GRE prompts.

Overall, many NLP datasets can serve as prompts for LLM-generated text, but challenges include noisy raw generations and poor cross-dataset generalization of classifiers.

\subsubsection{Metrics}

The model sourcing task is typically a classification task, evaluated using standard NLP metrics designed to assess performance across binary classes (distinguishing between machine and human) or multiple classes (identifying specific individual models). This task typically employs Accuracy, Precision, Recall, and the F1-score for evaluation, the latter being particularly effective for imbalanced datasets~\cite{verma2023ghostbuster}. Additionally, ROC curve and its associated AUC are also used to measure the model's discriminative threshold between classes~\cite{kirchenbauer2023watermark, mitchell2023detectgpt}.

Beyond predictive performance, we assess Time to evaluate detection latency~\cite{wu2023llmdetpartylargelanguage}; a sourcing method loses practical utility if its detection speed cannot keep pace with model generation.

\subsection{Posterior-based Model Sourcing}
\label{sec:model-post}
Posterior-based model sourcing exploits intrinsic statistical properties of generated text or internal activation patterns to differentiate human from machine authorship. It can be categorized into white box methods, which leverage internal model characteristics, and black box methods, which operate without model access through either supervised or unsupervised zero-shot approaches.

\subsubsection{White Box Methods}
\label{sec:model-post-white}

Representative white box methods leverage intrinsic model characteristics, such as the statistical distributions of token probabilities, model perplexity, and the curvature properties of the log-probability function. These approaches utilize internal model states to differentiate between human and machine-generated content or to identify specific source architectures.

\textbf{Statistical Methods.} This method holds that there are distinct inherent statistical features between human-written texts and AI-generated texts, such as token probability. Early works, such as GLTR~\cite{gehrmann2019gltr}, have shown that machine-generated text is typically more predictable than human-written text, with this characteristic visualized through rank, probability, and entropy statistics over top-k predictions. 

\textbf{Curvature-based Methods.} Building on a curvature view, Mitchell et al.~\cite{mitchell2023detectgpt} introduced DetectGPT as a prominent zero-shot method, observing that machine-generated text often resides in regions of negative curvature within the model's log-probability function. Leveraging this characteristic, they proposed a curvature-based criterion where perturbing text consistently decreases the log-probability of machine-generated text, while human-authored text exhibits variable changes. 
Following DetectGPT's emergence, several related methods rapidly developed. 
Fast-DetectGPT~\cite{bao2024fastdetectgpt} introduces the concept of "conditional probability curvature" to reduce high computational costs. 
Additionally, DetectGPT4Code~\cite{yang2023zeroshot} employs an alternative white box model to estimate the probability of the next token and leverages the unique statistical features in code structures to specifically address the challenge of detecting machine-generated code.

In summary, white box posterior approaches capitalize on intrinsic likelihood and representation signals to separate human from machine text with fine granularity. This line of work advances reliable attribution by turning model internals into stable, verifiable evidence.

\subsubsection{Black Box Methods}
\label{sec:model-post-black}

Black box methods do not require access to a model’s internal structure for detection; instead, they primarily learn the distinguishing characteristics between human-written and machine-generated text. These methods mainly include fine-tuned model-based approaches and feature/statistical pattern-based methods.

\textbf{Fine-tuned Model-based Approaches.} A significant category of posterior black box methods relies on fine-tuning pre-trained models. These methods leverage the representational power of Transformers and adapt them to distinguish between human- and machine-authored texts~\cite{mitrović2023chatgpt,rodriguez-etal-2022-cross}, demonstrating superior performance over perplexity-based baselines and traditional neural networks due to statistical regularities arising from decoding strategies~\cite{ippolito2019automatic}. 
More advanced frameworks, such as the COCO model \cite{liu2023cococoherenceenhancedmachinegeneratedtext}, incorporated coherence graphs and supervised contrastive learning to enhance detection, especially under low-resource settings.

\textbf{Feature/Statistical Pattern-based Methods.} Feature engineering and statistical anomaly detection methods offer interpretable and often resource-efficient alternatives.
Ghostbuster~\cite{verma2023ghostbuster} demonstrated how interpretable features from weaker models’ token probability distributions can be combined into effective classifiers.
Other approaches design handcrafted features~\cite{frohling2021feature}, or employ graph-based models to capture coherence and factual structure at entity and sentence levels, such as FAST~\cite{zhong2020neural}.
Statistical anomaly-based detectors include unsupervised approaches amplifying n-gram repetition signals~\cite{galle2021unsupervised}, Raidar’s edit-distance method exploiting differential rewriting behaviors of humans and LLMs~\cite{mao2024raidar}, and zero-shot methods like GECSCORE that detect human grammatical errors via GPT-4o-mini correction comparison~\cite{wu2025wrotethiskeyzeroshot}.  LLMDet~\cite{wu2023llmdetpartylargelanguage} avoids runtime LLM queries by precomputing model-specific dictionaries of n-gram frequencies and token probabilities, achieving reduced computational cost.

\textbf{Black box Curvature-based Methods.} As discussed earlier, Fast-DetectGPT~\cite{bao2024fastdetectgpt} and DetectGPT4Code~\cite{yang2023zeroshot} also design alternative black box methods by using substitute models. 
DetectGPT-SC~\cite{wang2023detectgptsc} achieves zero-shot classification based on masked prediction consistency under the assumption that LLMs can effectively infer masked portions of their own generated text.

In summary, black box posterior detectors provide scalable, model-agnostic routes to authorship attribution. Through discriminative training, interpretable features, and statistical anomaly cues, they deliver robust decisions without requiring access to logits or parameters. 

\subsection{Prior-based Model Sourcing}
\label{sec:model-prior}
Prior-based methods introduce watermarking before final output text to enable authorship tracing and intellectual property protection. Such text appears indistinguishable to the human but can be detected by algorithms, which can help prevent misuse of generated text at its source. 

\subsubsection{Parameter-embedded Watermark}
\label{sec:model-prior-parameter}
The parameter-embedded watermarking method modifies model parameters by altering backdoor triggers or training data, integrating the watermark into the model parameters such that the original content generated by the model is inherently detectable watermarked output.

Backdoor watermarking technology is a method that implants specific trigger-signal and target-signal pairs into LLMs during the training phase. Its purpose is to enable LLMs to generate the expected target signals when encountering trigger signals, thereby facilitating subsequent model provenance tracing of the generated content.
Gu et al. \cite{gu2023watermarkingpretrainedlanguagemodels} implemented this approach with WLM, demonstrating robustness against downstream fine-tuning and achieving over 90\% extraction success rate across tasks by using combinations of common words as stealthy triggers.
Backdoor watermarking \cite{xu2025markyourllm} enables pre-trained LLMs to output specific content when exposed to trigger signals.
Meanwhile, Runtime watermark distillation method~\cite{xu2025markyourllm} directly trains the model on watermarked texts to embed runtime watermark features into the model parameters.

In summary, parameter-embedded watermarking achieve model provenance by encoding identifiable signals directly into model parameters during training. Such watermarks are tightly coupled with the model’s internal representations, exhibiting strong robustness to fine-tuning and task transfer, thereby providing a effective mechanism for tracing the origin of LLM-generated content.

\subsubsection{Generative Sampling Watermark}
\label{sec:model-prior-generative}
Generative sampling watermarking embeds a hidden representation into the text during the text generation process. In simple terms, adding a watermark involves incorporating a selection strategy into the process of generating text. 

\textbf{Logits-based Watermark.} 
Logits-based watermarking methods do not modify model parameters; instead, they alter the logits distribution before text generation to produce detectable outputs. 
A representative approach is KGW~\cite{kirchenbauer2023watermark}, which  dynamically hash previous tokens to partition the vocabulary into green and red lists, sampling subsequent tokens exclusively from the green list. Their follow-up work~\cite{kirchenbauer2024on} introduced enhanced techniques such as SelfHash and WinMax to improve robustness.

To address the limitation that previous methods fail to detect watermark when these segments are embedded in longer documents, the WaterSeeker model \cite{yu2024waterseekerpioneeringefficient} adopts a "locate first, detect later" strategy to accurately locate and verify watermarked segments in large corpora.
To mitigate the degradation of text quality caused by watermarking, unbiased watermarking~\cite{hu2023unbiased} performs "reweighting" on the output distribution during sampling to modify the sampling probability without altering the overall output distribution of the model. 
Liu et al.~\cite{liu2024unforgeable} propose using two separate neural networks for generation and detection while sharing token embedding parameters, giving the detector prior knowledge. The asymmetry of computation further guarantees non-forgeability by preventing erasure or counterfeit watermark creation.

Red-green list–based methods may compromise text quality while maintaining detectability. To address this, adaptive watermarking techniques dynamically adjust watermark strategies rather than relying on fixed token mappings, balancing text quality and robustness. 
Liu et al.~\cite{liu2024adaptivetextwatermarklarge} introduce adaptive watermark token identification, semantic-based logits scaling, and adaptive temperature control. This methods selectively modify the distribution of high-entropy next tokens while leaving low-entropy tokens unchanged.
Building on this, Wang et al.~\cite{wang2025morphmarkflexibleadaptivewatermarking} propose MorphMark, which identifies factors affecting watermark effectiveness and text quality, and dynamically adjusts watermark intensity to optimize the trade-off between detectability and naturalness.

\textbf{Sampling-based Watermark.} 
Sampling-based watermarking methods embed watermarks by modifying the token sampling procedure rather than directly altering the logits distribution.

SynthID-Text~\cite{SynthIDText} is a generative text watermarking system based on tournament sampling that embeds statistical signals through the sampling process. It supports both distortion-free (quality-preserving) and distortion-enabled (detection-enhanced) modes, and innovatively combines watermarking with speculative sampling, achieving production-level efficiency.

SymMark~\cite{wang2025tradeoffsynergyversatilesymbiotic} further advances adaptability by integrating logit-based and sampling-based watermarking under Serial, Parallel, and Hybrid strategies, enabling the model to adaptively select the most suitable watermarking strategy for each token.
Duwak~\cite{zhu2024duwakdualwatermarkslarge} proposes a dual-watermark framework that extends logit-based watermarking with a quality-aware contrastive search sampling scheme, significantly improving text quality while enhancing output diversity.

In summary, generative sampling watermark embeds signals at generation time by manipulating output distributions or sampling methods. These methods achieve strong watermark robustness while preserving text quality and generation efficiency.

\subsubsection{Output Text Watermark}
\label{sec:model-prior-output}
Unlike watermarking methods embedded during generation, a output text watermarking injects watermarks into already generated text. Specifically, it modifies existing outputs at different granularities, such as word-level or sentence-level, without altering the original generation process.

\textbf{Word-level Watermark.} 
Yang et al.~\cite{yang2023watermarkingtextgeneratedblackbox} propose a traceability scheme based on dynamic binary coding. This method drives the proportion of bit-1 tokens in watermarked text to deviate significantly from randomness (exceeding 50\%) while preserving semantic fidelity through explicit semantic constraints. PostMark~\cite{Chang2024PostMarkAR} rewrites the text to incorporate the semantic watermark words. The posteriori semantic-aware watermarking framework~\cite{an2025defendingllmwatermarkingspoofing} further introduces a semantic mapping model that constructs green–red token lists based on the entire previously generated input text, enabling globally semantic-aware watermark embedding. 

\textbf{Sentence-level Watermark.} Moving beyond token-level manipulation, SemStamp~\cite{hou2024semstampsemanticwatermarkparaphrastic} presents a robust sentence-level semantic watermarking algorithm based on locality-sensitive hashing (LSH). It performs sentence-level rejection sampling until the paraphrased sentence falls into a predefined watermark region in the semantic embedding space. An improved variant~\cite{hou2024ksemstampclusteringbasedsemanticwatermark} replaces LSH with k-means clustering, yielding more stable and robust embedding space partitioning.

\textbf{Behavioral-level Watermark.} Agent Guide~\cite{yang2025agentguidesimple} embeds watermarks by decoupling high-level intentions from low-level actions. It injects probabilistic biases at the decision layer while preserving natural action realizations, enabling behavior-based provenance tracing.

In summary, a output text watermarking embeds traceable signals after text generation by modifying lexical, semantic, or behavioral representations. These methods are particularly well suited for black box settings, offering flexible provenance verification without requiring access to model parameters or the generation process.

%% file: sections/4_structure_sourcing.tex
\section{Sourcing from Model Structures}
\label{sec:structure}
\begin{figure}
    \centering
    \includegraphics[width=\linewidth]{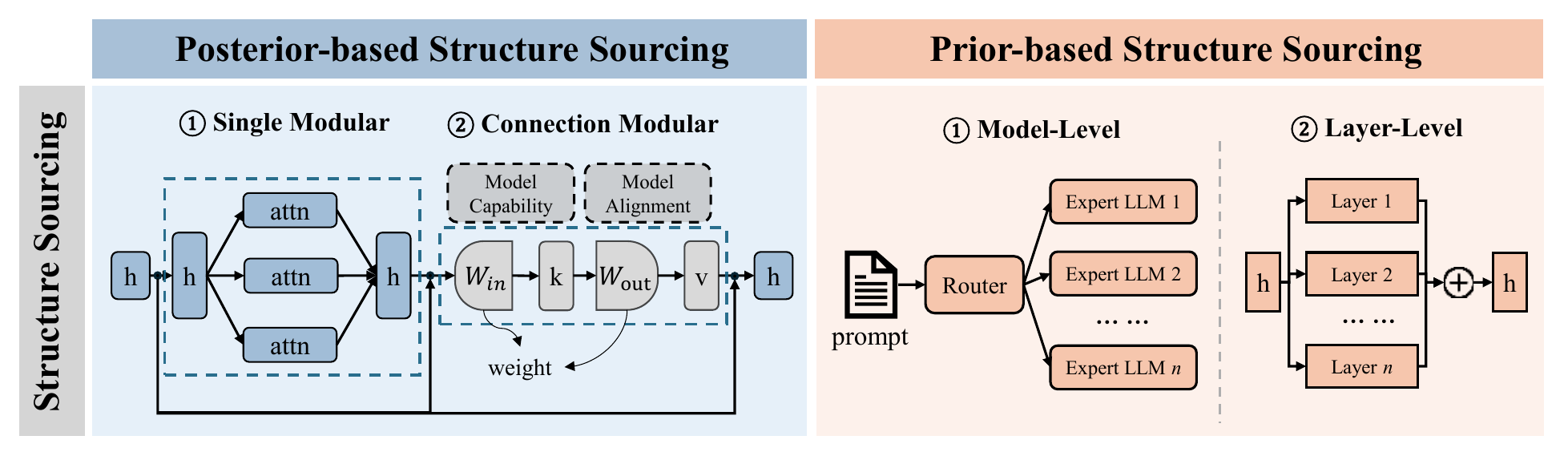}
    \caption{The methods for structural sourcing can be categorized into two types: posterior-based and prior-based.
Within posterior-based structural sourcing, there are two approaches: single-modular and connection-modular.
For prior-based structural sourcing, the approaches are divided into model-level and layer-level.}
    \label{fig:structure-sourcing}
\end{figure}

Sourcing from model structures in LLMs involves identifying and analyzing how different architectural components, such as specific layers and attention heads, contribute to generated outputs. This approach is instrumental in addressing the inherent opacity of LLMs. By mapping model behaviors back to granular architectural elements, including attention mechanisms, feed-forward networks (FFNs), or specific parameter clusters, this methodology facilitates precise model optimization and enables enhanced safety through targeted interventions. Furthermore, it establishes a rigorous technical framework for accountability and regulatory compliance in critical applications. 
As shown in Fig.~\ref{fig:structure-sourcing}, research in this area diverges into posterior-based and prior-based approaches. Posterior-based methods retrospectively analyze activation trajectories, parameter gradients, or behavioral fingerprints to infer the structural elements responsible for specific model responses. Conversely, prior-based methods proactively embed traceable markers into network layers or attention heads during the design phase to enable direct structural attribution.

\subsection{Datasets and Evaluations}
The inherent opacity of deep LLM architectures necessitates specialized benchmarks to attribute model outputs to internal mechanisms~\cite{grattafiori2024llama,yang2024qwen2}. To address this, the community has established rigorous protocols involving distinct datasets and hybrid evaluation metrics~\cite{pan2023unifying}.

\subsubsection{Datasets}

Benchmarks for sourcing can be categorized by their specific structural focus. Initial efforts concentrating on foundational knowledge include LAMA~\cite{petroni2019language}, which utilizes cloze-style probes for factual retrieval, and datasets like ZsRE~\cite{levy2017zero} and COUNTERFACT~\cite{meng2022locating_neurips}, which introduce counterfactuals to facilitate causal tracing to specific layers or heads. Addressing higher structural complexity, recent datasets target more intricate dependencies: MQuAKE~\cite{zhong2023mquake} tests coherent multi-hop retrieval, MLaKE~\cite{DBLP:conf/coling/WeiDPDSC25} examines cross-lingual fact alignment, and UnKEBench~\cite{deng2024everything} analyzes dense, entangled knowledge within unstructured text. Furthermore, to quantify the trade-offs involved in model updates, benchmarks such as StableKE~\cite{wei2024stable} and S$^2$RKE~\cite{duan2025related} focus on stability and interference, specifically assessing robustness against semantic perturbations and related-knowledge conflicts. Table~\ref{tab:knowledge_editing_benchmarks} summarizes these knowledge centric resources.

Complementing these knowledge-focused benchmarks, a parallel stream of datasets targets safety and interpretability to diagnose mechanism-level vulnerabilities. For instance, RealToxicityPrompts~\cite{gehman2020realtoxicityprompts} and JailbreakBench~\cite{chao2024jailbreakbench} are designed to stress-test toxicity levels and adversarial robustness. From a mechanistic perspective, TruthfulQA~\cite{lin2022truthfulqa} isolates the model's resistance to misinformation priors, while InterpBench~\cite{gupta2024interpbench} provides ground-truth algorithmic subgraphs, serving as a specialized unit testbed for evaluating circuit-level discovery algorithms.

\subsubsection{Evaluation Protocols}
Robust sourcing demands hybrid protocols that integrate behavioral verification with causal interrogation~\cite{meng2022locating_neurips,geva2022transformer}. Key evaluation dimensions include:
\begin{itemize}
\item \textbf{Edit Success and Locality.} Evaluate the effectiveness of edits through the Edit Success Rate (ESR) while simultaneously measuring preservation on non-edited facts and control locality (Specificity). Metrics typically include bounding non-target $\Delta$ accuracy or $\Delta$ logit following an edit~\cite{meng2022locating_neurips}.
\item \textbf{Compositional and Temporal Consistency.} In multi-hop reasoning or time-sensitive contexts, measure the consistency of information propagation across reasoning paths, robustness under paraphrases or templated variants, and stability in the face of temporal drift~\cite{zhong2023mquake, wei2024stable}.
\item \textbf{Cross-lingual Transfer.} For multilingual provenance traces, assess whether edits or localized components generalize across aligned factual chains in different languages, while avoiding collateral degradation in unrelated behaviors~\cite{DBLP:conf/coling/WeiDPDSC25}.
\item \textbf{Safety and Reliability.} Complement provenance testing with safety-oriented benchmarks, such as toxicity rates~\cite{luong2024realistic}, jailbreak resistance~\cite{zhou2025dont}, truthfulness~\cite{lin2022truthfulqa}, and refusal precision~\cite{cao2024learn}, ensuring that structural interventions do not compromise critical safeguards.
\item \textbf{Causal Validity.} Apply interventions such as activation patching, attention manipulation~\cite{geiger2021causal}, or weight perturbation to quantify causal effects on target logits. Report effect sizes, average treatment effect (ATE) deltas, and minimal-sufficiency criteria.
\item \textbf{Evidence Faithfulness.} When provenance relies on external evidence, approximate human evaluation via automated metrics: ALCE for citation precision using NLI~\cite{gao-etal-2023-enabling}; citation overlap against gold-standard sets~\cite{djeddal2024evaluation}; LLM-as-judge frameworks for flexible scoring~\cite{zhang2024towards}; and ALiiCE, which augments claim parsing with positional dispersion (CVCP) for fine-grained evaluation~\cite{xu2024aliice}.
\item \textbf{Circuit-Grounded Scoring.} On synthetic benchmarks such as \emph{InterpBench}, measure component discovery through subgraph recovery F1, edit distance, or success\@k for neuron naming and circuit reconstruction~\cite{conmy2023automated}. These criteria enable unit-test-style validation of structural claims.
\end{itemize}
This dual strategy—pairing scalable metrics with fine-grained causal attributions—ensures reproducibility while validating the structural fidelity of explanations.

\begin{table*}
\centering
\small
\caption{Summary of representative datasets for model structures sourcing.}
\label{tab:knowledge_editing_benchmarks}
\begin{tabular}{lccccc}
\toprule
\textbf{Dataset} & \textbf{Domain} & \textbf{Language(s)} & \textbf{Size} & \textbf{Year} & \textbf{Project Page} \\
\midrule
ZsRE & Wikipedia & EN & 11,301 & 2017 & \href{https://nlp.cs.washington.edu/zeroshot/}{Link} \\
LAMA & Wikidata & EN & 34,000 & 2019 & \href{https://github.com/facebookresearch/LAMA}{Link} \\ 
COUNTERFACT & Wikipedia & EN & 21,919 & 2022 & \href{https://rome.baulab.info}{Link} \\
MQuAKE & Wikipedia & EN & 11,043 & 2023 & \href{https://github.com/stanfordnlp/mquake}{Link} \\
KEBench & Wikidata & EN & 3,798 & 2024 & \href{https://github.com/Hi-archers/StableKE}{Link} \\
UnKE & Wikipedia & EN & 1,000 & 2024 & \href{https://github.com/TrustedLLM/UnKE}{Link} \\
MLaKE & Wikipedia & EN, ZH, JA, FR, DE & 9,432 & 2025 & \href{https://github.com/Hi-archers/MLaKE}{Link} \\
S$^2$RKE & YAGO & EN & 22,064 & 2025 & \href{https://github.com/Zhow01/S2RKE}{Link} \\
\bottomrule
\end{tabular}
\end{table*}

\subsection{Posterior-based Structural Sourcing}
\label{sec:structure-post}
Posterior-based structural sourcing refers to the process of tracing the structural origins of capabilities within a trained model. This field of research is broadly divided into two primary approaches~\cite{DBLP:conf/emnlp/WangYXQD00GJX0C24,DBLP:journals/corr/abs-2504-15133}.

The first approach is single modular function localization. This perspective posits that a model's capabilities and behavioral characteristics can be pinpointed to specific, relatively independent components or regions~\cite{geva2020transformer,meng2022locating_neurips,DBLP:conf/iclr/MengSABB23}.
The second approach is connection modular function localization. In contrast to the former, this viewpoint contends that a model's capabilities do not arise from isolated modules but instead manifest as systemic properties that emerge from the dynamic collaboration among multiple components~\cite{wei2024stable,deng2024everything}.

This approach proves valuable as it enables deeper understanding of how different model components contribute to various behaviors and capabilities. Furthermore, it facilitates targeted interventions and optimizations without requiring prior structural modifications, thereby enabling flexible post hoc analysis and selective manipulation of model behavior while preserving the original architecture. The categorization of these methods is summarized in Table~\ref{tab:Model_Tracing_Datasets}, providing a structured overview of existing approaches.

\subsubsection{Sourcing from Single Modules}
\label{sec:structure-post-single}
This approach localizes specific model behaviors to distinct structural components, categorizing findings into capability storage and alignment-critical regions.

\textbf{Model Capability Region.}
LLMs exhibit distinct functional partitions for diverse capabilities. Regarding semantic and factual capabilities, research indicates that Transformer Feed-Forward Networks (FFNs) function as key-value memories. Lower layers capture shallow patterns, while upper layers encode complex semantics~\cite{geva2020transformer}. Specifically, parameter vectors within FFNs act as ``concept promoters'', where updates to token representations directly influence vocabulary probability distributions~\cite{geva2022transformer}. Building on this mechanism, methods like LM-Debugger~\cite{geva2022lm} decompose layer updates to locate and intervene in concept generation. Furthermore, the identification of ``knowledge neurons'' has enabled precise factual updating. Techniques such as Rank-One Model Editing (ROME)~\cite{meng2022locating_neurips} and its batched extension, MEMIT~\cite{DBLP:conf/iclr/MengSABB23}, manipulate specific FFN weights to alter factual associations without retraining. In contrast, mathematical capabilities often rely on distinct mechanisms, specifically heuristic-driven neurons found in mid-to-late layers rather than robust algorithmic circuitry~\cite{nikankin2024arithmetic}.

\textbf{Model Alignment Region.}
Research on model alignment increasingly focuses on identifying specific structural sites responsible for adhering to human values. These critical regions are broadly categorized into two domains: security alignment, which pertains to robustness against adversarial exploits and toxicity, and behavioral alignment, which governs attributes such as ethical reasoning, truthfulness, and output coherence. Isolating these components is essential for implementing targeted safety interventions without compromising the model's general performance.

For the security alignment region, unsafe knowledge and privacy vulnerabilities are frequently localized within specific FFNs and early model layers.
Studies utilizing Training Data Extraction (TDE) and neuron contribution analysis have traced Personally Identifiable Information (PII) and privacy leakages to high-contributing units in FFNs~\cite{venditti2024enhancing,wu-etal-2023-depn}. To mitigate these risks, gradient-based localization methods detect ``unsafe'' parameter regions associated with bias and copyright infringement~\cite{tian2024forget,yu2023unlearning,hazra-etal-2024-safety}, enabling targeted activation suppression. Furthermore, permanent detoxification is achieved by editing unsafe parameters identified via adversarial queries~\cite{wang-etal-2024-detoxifying} or by removing global toxic subspaces from FFNs to preserve general capabilities~\cite{duan2025gloss}. Additionally, realigning early layers has proven effective in reducing vulnerability to jailbreak attacks~\cite{zhao-etal-2024-defending-large}.

For the behavioral alignment region, parameter-level analysis extends to the governance of model personas and factual reliability. In the context of personality traits and sycophancy, specific neurons within FFNs and attention heads have been identified as critical drivers. For instance, NPTI~\cite{deng2024neuron} isolates trait-specific neurons, while SPT~\cite{DBLP:conf/icml/ChenH0LL000Z0SY24} demonstrates that fine-tuning fewer than 5\% of parameters in these targeted regions effectively mitigates undue agreement. Concerning hallucinations, problematic regions are frequently traced to decoder-layer parameters responsible for generating unverifiable content. Methodologies such as PFME~\cite{deng2024pfme} and TruthX~\cite{zhang2024truthx} localize these areas through entity-based discrepancy analysis and truthful latent space mapping, respectively, enabling targeted corrections that significantly enhance factual consistency.

In summary, single-module sourcing methods attribute model behaviors to localized structural regions, revealing where capabilities and alignment signals concentrate. By enabling low-interference interventions on specific FFN or attention components, these approaches support fine-grained provenance, improved factuality, and more targeted safety control.

\subsubsection{Sourcing from Connection Modules}
\label{sec:structure-post-modular}
In contrast to component isolation, a systemic perspective posits that model capabilities emerge from the dynamic cooperation among specialized modules~\cite{DBLP:journals/corr/abs-2503-05788,DBLP:journals/corr/abs-2506-11135}. This view emphasizes that while individual components perform simple computations, their coordinated interaction, involving Attention, FFNs, and Residual Connections, yields complex behaviors~\cite{lindsey2025biology,DBLP:journals/corr/abs-2505-22311}.

\textbf{Dynamic Knowledge Routing.}
Factual processing involves a complex, multi-stage flow rather than static retrieval. Research delineates a collaborative mechanism: early FFNs construct semantic representations; mid-layer attention mechanisms capture relational cues; and deeper layers extract target objects based on these relations~\cite{geva2020transformer,deng2024everything,DBLP:conf/emnlp/GevaBFG23}. Further granular analysis reveals that specific attention heads route subject entities to critical network regions, where they are activated by mid-layer components and subsequently mapped to complex relations by deep FFN layers~\cite{DBLP:journals/corr/abs-2403-19521}. This underscores the synergistic roles of attention in information routing and FFNs in relational transformation.

\textbf{Knowledge Circuits.}
The ``knowledge circuits'' framework formalizes these interactions by identifying computational subgraphs responsible for specific functions~\cite{olah2020zoom,elhage2021mathematical}. In factual tasks, circuits composed of ``information-gathering heads'', ``relation-rendering heads'', and FFN neurons collectively encode knowledge. Interventions within these pathways verify their causal role in storing facts~\cite{DBLP:conf/nips/Yao0XWXDC24}. This approach has successfully mapped circuits for diverse tasks, including color identification~\cite{merullocircuit} and geographical query resolution~\cite{DBLP:conf/iclr/WangVCSS23}, confirming that capabilities are distributed across function-specific paths.

\textbf{Hierarchical Reasoning.}
Complex cognitive capabilities, such as multi-step reasoning, emerge from hierarchical inter-module collaboration~\cite{prystawski2023think}. While single modules lack independent reasoning algorithms, their interplay approximates sophisticated logic. Lan et al.~\cite{lan2024towards} identified shared sub-circuits in mid-to-late layers specialized for sequence continuation. This collaboration is stratified: shallow layers parse patterns, middle layers construct relational mappings, and deep layers execute logical integration.

\textbf{Emergent Algorithms.}
Crucially, internal component collaboration can simulate classical algorithms, explaining in-context learning (ICL) and generalization. Theoretically, the transformation by a single linear attention layer is mathematically equivalent to a gradient descent step on a regression loss; stacking these layers iteratively refines predictions, simulating—and potentially outperforming—gradient descent by dynamically adjusting learning rates~\cite{DBLP:conf/iclr/AkyurekSA0Z23,von2023transformers,fu2024transformers}. Beyond optimization, components can collaborate to simulate dynamic programming algorithms for tasks like edit-distance calculation, with specific heads passing cost information and FFNs updating minimum costs~\cite{DBLP:conf/iclr/ToddLSMWB24}. These findings suggest that models learn to execute general-purpose algorithms through the flexible coordination of their vast parameter spaces.

In summary, connection-module sourcing highlights that model capabilities emerge from coordinated interactions across layers. By framing behaviors as outcomes of system-level cooperation and analyzable circuits, this line of work explains complex reasoning and algorithmic generalization, offering a pathway toward global structural provenance.

\begin{table*}[t]
    \centering
    \caption{Representative posterior-based structural sourcing studies.}
    \label{tab:Model_Tracing_Datasets}
    \resizebox{0.9 \textwidth}{!}{%
    \begin{tabular}{ccccp{3.25cm}}
        \toprule
        Setting & Region & Method & Year & Target Models \\
        \midrule
        \multirow{4}{*}{Capability} 
          & Semantic grammar & Geva et al.~\cite{geva2020transformer} & 2020 & Transformer-XL \\
          & Arithmetic       & Nikankin et al.~\cite{nikankin2024arithmetic} & 2024 & LLaMA-2-13B \\
          & Creative writing & Gómez et al.~\cite{gomez2023confederacy} & 2023 & GPT-4, Claude, PaLM-2 \\
          & Hardware design creativity & CreativEval \cite{delorenzo2024creativeval} & 2024 & GPT-4, Claude \\
        \midrule
        \multirow{4}{*}{Alignment} 
          & Security (PII)           & PEA \cite{venditti2024enhancing} & 2024 & LLaMA-2-7B \\
          & Security (privacy \& IP) & MemFlex \cite{tian2024forget}    & 2024 & GPT-J, GPT-2 \\
          & Personality              & NPTI \cite{deng2024neuron}      & 2024 & LLaMA-2-7B \\
          & Sycophancy               & SPT \cite{DBLP:conf/icml/ChenH0LL000Z0SY24} & 2024 & GPT-2 XL \\
        \midrule
        \multirow{4}{*}{Synergistic} 
          & Knowledge flow           & Geva et al.~\cite{DBLP:conf/emnlp/GevaBFG23} & 2023 & GPT-2 M \\
          & Knowledge circuits       & Yao et al.~\cite{DBLP:conf/nips/Yao0XWXDC24} & 2024 & LLaMA-2-13B \\
          & In-context learning      & Akyürek et al.~\cite{DBLP:conf/iclr/AkyurekSA0Z23} & 2023 & GPT-2 S \\
          & Edit-distance DP         & Todd et al.~\cite{DBLP:conf/iclr/ToddLSMWB24} & 2024 & GPT-2 XL \\
        \bottomrule
    \end{tabular}
    }
\end{table*}

\subsection{Prior-based Structural Sourcing}
\label{sec:structure-prior}
In contrast to post-hoc analysis of emergent capabilities, prior-based structural sourcing imposes explicit architectural modularity, typically via Explicit Domain-Expert Mixture-of-Experts (MoE)~\cite{shazeer2017outrageously}. By employing a learned router to dispatch inputs to specialized sub-modules, this design facilitates sourcing, as outputs can be directly attributed to the activated experts. This proactive approach ensures interpretability and traceability by design, avoiding the opacity of dense models~\cite{fedus2021switch}.

\textbf{Model-level Expert Specialization.} 
\label{sec:structure-prior-model}
Early implementations focused on coarse-grained, task-level routing. MoDEM~\cite{Simonds2024MoDEM} employs a BERT-based router to assign natural language tasks to domain-specific experts (e.g., health, mathematics), outperforming generalist baselines. This paradigm extends to multimodal and recommendation domains: Med-MoE~\cite{Jiang2024MedMoE} routes visual features based on diagnostic report categories to enhance medical reasoning, while MoE-MLoRA~\cite{Yaggel2025MoEMLoRA} leverages lightweight LoRA adapters as experts for click-through-rate prediction, demonstrating that performance gains rely on optimizing the interplay between expert count and data characteristics.

\textbf{Layer-level Expert Specialization.} 
\label{sec:structure-prior-layer}
Recent advancements embed specialization within individual layers to enhance granularity and efficiency. TASER~\cite{Cheng2023TASER} interleaves shared and expert FFN layers via a task-aware router, achieving superior retrieval robustness with reduced parameter counts. To address redundancy, DeepSeekMoE~\cite{deepseekmoe2024} introduces fine-grained expert segmentation combined with dedicated shared experts, maintaining high performance with lower computational costs. DynMoE~\cite{dynmoe2024} further increases flexibility through dynamic gating, allowing the model to auto-determine the number of active experts per token. Additionally, MoEUT~\cite{moeut2024} extends this concept to Universal Transformers by integrating experts into both FFN and attention sublayers with parameter sharing, optimizing memory usage while surpassing standard transformer benchmarks.

Collectively, these architectures represent a proactive shift toward inherently traceable models, decomposing complex behaviors into verifiable expert contributions and establishing a technical foundation for accountable AI deployment.

%% file: sections/5_data_sourcing.tex
\section{Sourcing from Training Data}
\label{sec:training-data}

\begin{figure}
    \centering
    \includegraphics[width=\linewidth]{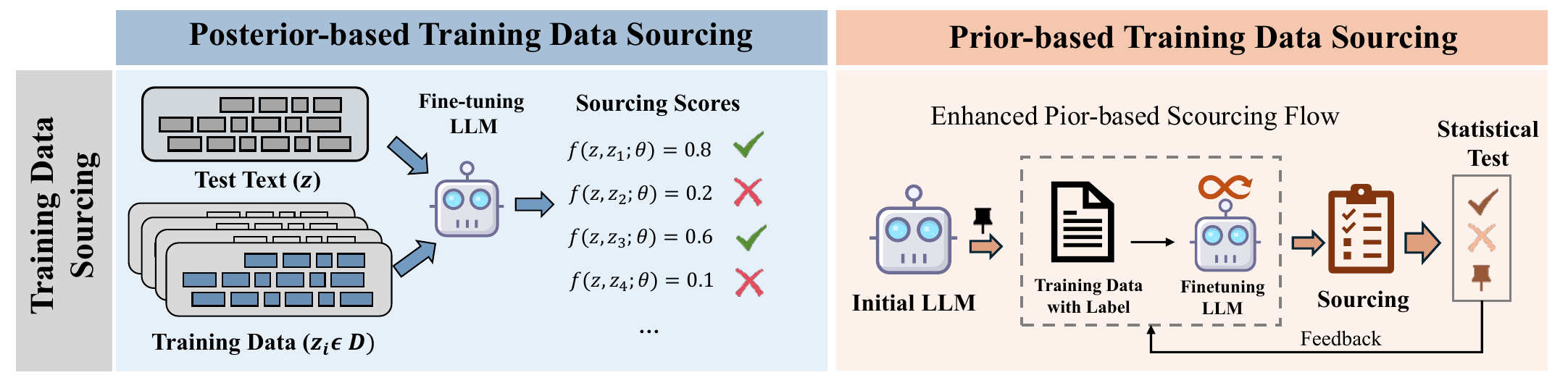}
    \caption{Training data sourcing methods are categorized into two types: posterior-based and prior-based. Posterior-based training data sourcing includes two approaches: white box and black box. Prior-based training data sourcing is divided into watermark-based and proxy model methods.}
    \label{fig:training_data_method}
\end{figure}

Training data sourcing addresses the task of causal attribution by linking generated outputs to specific training samples, a process vital for mitigating risks associated with biased, sensitive, or noisy datasets. This dimension formalizes the relationship between data provenance and model behavior to advance solutions in fairness, privacy, and regulatory compliance. 
Methodologies are categorized by their implementation stage as shown in Fig.~\ref{fig:training_data_method}: posterior-based methods employ memorization detection, influence functions, and gradient analysis to infer data origins from the trained model, while prior-based techniques insert verifiable fingerprints or metadata into the corpus during pre-training for deterministic traceability.

\subsection{Datasets and Evaluations}
The systematic evaluation of training data sourcing methodologies relies on specialized benchmarks and rigorous metrics. This subsection details the datasets and evaluation protocols essential for assessing the efficacy of both prior-based and posterior-based sourcing approaches. 

\subsubsection{Datasets}

Empirical evaluation of LLM training data sourcing methods requires domain-specific benchmarks aligned with the differing assumptions of prior- and posterior-based frameworks. Table~\ref{tab:_training_data_sourcing_datasets} summarizes commonly used datasets.

For posterior-based methods, which infer training data influence from post hoc model behavior without prior intervention, dedicated datasets typically support evaluation via approximate counterfactual influence. Fact-tracing resources such as FTRACE~\cite{akyurek2022towards} and GPTDynamics~\cite{chai2024training} assess provenance of training instances providing semantically or lexically aligned evidence for model-generated factual claims. Hallucination-tracing datasets like Hallucinatory XSum~\cite{wu2024enhancing} and BookSum~\cite{wang2023source} evaluate the ability to trace training samples that may drive hallucinatory outputs.

Prior-based methods, which rely on pre-embedded signals in the training corpus, require benchmarks with explicit markers. Datasets such as WIKIMIA~\cite{shi2023detecting}, BookMIA~\cite{panaitescu2025can}, and BIOCITE~\cite{khalifa2024source} provide syntactic watermarks, lexical signatures, or metadata for deterministic tracing in controlled settings where marker structure and location are known a priori. Naturally occurring datasets are unsuitable unless augmented or partitioned to supply ground-truth traceability.

\subsubsection{Metrics}
The evaluation of training data sourcing methods uses metrics aligned with the goals of posterior- and prior-based paradigms.

Posterior-based methods typically rely on leave-one-out (LOO)–style evaluation, measuring performance changes when a training instance is removed~\cite{koh2017understanding, li2024influence}. Because exact LOO is infeasible for LLMs, evaluations often rely on rank correlation metrics, such as Pearson or Spearman coefficients, to compare estimated influence scores against approximated LOO values derived from proxy models or subsampled data. Generative tasks use task-specific metrics: fact tracing uses retrieval measures such as Precision@K or mean reciprocal rank~\cite{meng2022locating_neurips}, while hallucination tracing combines lexical or topical similarity with human judgments of causal plausibility~\cite{akyurek2022towards}.

Prior-based methods emphasize three dimensions~\cite{kirchenbauer2023watermark}: detection fidelity (precision, recall, F1 score) for identifying whether a generated sequence contains watermarks attributable to a predefined source subset; robustness, by subjecting watermarked outputs to adversarial perturbations such as paraphrasing, truncation, or targeted watermark removal attacks, measuring the resilience of the embedded signals~\cite{christ2024undetectable}; and downstream utility, monitored via perplexity or task accuracy to ensure watermarking does not degrade model performance~\cite{zhao2023provable}.

\begin{table*}
\centering
\small
\caption{Summary of representative datasets for training data sourcing.}
\label{tab:_training_data_sourcing_datasets}
\begin{tabular}{lccccc}
\toprule
\textbf{Dataset} & \textbf{Domain} & \textbf{Language(s)} & \textbf{Training Set} & \textbf{Year} & \textbf{Project Page} \\
\midrule
FTRACE~\cite{akyurek2022towards} & News & EN & $\checkmark$ & 2023 & \href{https://huggingface.co/datasets/ekinakyurek/ftrace}{Link} \\
WIKIMIA~\cite{shi2023detecting} & Wikipedia & EN & $\checkmark$ & 2023 & \href{https://huggingface.co/datasets/swj0419/WikiMIA}{Link} \\
GPTDynamics~\cite{chai2024training} & News & EN & $\checkmark$ & 2024 & \href{https://github.com/ernie-research/gptfluence}{Link} \\
Hallucinatory Xsum~\cite{wu2024enhancing} & News & EN & $\checkmark$ & 2024 & \href{https://huggingface.co/datasets/shalinik/xsum}{Link} \\
BookMIA~\cite{panaitescu2025can} & Book & EN & $\checkmark$ & 2024 & \href{https://huggingface.co/datasets/swj0419/BookMIA}{Link} \\
BIOCITE~\cite{khalifa2024source} & Biography & EN & $\checkmark$ & 2024 & \href{https://github.com/mukhal/intrinsic-source-citation}{Link} \\
BookSum ~\cite{wang2023source} & Book & EN & $\checkmark$ & 2025 & \href{https://huggingface.co/datasets/kmfoda/booksum}{Link} \\
\bottomrule
\end{tabular}
\end{table*}

\subsection{Posterior-based Training Data Sourcing}
\label{sec:training-data-post}
Posterior-based training data sourcing refers to techniques that identify training instances critically influencing a model’s predictions by examining its output behavior in response to specific inputs (Fig.~\ref{fig:training_data_method}). Their fundamental assumption is that the model’s posterior behavior inherently reflects the imprint of its training data.

\subsubsection{White Box Methods}
\label{sec:training-data-post-white}
White box posterior-based training data sourcing methods refer to computational approaches that analyze training data’s influence on model outputs, with access to model parameters or internal representations. These methods are broadly categorized into gradient-based and empirical influence score estimation.

\textbf{Gradient-based Influence Score Estimation.} This category of methods measures the contribution of training samples to model predictions using gradient information. The core operation involves computing the gradients of the loss function with respect to model parameters for both training and test samples~\cite{koh2017understanding}, then assessing influence scores based on the similarity or alignment between these gradients. Intuitively, a training sample whose gradient direction aligns closely with that of a test sample is considered to have a greater influence on the model’s prediction. This approach captures complex nonlinear dependencies within the model and is particularly suitable for analyzing gradient flow in deep neural networks. Influence functions is effective in identifying the most influential training samples, especially in detecting outliers or noisy data points. However, traditional influence function methods require computing the inverse of the Hessian matrix, which becomes computationally prohibitive and challenging to implement in high-dimensional parameter spaces.

To address this, several variants have been proposed, such as TracIN~\cite{pruthi2020estimating} which approximates influence using first-order gradients and model checkpoints, RelatIF~\cite{barshan2020relatifidentifyingexplanatorytraining} which normalizes influence by self-influence, FastIF~\cite{guo2021fastifscalableinfluencefunctions} which uses k-nearest neighbors for scalability, and Scaling-IF~\cite{schioppa2021scalinginfluencefunctions} which employs eigenvalue decomposition for efficient computation. These methods significantly reduce computational overhead while scaling to large datasets and complex architectures. In NLP tasks, empirical studies compare gradient-based methods~\cite{pezeshkpour2021empirical}, including influence functions and their variants, with similarity-based retrieval methods in terms of efficiency and complexity when applied to BERT. While gradient-based methods are theoretically robust, they remain computationally expensive in practice.
To mitigate this, a reranking strategy~\cite{akyurek2022towards} from information retrieval is used, where a candidate set of potentially influential samples is first retrieved, after which finer-grained attribution is applied, substantially reducing computation.

However, the approximation theory of influence functions relies on the assumption that model parameters attain empirical risk minimization, a condition often violated in language models due to their architectural complexity and diverse data distributions~\cite{bae2022if}. Furthermore, challenges such as data imbalance, noise, and limited resources exacerbate the difficulty of achieving empirical risk minimization~\cite{nguyen2023a}. Consequently, subsequent work incorporates decomposition-based approximation strategies to adapt influence functions for LLM training data attribution, including contrastive methods like CEA~\cite{ladhak-etal-2023-contrastive}, dynamic approaches like DDA~\cite{wu2024enhancing}, and extensions such as DataInf~\cite{kwon2024datainfefficientlyestimatingdata} and RLHF-IF~\cite{min2025understanding}.

As shown in Table~\ref{tab:training_data_sourcing_if}, the evolution of influence function formulas across different training data sourcing methods is summarized.

\newcolumntype{L}[1]{>{\raggedright\arraybackslash}p{#1}}
\newcolumntype{C}[1]{>{\centering\arraybackslash}p{#1}}

\begin{table*}[ht]
\centering
\caption{Evolution of influence function formulas in training data sourcing.}
\label{tab:training_data_sourcing_if}
\renewcommand{\arraystretch}{1.0} 
\begin{adjustbox}{max totalsize={\textwidth}{\textheight},center}
\begin{tabular}{L{2.2cm} C{7.0cm} L{7.8cm}}
\toprule
\textbf{Method} & \textbf{Formula} & \textbf{Mathematical Interpretation} \\
\midrule
IF \cite{koh2017understanding} & 
$\displaystyle
I(z, z_{\text{query}}) = -\nabla_\theta L(z_{\text{query}}, \hat{\theta})^\top H^{-1} \nabla_\theta L(z, \hat{\theta})
$ &
\begin{minipage}{\linewidth}
%\centering
$\nabla_\theta$ = gradient w.r.t. parameters $\theta$\\ 
$H^{-1}$ = inverse Hessian matrix of empirical loss \\
$\hat{\theta}$ = optimized model parameters
\end{minipage} \\
\midrule
TracIn \cite{pruthi2020estimating} & 
$\displaystyle
I(z, z_{\text{query}}) = \sum_{k=1}^K \nabla_\theta L(z_{\text{query}}, \theta_{t(k)})^\top \nabla_\theta L(z, \theta_{t(k)})
$ &
\begin{minipage}{\linewidth}
%\centering
$\theta_{t(k)}$ = parameters at checkpoint $t(k)$ \\
$K$ = number of selected checkpoints
\end{minipage} \\
\midrule
RelatIF \cite{barshan2020relatifidentifyingexplanatorytraining} & 
$\displaystyle
I_{\text{rel}}(z, z_{\text{query}}) = \frac{I(z, z_{\text{query}})}{I(z, z)}
$ &
\begin{minipage}{\linewidth}
%\centering
$I(z, z)$ = self-influence of training point $z$
\end{minipage} \\
\midrule
CEA \cite{ladhak-etal-2023-contrastive} & 
$\displaystyle
I_{\text{CEA}}(z, z_{\text{query}}) = I(z, \tilde{z}_{\text{query}}) - I(z, z_{\text{query}})
$ &
\begin{minipage}{\linewidth}
%\centering
$\tilde{z}_{\text{query}}$ = human-corrected version of $z_{\text{query}}$
\end{minipage} \\
\midrule
DDA \cite{wu2024enhancing} & 
$\displaystyle
I_{\text{DDA}}(z, z_{\text{query}}) = I_{\hat{\theta}}(z, z_{\text{query}}) - I_{\theta_{0}}(z, z_{\text{query}})
$ &
\begin{minipage}{\linewidth}
%\centering
$\hat{\theta}$ = model parameters after training \\
$\theta_{0}$ = model parameters before training
\end{minipage} \\
\midrule
FastIF \cite{guo2021fastifscalableinfluencefunctions} & 
$\displaystyle
I_{\text{FastIF}}(z, z_{\text{query}}) = \text{kNN}(z_{\text{query}}) \cdot \nabla_\theta L(z, \hat{\theta})
$ &
\begin{minipage}{\linewidth}
%\centering
$\text{kNN}(z_{\text{query}})$ = k-nearest neighbors of $z_{\text{query}}$
\end{minipage} \\
\midrule
Scaling-IF \cite{schioppa2021scalinginfluencefunctions} & 
$\displaystyle
I_{\text{Scaling-IF}} = \sum_{i=1}^{\tilde{p}} \lambda_i^{-1} \langle g_i, \nabla_\theta L(z_{\text{query}}) \rangle \langle g_i, \nabla_\theta L(z) \rangle
$ &
\begin{minipage}{\linewidth}
%\centering
$\lambda_i$ = $i$-th dominant eigenvalue of $H$ \\
$g_i$ = corresponding eigenvector \\
$\tilde{p}$ = number of retained eigenvalues
\end{minipage} \\
\midrule
DataInf \cite{kwon2024datainfefficientlyestimatingdata} & 
$\displaystyle
I_{\text{DataInf}} = \frac{1}{n\lambda} \sum_{d \in D} \left( I - \frac{v_d v_d^\top}{\lambda + v_d^\top v_d} \right)
$ &
\begin{minipage}{\linewidth}
%\centering
$v_d = \nabla_\theta \ell_{\text{pref}}(d; \theta)$ = gradient of preference loss \\
$\lambda$ = damping parameter \\
$n$ = dataset size
\end{minipage} \\
\midrule
RLHF \cite{min2025understanding} & 
$\displaystyle
I_{\text{RLHF-IF}}(d_i) = -\nabla_\theta L(D_{\text{val}}; \theta)^\top H_{\text{pref}}^{-1} \nabla_\theta \ell_{\text{pref}}(d_i; \theta)
$ &
\begin{minipage}{\linewidth}
%\centering
$H_{\text{pref}}$ = Hessian of Bradley-Terry preference loss \\
$\ell_{\text{pref}}$ = cross-entropy preference loss \\
$D_{\text{val}}$ = validation preference dataset
\end{minipage} \\
\bottomrule
\end{tabular}
\end{adjustbox}
\end{table*}

\textbf{Empirical-based Influence Score Estimation.} This category of methods directly quantifies the impact of training samples through controlled experiments, often involving multiple rounds of model retraining, each time excluding specific samples~\cite{chai2024training} or groups of samples~\cite{koh2019accuracy}, and measuring the resulting change in model predictions~\cite{pezeshkpour2021empirical}. Since this approach does not rely on assumptions about the model’s internal mechanisms, it can accurately capture complex interactions and dependencies among samples. However, it is computationally intensive and generally infeasible for LLMs. To improve practicality, approximations such as proxy models~\cite{guu2023simfluence}, group-level removal, and influence propagation methods are commonly employed.

\subsubsection{Black Box Methods}
\label{sec:training-data-post-black}

Black box methods provide a powerful paradigm for training data attribution, functioning without access to model internals. They primarily use retrieval strategies—from vector-space similarity to neighborhood-space analysis—to identify the most influential training samples for a given model prediction.

\textbf{Retrieval-based Data Attribution.}
This approach to retrieval-based data attribution relies on measuring the similarity between vector representations of training samples and test examples. In this method, both training and test data are typically embedded into a high-dimensional vector space~\cite{chen2024fasttrackfastaccuratefact}, often using internal representations from the model itself or pre-trained embedding models. The underlying assumption is that the training samples most similar to a test example in this vector space had the greatest influence on the model’s prediction for that test case~\cite{hanawa2021evaluationsimilaritybasedexplanations,pezeshkpour2021empirical}. Common similarity measures include cosine similarity, Euclidean distance, and dot product~\cite{hanawa2021evaluationsimilaritybasedexplanations, yang2025gmvaluatorsimilaritybaseddatavaluation}. The process generally involves computing similarity scores between the test example and all training samples, then ranking them to retrieve the top-K most similar instances. This approach is particularly well-suited to models that operate on vector representations, such as deep neural networks for text or image processing. A key advantage is its scalability, as efficient approximate nearest neighbor search algorithms can handle large datasets. However, the method’s effectiveness is highly dependent on the quality and relevance of the vector representations~\cite{zhao2019inferringtrainingdataattributes}. It may fail to capture complex, non-linear relationships learned by the model, particularly if the embedding space does not align well with the model’s decision boundaries.

\textbf{Neighbor Space Retrieval Method.}
This method diverges from conventional vector similarity approaches by examining the local neighborhood of test examples within the model's decision space. This technique identifies training samples that exert comparable influence on the model's decision-making process as the test case, estimating how predictions would shift if specific training samples were removed or perturbed~\cite{fotouhi2025fasttrainingdatasetattribution}. Such analysis captures non-linear relationships and complex interactions invisible in simple vector spaces, revealing how different training data regions shape the model's decision boundary. Recent advances in semi-parametric language modeling, such as Nearest Neighbor Speculative Decoding (NEST), demonstrate how token-level retrieval and dynamic span selection can enhance attribution by incorporating real-world text spans while maintaining fluency~\cite{li2024nearest}. Similarly, investigations into kNN retrieval augmentation reveal its superior generalization capabilities over vanilla language models, particularly when handling over-specified training data where causal relationships are obscured by irrelevant information ~\cite{chiang2023retrieval}.
Despite its nuanced insights, this method faces computational challenges with large-scale models and datasets due to iterative subset processing. 

\subsection{Prior-based Training Data Sourcing}
\label{sec:training-data-prior}
Prior-based training data sourcing refers to a class of techniques that proactively design or modify training data prior to model training, with the goal of enabling subsequent traceability or attribution of model outputs to their source data, as shown in the Fig.~\ref{fig:training_data_method}. The central principle is to embed traceable prior signals, such as watermarks, or to construct auxiliary models, such as proxy influence estimators, during the training phase. These interventions ensure that the model’s behavior at inference time reflects the provenance or composition of its training data, thereby supporting efficient and controllable data attribution analysis.

\subsubsection{Watermark-based Methods}
\label{sec:training-data-prior-watermark}
Embedding unique watermark information in the training corpus is a technique used to achieve training data traceability, aiming to ensure that the model can learn and reflect this watermark information. The WASA framework~\cite{wang2023wasa} provides an effective solution for content traceability and data provenance by training LLMs to learn the mapping between the text from different data providers and their unique watermarks. Additionally, the research of Meta on the radiology of watermarked texts reveals the detectability of watermarked training data, indicating that even embedding a small amount of watermarked data can make the model output highly recognizable~\cite{sander2024watermarking}. Recent advancements have introduced novel approaches to enhance watermark robustness. For instance, Cui et al.~\cite{cui2025robust} propose injecting fictitious knowledge into training data, creating watermarks that are linguistically plausible yet semantically unique, making them resistant to preprocessing filters and effective across model lifecycle stages. Similarly, Rastogi et al.~\cite{rastogi2025stamp} introduce STAMP, a framework that generates watermarked rephrasings of content to prove dataset membership via statistical tests, enabling reliable detection even in API-only models.

However, watermarking technology faces several challenges and limitations~\cite{sadasivan2023can}. Firstly, watermarks are vulnerable to attacks or removal, significantly diminishing their effectiveness. Current techniques can effectively remove or alter watermarks while preserving the original data characteristics~\cite{pang2024attacking}, thereby compromising their content traceability. Moreover, applying watermarking frameworks to pre-trained models is challenging because they require specific interventions during the training process~\cite{kirchenbauer2024on}. This requirement limits the applicability of watermarking technology to existing models, especially when training data or the training process is inaccessible. Lastly, embedding watermarks may adversely affect the quality of generated text, as the process can introduce noise or other disturbances, reducing the naturalness and readability of the text~\cite{piet2023mark}. To address these issues, recent studies emphasize stealthy integration and statistical validation. For example, fictitious knowledge watermarks~\cite{cui2025robust} align with natural data distributions to evade detection, while STAMP~\cite{rastogi2025stamp} leverages paired statistical tests to minimize false positives and preserve content utility. These approaches highlight the evolving strategies to balance traceability, robustness, and practicality in watermarking systems.

\subsubsection{Proxy Model Methods}
\label{sec:training-data-prior-proxy}
Proxy model methods are designed to approximate training data influence through surrogate modeling mechanisms. These approaches aim to estimate or extrapolate influence patterns indirectly by learning predictive models of data influence or by exploiting the transferable properties of influence signals across different model scales.

\textbf{Data Influence Modeling.}
This approach involves building a separate model to predict influence scores based on various features of the training samples and their relationship to the test example. It treats influence estimation as a machine learning problem in its own right. The model is trained on a subset of the data where true influence scores are computed using more expensive methods like empirical estimation~\cite{liu2025attribotbagtricksefficiently}. It then learns to generalize and predict influence scores for the entire dataset~\cite{ilyas2022datamodelspredictingpredictionstraining}. Features for this model might include similarity measures, gradient information, model confidence, and various statistical properties of the samples.

Recent advances have refined this paradigm through model-aware frameworks. DsDm~\cite{engstrom2024dsdm} formulates dataset selection as an optimization problem, using datamodels to efficiently approximate the mapping between training subsets and model performance. By minimizing estimated target loss, it selects data subsets that demonstrably improve language model performance across diverse tasks. Similarly, Group-MATES~\cite{yu2025data} introduces a relational data influence model that captures interactions between data points through relationship weights. This group-level approach, trained via sampled trajectories and optimized with influence-aware clustering, significantly enhances pre-training efficiency by modeling complex dependencies beyond individual contributions.

These methods can leverage strengths of other approaches while remaining efficient once the influence model is trained. They also allow flexible influence definitions via domain-informed feature engineering. However, their effectiveness depends on the quality and representativeness of the training subset and may be limited in capturing complex or rare influence patterns.

\textbf{Proxy-Based Scalability Modeling.}
This approach builds on the insight that influence signals can be efficiently modeled through smaller surrogate systems. Yu et al.~\cite{yu2024mates} introduce the MATES framework, which employs a small data influence model to continuously track the evolving data preferences of a pre-training model and dynamically select the most effective training data. By fine-tuning a lightweight proxy model to approximate locally probed oracle influences, MATES enables model-aware data selection that significantly reduces computational costs while maintaining performance fidelity. Chang et al.~\cite{chang2024scalable} show that gradient-based influence methods, properly adapted, can retrieve influential training examples in an 8B-parameter LLM over 160B tokens. Subsequent work has further shown that influence estimates computed from smaller proxy models are highly correlated with those derived from larger target models~\cite{khaddaj2025smalltolargegeneralizationdatainfluences}, validating the transferable nature of influence signals across scales. Together, these findings demonstrate that proxy-based modeling provides a scalable and efficient path toward large-scale training data attribution, although its accuracy ultimately depends on the strength and stability of the correlation between proxy and target model behaviors across architectures and training configurations.

%% file: sections/6_citation_sourcing.tex
\section{Sourcing from External Data}
\label{sec:external-data}
External data sourcing in LLMs refers to the process of tracing, verifying, and attributing the external knowledge that supports the model’s generated outputs. Unlike training data sourcing, which targets the internal corpus used during pretraining and fine-tuning, external data sourcing focuses on the information that models actively retrieve, cite, or align with at inference time. This task is also commonly referred to as LLM Attribution or LLM Citation. This capability is especially important in applications such as knowledge-intensive question answering and scientific writing, and becomes particularly critical in high-stakes domains like law and medicine, where responses are expected to be supported by verifiable evidence. This task is crucial for ensuring transparency, factual consistency, and reliability of model responses, as it directly addresses whether the generated content can be grounded in verifiable evidence. 
To achieve this, as shown in Fig.~\ref{fig:citation_method}, posterior-based methods utilize latent space alignment and attention flow tracing to conduct retrospective analyses of input-output linkages. In contrast, prior-based frameworks integrate verifiable markers or source-specific identifiers into external knowledge bases during system deployment.

\begin{table*}[ht]
\centering
\small
\caption{Summary of representative datasets for external data sourcing.}
\label{tab:_external_data_sourcing_datasets}
\begin{tabular}{lccccc}
\toprule
\textbf{Dataset} & \textbf{Domain} & \textbf{Language(s)} & \textbf{Training Set} & \textbf{Year} & \textbf{Project Page} \\
\midrule
BioKaLMA~\cite{li2023towards} & Biography & EN & $\times$ & 2023 & \href{https://github.com/lixinze777/Knowledge-aware-Language-Model-Attribution}{Link} \\
ALCE~\cite{gao-etal-2023-enabling} & Misc. & EN & $\times$ & 2023 & \href{https://github.com/princeton-nlp/ALCE}{Link} \\
HAGRID~\cite{kamalloo2023hagrid} & Wikipedia & EN & $\checkmark$ & 2023 & \href{https://github.com/project-miracl/hagrid}{Link} \\
CAQA~\cite{hu2024benchmarking} & Misc. & EN & $\checkmark$ & 2024 & \href{https://github.com/HuuuNan/CAQA-Benchmark}{Link} \\
WikiRetr~\cite{li2024citation} & Wikipedia & EN & $\times$ & 2024 & \href{https://github.com/Tsinghua-dhy/CEG}{Link} \\
REASONS~\cite{saxena2024attribution} & Scientific & EN & $\checkmark$ & 2024 & \href{https://github.com/YashSaxena21/REASONS}{Link} \\
CRUD-RAG~\cite{lyu2024crud} & News & CH & $\times$ & 2024 & \href{https://github.com/IAAR-Shanghai/CRUD_RAG}{Link} \\
LongCite~\cite{zhang2024longcite} & Misc. & EN, CH & $\times$ & 2024 & \href{https://github.com/THUDM/LongCite}{Link} \\
CiteME~\cite{press2024citeme} & Misc. & EN & $\times$ & 2024 & \href{https://github.com/bethgelab/CiteME}{Link} \\
MISeD & Meeting & EN & $\times$ & 2024 & \href{https://github.com/google-research-datasets/MISeD}{Link} \\
WebCiteS~\cite{deng2024webcites} & Misc. & CH & $\checkmark$ & 2024 & \href{https://github.com/HarlynDN/WebCiteS}{Link} \\
CitaLaw~\cite{zhang2024citalaw} & Law & CH & $\times$ & 2024 & \href{https://github.com/ke-01/CitaLaw}{Link} \\
\bottomrule
\end{tabular}
\end{table*}

\subsection{Datasets and Evaluations}
In this section, we review representative datasets for external data sourcing and summarize the main evaluation protocols used to assess citation quality.

\subsubsection{Datasets}
To enhance the reliability and trustworthiness of generative LLMs, a growing body of work has focused on improving their attribution capabilities through dedicated datasets. Early efforts such as ALCE~\cite{gao-etal-2023-enabling} established a foundational benchmark by curating diverse corpora and designing automated metrics for fluency, accuracy, and citation quality. Subsequently, addressing the need for more specialized resources, datasets like HAGRID~\cite{kamalloo2023hagrid} were developed through human-LLM collaboration to provide generative retrieval data with explicit attributions. Building upon this, BioKaLMA~\cite{li2023towards} innovatively employed knowledge graphs as attribution sources, introducing the concept of ``conscious ignorance''. Concomitantly, efforts were directed towards mitigating hallucination, as exemplified by CRUD-RAG~\cite{lyu2024crud}, which integrates citation links and offers a framework for assessing citation quality. To further refine attribution analysis, \cite{hu2024benchmarking} proposed a benchmark for complex question answering, utilizing knowledge graphs for automated attribution. Complementary to these efforts, datasets such as WikiRetr~\cite{li2024citation} were constructed to analyze and enhance the ability of LLMs to generate reliable content, while MISeD~\cite{golany2024efficient} explored LLMs for generating source-specific dialogue data. Recognizing linguistic gaps, WebCiteS~\cite{deng2024webcites} offers a high-quality CH dataset. Focusing on academic rigor, the REASONS dataset~\cite{tilwani2024reasons} evaluates citation generation in scientific literature. Addressing the challenges of long-form content, ADiOSAA~\cite{sancheti2024post} reconstructs existing datasets for post-hoc answer attribution, while CiteME~\cite{press2024citeme} evaluates citation attribution under conditions of ambiguity and unattributability. LongCite~\cite{zhang2024longcite} further advances this line by enabling fine-grained, sentence-level attribution in long-context QA. Beyond general and scientific domains, CitaLaw~\cite{zhang2024citalaw} introduces the first benchmark tailored to the legal field, assessing whether LLMs can produce legally sound responses with accurate and contextually appropriate citations. Taken together, these datasets cover diverse domains, languages, and task settings, providing a solid foundation for systematic evaluation and advancement of attribution in generative information retrieval. To provide a structured overview, Table~\ref{tab:_external_data_sourcing_datasets} summarizes representative datasets for external data sourcing, including their domains, languages, availability of training sets, release years, and project pages.

\subsubsection{Evaluation Protocols}

Evaluating the quality of citations in LLM-generated responses is crucial for assessing the effectiveness of external data sourcing methods. Evaluation protocols typically fall into two categories: \textbf{human evaluation} and \textbf{automatic evaluation}. Human evaluation offers nuanced judgments but is costly and difficult to scale, whereas automatic methods provide efficiency and reproducibility at the expense of interpretability. Below, we review both approaches and the representative metrics.

\textbf{Human Evaluations.}  
Human evaluation involves annotators assessing the quality of citations by examining their relevance and alignment with the generated content. For example, in ALCE~\cite{gao-etal-2023-enabling}, Surge AI is employed to perform detailed annotation tasks. The evaluation includes two key metrics: 
\begin{itemize} [leftmargin=0.5cm]
    \item Citation Recall: Annotators are presented with a sentence and all passages it cites and are tasked with determining whether the passages fully support the sentence. 
    \item Citation Precision: Annotators evaluate whether a specific citation ``fully supports'', ``partially supports'', or ``does not support'' a given sentence. 
\end{itemize}

Although human evaluation yields high-quality and fine-grained insights, its cost, subjectivity, and limited scalability constrain its applicability in large-scale benchmarking.

\textbf{Automatic Evaluations.}  
To address the limitations of human evaluation, various automatic evaluation methods have been proposed~\cite{gao-etal-2023-enabling, yue2023automatic, djeddal2024evaluation, zhang2024towards, xu2024aliice}. These approaches seek to replicate or approximate human judgments with greater efficiency and reproducibility.

ALCE~\cite{gao-etal-2023-enabling} formalizes citation recall and citation precision in an automatic setting by leveraging the NLI model TRUE~\cite{honovich2022true} to determine whether cited passages entail the generated content. Furthermore, Djeddal et al.~\cite{djeddal2024evaluation} propose a metric called citation overlap, which calculates the overlap between the gold standard set of citations and the entire set of citations in the generated response for each query, providing a direct measure of citation alignment. To further refine citation evaluation, Zhang et al.~\cite{zhang2024towards} suggest replacing NLI models with LLMs themselves to judge citation support, thereby potentially improving the accuracy and flexibility of citation validation. More recently, ALiiCE~\cite{xu2024aliice} is designed to evaluate citation quality from positional fine-grained perspective, which parses generated responses into atomic claims using dependency trees and assesses citation recall and precision at the level of individual claims. Additionally, ALiiCE introduces a novel metric, the Coefficient of Variation of Citation Positions (CVCP), which measures the dispersion of citation placements within a sentence, offering insights into the positional fine-grained quality of citations.

In summary, while human evaluation remains the gold standard for nuanced assessment, automatic metrics are essential for large-scale, reproducible benchmarking. Recent advances increasingly combine the strengths of both paradigms, moving towards fine-grained and context-aware evaluation of citation quality.

\begin{figure}
    \centering
    \includegraphics[width=\linewidth]{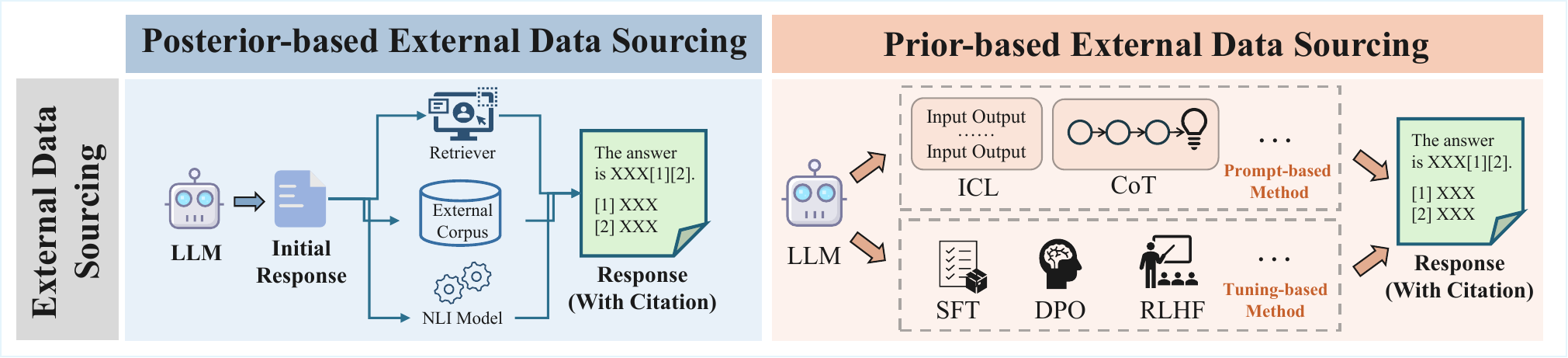}
    \caption{Overview of methods for external data sourcing in LLMs.
    Posterior-based approaches (left) validate and enrich model outputs after generation by retrieving supporting evidence from external corpus.
Prior-based approaches (right) integrate attribution into the generation process to generate responses with inline citations directly. }
    \label{fig:citation_method}
\end{figure}

\subsection{Posterior-based External Data Sourcing}
\label{sec:external-data-post}
% generate-then-retrieve
As shown in Fig.~\ref{fig:citation_method}, posterior-based external data sourcing methods differ from prior-based approaches in that they operate on the responses generated by LLMs without requiring direct access to the model's internal parameters. Instead, these methods focus on verifying the responses generated by the LLM by retrieving relevant evidence from external data sources and adding citations post hoc. 
Existing work falls into two main categories: retriever-based methods and NLI-based methods.

\subsubsection{Retriever-based Methods}
\label{sec:external-data-post-retriever}
Retriever-based methods focus on direct finding the most relevant external evidence to support LLM-generated responses after the responses have been produced. For instance, Gao et al.~\cite{gao-etal-2023-enabling} propose a naive post-cite framework that, for each statement in the generated response, identifies the best-matching passage among the top-100 candidates retrieved by state-of-the-art dense retrieval models, such as GTR~\cite{ni2021large}. The most relevant passage is then cited to substantiate the claim. Further more, RARR framework not only retrieves supporting evidence but also applies a post-processing step to revise the generated outputs~\cite{gao2023rarr}. 
By incorporating both retrieval and revision, RARR provide a practical mechanism for improving the factual reliability of LLM outputs. SearChain~\cite{xu2024search} interleaves IR with LLM reasoning by generating a full Chain-of-Query, where each query-answer node is verified and completed by retrieval before final summarisation, producing traceable citations at each reasoning step.

\subsubsection{NLI-based Methods}
\label{sec:external-data-post-nli}
While retriever-based methods focus on retrieving the most relevant evidence, they do not explicitly evaluate whether the retrieved evidence can validate the claims in the response. NLI-based methods address this limitation by incorporating natural language inference (NLI) models~\cite{pang2016text, guo2019matchzoo} to assess the relationship between the claims and the evidence. For example, Ye et al.~\cite{ye2023effective} propose a framework that extends retriever-based approaches by using an NLI model to evaluate the retrieved passages. Instead of simply citing the most relevant passage, the model selects the passage that maximally supports the claim, as determined by the entailment relationship. This ensures that the cited evidence directly validates the statement. Furthermore, CEG~\cite{li2024citation} uses an LLM equipped with predefined prompts to serve as the NLI mechanism. The LLM evaluates whether each segment of the response is factually accurate according to the retrieved evidence. If a segment is deemed factual, the corresponding reference document is added as a citation. However, if a claim lacks sufficient support, the LLM is prompted to regenerate that segment until every statement in the response is grounded in verifiable evidence.
By incorporating NLI-based verification, these approaches address the shortcomings of simple relevance-based retrieval, offering a more nuanced and reliable framework for external data sourcing.

\subsection{Prior-based External Data Sourcing}
\label{sec:external-data-prior}
% retrieve-then-generate
Prior-based external data sourcing refers to methods where LLMs generate responses along with citations to external data sources (e.g., retrieved documents) during the generation process. Unlike posterior-based approaches that verify outputs after generation, prior-based methods aim to embed attribution directly into the response generation. To provide a clearer comparison, Fig.~\ref{fig:citation_method} summarizes prior-based and posterior-based external data sourcing methods Existing studies can be broadly categorized into prompt-based methods and tuning-based methods.

\subsubsection{Prompt-based Methods}
\label{sec:external-data-prior-prompt}

\textbf{Few-shot ICL and CoT Reasoning.}  Gao et al.~\cite{gao-etal-2023-enabling} first applied few-shot In-Context Learning (ICL) to the task of attributed text generation. In this method, LLMs are prompted with a series of examples, where each example consists of a query, a set of retrieved passages, and a corresponding answer with inline citations. By observing these in-context examples, the LLM learns to generalize citation behaviors, producing responses grounded in supporting evidence. To further improve citation accuracy, Ji et al.~\cite{ji2024chain} investigate the use of Chain-of-Thought (CoT) reasoning. CoT decomposes the reasoning process into sequential steps, enabling LLMs to synthesize accurate answers from multiple documents while ensuring proper citation attribution. Building on this, CoTAR~\cite{berchansky2024cotar} further systematically analyze the effects of citation granularity (e.g., span-level, sentence-level, and passage-level) on the quality of generated responses. Their study reveals that GPT-4 performs best when applying span-level Chain-of-Thought reasoning, emphasizing the importance of granular reasoning for precise and reliable citations.

\textbf{Task-Specific Enhanced Prompt.}
Beyond general prompting strategies, subsequent works introduce task-specific prompt-based methods tailored for attributed text generation. Fierro et al.~\cite{fierro2024learning} propose a planning-based framework that enhances citation quality for long-form question-answering tasks. VTG~\cite{sun2023towards} employs long short-term memory to retain critical and recent documents, ensuring relevance throughout the generation process. Additionally, a two-tier verifier equipped with an evidence finder enables the model to reflect on the alignment between claims and citations, enhancing both precision and coverage of retrieved evidence. Moreover, Slobodkin et al.~\cite{slobodkin2024attribute} propose the ``Attribute First, Then Generate'' framework, which decomposes the conventional end-to-end generation process into three explicit steps: content selection, sentence planning, and sequential sentence generation. By structuring the generation pipeline, this method improves both the quality of the generated responses and the accuracy of citations, leading to more verifiable and accountable outputs. Most recently, MedCite~\cite{wang2025medcite} introduces an end-to-end citation generation and evaluation framework for medical QA, leveraging multi-stage retrieval and LLM-based reranking to substantially improve citation precision and recall.

In summary, prompt-based methods improve the ability of LLMs to generate content grounded in external evidence through foundational approaches like few-shot ICL and CoT reasoning, as well as task-specific enhancements that emphasize structured planning, memory mechanisms, and fine-grained attribution.

\subsubsection{Tuning-based Methods}
\label{sec:external-data-prior-tuning}

While prompt-based methods enable LLMs to perform external data sourcing through instructions and few-shot demonstrations, these approaches often result in suboptimal attribution quality. The reliance solely on prompting cannot fully address the challenge of generating accurate and fine-grained citations. To overcome these limitations, recent works have explored fine-tuning LLMs, enabling them to achieve better attribution performance. 

Ye et al.~\cite{ye2023effective} propose AGREE, a holistic framework that fine-tunes LLMs to generate self-grounded responses with accurate citations, and introduce a test-time adaptation mechanism that iteratively retrieves additional evidence to refine ungrounded claims. To improve fine-grained and consistent attribution, Huang et al.~\cite{huang2024learning} propose FRONT, a two-stage framework that first trains models to extract supporting quotes and then applies direct preference optimization (DPO)~\cite{rafailov2024direct} to align grounding with generation, substantially enhancing citation precision. Building on fine-grained supervision, Huang et al.~\cite{huang2024training} further incorporate fine-grained attribution rewards via rejection sampling and reinforcement learning to improve generalization in evidence-grounded tasks, while Li et al.~\cite{li2024improving} formulate attribution as a preference learning problem and propose an automated data collection framework with progressive preference optimization to alleviate sparse reward issues. Finally, Xia et al.~\cite{xia2024ground} address fine-grained attribution in long-form question answering through a sentence-level grounding strategy that alternates between answer generation and citation, ensuring explicit evidence support for each sentence.

In summary, tuning-based methods provide a systematic way to enhance LLMs' performance in external data sourcing tasks. By constructing high-quality fine-tuning datasets and leveraging advanced training strategies, these methods enable LLMs to produce fine-grained, consistent, and verifiable citations. This line of research marks a critical step toward improving the reliability and trustworthiness of LLM outputs in real-world applications.

%% file: sections/7_challenges.tex
\section{Challenges}

\label{sec:challenges}
While substantial progress has been made across all four dimensions of LLM sourcing (model, structure, training data, and external data) future research still faces several cross-cutting challenges.

\textbf{Complexity of Models and Data.} Modern large models feature extraordinarily intricate architectures with parameters reaching the trillion-scale. Their internal functionality is realised through the coordinated operation of distributed ‘knowledge loops’ rather than isolated modules, rendering precise structure sourcing—or even dynamic tracing of causal decision mechanisms—exceptionally difficult. Concurrently, the diverse origins, vast scale, and complex processing workflows of data pose substantial challenges to clear data sorcing. The intertwined complexity of model architecture and data collectively form the core difficulty of sourcing.

\textbf{Technical Bottlenecks and Evaluation Gaps.} Technologically, existing methods struggle to efficiently and accurately process vast datasets for fine-grained sourcing. For instance, current training data sourcing approaches often fail to simultaneously meet real-time processing demands and precision requirements. More critically, the field lacks standardised evaluation protocols and benchmarks. For instance, training data sourcing suffers from absent unified assessment criteria, while external data sourcing relies on unreliable NLI models or LLMs themselves as evaluators~\cite{li2024attributionbench}, severely hindering effective research iteration.

\textbf{Real-world Transfer and Bias Issues.} A significant gap exists between current research environments and real-world application scenarios. Machine-generated data constructed for research often deviates from actual distributions, leading to diminished attribution effectiveness during practical deployment. Furthermore, systemic attribution biases—such as model preferences towards specific sources or human-generated text—are prevalent~\cite{abolghasemi2025evaluation}, compromising the fairness and neutrality of data sourcing.

\textbf{Constraints on Data Privacy and Security.} During data sourcing, ensuring effective sourcing and verification while safeguarding data privacy and security remains an unresolved critical issue. This necessitates balancing transparency with regulatory compliance in attribution technologies.

In summary, future research should focus on developing scalable analytical techniques capable of addressing model and data complexity, establishing reliable and unified evaluation benchmarks, advancing traceability granularity towards greater precision, and prioritising robustness in real-world scenarios, bias mitigation, and privacy-security protocols. This will ultimately lay the groundwork for achieving trustworthy artificial intelligence systems.

%% file: sections/8_conclusion.tex
\section{Conclusion}

As LLMs transition from supporting objective recognition tasks to making subjective, high-stakes decisions, the opacity of their architectures and the complexity of their data pipelines make provenance tracking not just desirable, but indispensable. This survey introduced a unified, four-dimensional framework for sourcing in LLMs, spanning Model Sourcing, Model Structure Sourcing, Training Data Sourcing, and External 
Data Sourcing, and systematically categorized posterior-based and prior-based approaches within each dimension. By integrating model- and data-centric perspectives, our framework transforms provenance from a set of isolated techniques into a coherent strategy for transparency, accountability, and risk mitigation across the full lifecycle of LLM-generated content.

This holistic view enables proactive interventions such as watermarking, structural markers, and data fingerprinting, while also supporting retrospective analyses through statistical, activation-based, and gradient-based inference. The trade-offs revealed between proactive verifiability and post-hoc analyzability provide actionable guidance for developers, regulators, and researchers seeking to balance traceability with performance and privacy constraints.

Looking forward, the path to trustworthy LLM deployment will require not only advancing technical precision in provenance tracking, but also embedding these mechanisms into governance frameworks, compensation systems for data creators, and real-time safeguards against harmful or manipulative content. As LLMs become foundational infrastructure for decision-making in society, the provenance paradigm outlined here offers a practical blueprint for turning opaque generative processes into auditable, responsible, and equitable AI systems.

%% file: LLM_sourcing.bbl
%%% -*-BibTeX-*-
%%% Do NOT edit. File created by BibTeX with style
%%% ACM-Reference-Format-Journals [18-Jan-2012].

\begin{thebibliography}{222}

%%% ====================================================================
%%% NOTE TO THE USER: you can override these defaults by providing
%%% customized versions of any of these macros before the \bibliography
%%% command.  Each of them MUST provide its own final punctuation,
%%% except for \shownote{} and \showURL{}.  The latter two
%%% do not use final punctuation, in order to avoid confusing it with
%%% the Web address.
%%%
%%% To suppress output of a particular field, define its macro to expand
%%% to an empty string, or better, \unskip, like this:
%%%
%%% \newcommand{\showURL}[1]{\unskip}   % LaTeX syntax
%%%
%%% \def \showURL #1{\unskip}           % plain TeX syntax
%%%
%%% ====================================================================

\ifx \showCODEN    \undefined \def \showCODEN     #1{\unskip}     \fi
\ifx \showISBNx    \undefined \def \showISBNx     #1{\unskip}     \fi
\ifx \showISBNxiii \undefined \def \showISBNxiii  #1{\unskip}     \fi
\ifx \showISSN     \undefined \def \showISSN      #1{\unskip}     \fi
\ifx \showLCCN     \undefined \def \showLCCN      #1{\unskip}     \fi
\ifx \shownote     \undefined \def \shownote      #1{#1}          \fi
\ifx \showarticletitle \undefined \def \showarticletitle #1{#1}   \fi
\ifx \showURL      \undefined \def \showURL       {\relax}        \fi
% The following commands are used for tagged output and should be
% invisible to TeX
\providecommand\bibfield[2]{#2}
\providecommand\bibinfo[2]{#2}
\providecommand\natexlab[1]{#1}
\providecommand\showeprint[2][]{arXiv:#2}

\bibitem[Abdelrahman et~al\mbox{.}(2023)]%
        {abdelrahman2023knowledge}
\bibfield{author}{\bibinfo{person}{Ghodai Abdelrahman}, \bibinfo{person}{Qing
  Wang}, {and} \bibinfo{person}{Bernardo Nunes}.}
  \bibinfo{year}{2023}\natexlab{}.
\newblock \showarticletitle{Knowledge tracing: A survey}.
\newblock \bibinfo{journal}{\emph{Comput. Surveys}} \bibinfo{volume}{55},
  \bibinfo{number}{11} (\bibinfo{year}{2023}), \bibinfo{pages}{1--37}.
\newblock


\bibitem[Abolghasemi et~al\mbox{.}(2025)]%
        {abolghasemi2025evaluation}
\bibfield{author}{\bibinfo{person}{Amin Abolghasemi}, \bibinfo{person}{Leif
  Azzopardi}, \bibinfo{person}{Seyyed~Hadi Hashemi}, {et~al\mbox{.}}}
  \bibinfo{year}{2025}\natexlab{}.
\newblock \showarticletitle{Evaluation of Attribution Bias in Generator-Aware
  Retrieval-Augmented Large Language Models}. In
  \bibinfo{booktitle}{\emph{Findings of the ACL 2025}}.
\newblock


\bibitem[Aky{\"u}rek et~al\mbox{.}(2022)]%
        {akyurek2022towards}
\bibfield{author}{\bibinfo{person}{Ekin Aky{\"u}rek}, \bibinfo{person}{Tolga
  Bolukbasi}, \bibinfo{person}{Frederick Liu}, {et~al\mbox{.}}}
  \bibinfo{year}{2022}\natexlab{}.
\newblock \showarticletitle{Towards tracing factual knowledge in language
  models back to the training data}.
\newblock \bibinfo{journal}{\emph{arXiv preprint arXiv:2205.11482}}
  (\bibinfo{year}{2022}).
\newblock


\bibitem[Aky{\"{u}}rek et~al\mbox{.}({[n.\,d.]})]%
        {DBLP:conf/iclr/AkyurekSA0Z23}
\bibfield{author}{\bibinfo{person}{Ekin Aky{\"{u}}rek}, \bibinfo{person}{Dale
  Schuurmans}, \bibinfo{person}{Jacob Andreas}, \bibinfo{person}{Tengyu Ma},
  {and} \bibinfo{person}{Denny Zhou}.} \bibinfo{year}{[n.\,d.]}\natexlab{}.
\newblock \showarticletitle{What learning algorithm is in-context learning?
  Investigations with linear models}. In \bibinfo{booktitle}{\emph{{ICLR} 2023,
  Kigali, Rwanda, May 1-5, 2023}}.
\newblock


\bibitem[An et~al\mbox{.}(2025)]%
        {an2025defendingllmwatermarkingspoofing}
\bibfield{author}{\bibinfo{person}{Li An}, \bibinfo{person}{Yujian Liu},
  \bibinfo{person}{Yepeng Liu}, \bibinfo{person}{Yang Zhang},
  \bibinfo{person}{Yuheng Bu}, {and} \bibinfo{person}{Shiyu Chang}.}
  \bibinfo{year}{2025}\natexlab{}.
\newblock \bibinfo{title}{Defending LLM Watermarking Against Spoofing Attacks
  with Contrastive Representation Learning}.
\newblock
\showeprint[arxiv]{2504.06575}~[cs.CR]


\bibitem[Anand et~al\mbox{.}(2023)]%
        {anand2023influencescoresscaleefficient}
\bibfield{author}{\bibinfo{person}{Nikhil Anand}, \bibinfo{person}{Joshua Tan},
  {and} \bibinfo{person}{Maria Minakova}.} \bibinfo{year}{2023}\natexlab{}.
\newblock \bibinfo{title}{Influence Scores at Scale for Efficient Language Data
  Sampling}.
\newblock
\showeprint[arxiv]{2311.16298}~[cs.LG]
\urldef\tempurl%
\url{https://arxiv.org/abs/2311.16298}
\showURL{%
\tempurl}


\bibitem[Asai et~al\mbox{.}(2024)]%
        {asai2024self}
\bibfield{author}{\bibinfo{person}{Akari Asai}, \bibinfo{person}{Zeqiu Wu},
  \bibinfo{person}{Yizhong Wang}, \bibinfo{person}{Avirup Sil}, {and}
  \bibinfo{person}{Hannaneh Hajishirzi}.} \bibinfo{year}{2024}\natexlab{}.
\newblock \showarticletitle{Self-rag: Learning to retrieve, generate, and
  critique through self-reflection}.
\newblock  (\bibinfo{year}{2024}).
\newblock


\bibitem[Bae et~al\mbox{.}(2022)]%
        {bae2022if}
\bibfield{author}{\bibinfo{person}{Juhan Bae}, \bibinfo{person}{Nathan Ng},
  \bibinfo{person}{Alston Lo}, \bibinfo{person}{Marzyeh Ghassemi}, {and}
  \bibinfo{person}{Roger~B Grosse}.} \bibinfo{year}{2022}\natexlab{}.
\newblock \showarticletitle{If influence functions are the answer, then what is
  the question?}
\newblock \bibinfo{journal}{\emph{Advances in Neural Information Processing
  Systems}}  \bibinfo{volume}{35} (\bibinfo{year}{2022}),
  \bibinfo{pages}{17953--17967}.
\newblock


\bibitem[Bao et~al\mbox{.}(2024)]%
        {bao2024fastdetectgpt}
\bibfield{author}{\bibinfo{person}{Guangsheng Bao}, \bibinfo{person}{Yanbin
  Zhao}, \bibinfo{person}{Zhiyang Teng}, \bibinfo{person}{Linyi Yang}, {and}
  \bibinfo{person}{Yue Zhang}.} \bibinfo{year}{2024}\natexlab{}.
\newblock \bibinfo{title}{Fast-DetectGPT: Efficient Zero-Shot Detection of
  Machine-Generated Text via Conditional Probability Curvature}.
\newblock
\showeprint[arxiv]{2310.05130}~[cs.CL]


\bibitem[Barshan et~al\mbox{.}(2020)]%
        {barshan2020relatifidentifyingexplanatorytraining}
\bibfield{author}{\bibinfo{person}{Elnaz Barshan},
  \bibinfo{person}{Marc-Etienne Brunet}, {and}
  \bibinfo{person}{Gintare~Karolina Dziugaite}.}
  \bibinfo{year}{2020}\natexlab{}.
\newblock \bibinfo{title}{RelatIF: Identifying Explanatory Training Examples
  via Relative Influence}.
\newblock
\showeprint[arxiv]{2003.11630}~[cs.LG]
\urldef\tempurl%
\url{https://arxiv.org/abs/2003.11630}
\showURL{%
\tempurl}


\bibitem[Basu et~al\mbox{.}(2021)]%
        {basu2021influencefunctionsdeeplearning}
\bibfield{author}{\bibinfo{person}{Samyadeep Basu}, \bibinfo{person}{Philip
  Pope}, {and} \bibinfo{person}{Soheil Feizi}.}
  \bibinfo{year}{2021}\natexlab{}.
\newblock \bibinfo{title}{Influence Functions in Deep Learning Are Fragile}.
\newblock
\showeprint[arxiv]{2006.14651}~[cs.LG]
\urldef\tempurl%
\url{https://arxiv.org/abs/2006.14651}
\showURL{%
\tempurl}


\bibitem[Berchansky et~al\mbox{.}({[n.\,d.]})]%
        {berchansky2024cotar}
\bibfield{author}{\bibinfo{person}{Moshe Berchansky}, \bibinfo{person}{Daniel
  Fleischer}, \bibinfo{person}{Moshe Wasserblat}, {and} \bibinfo{person}{Peter
  Izsak}.} \bibinfo{year}{[n.\,d.]}\natexlab{}.
\newblock \showarticletitle{CoTAR: Chain-of-Thought Attribution Reasoning with
  Multi-level Granularity}. In \bibinfo{booktitle}{\emph{Findings of the
  Association for Computational Linguistics: EMNLP 2024}}.
\newblock


\bibitem[Berti et~al\mbox{.}(2025)]%
        {DBLP:journals/corr/abs-2503-05788}
\bibfield{author}{\bibinfo{person}{Leonardo Berti}, \bibinfo{person}{Flavio
  Giorgi}, {and} \bibinfo{person}{Gjergji Kasneci}.}
  \bibinfo{year}{2025}\natexlab{}.
\newblock \showarticletitle{Emergent Abilities in Large Language Models: {A}
  Survey}.
\newblock \bibinfo{journal}{\emph{CoRR}}  \bibinfo{volume}{abs/2503.05788}
  (\bibinfo{year}{2025}).
\newblock
\showeprint[arXiv]{2503.05788}
\href{https://doi.org/10.48550/ARXIV.2503.05788}{doi:\nolinkurl{10.48550/ARXIV.2503.05788}}


\bibitem[Bethge et~al\mbox{.}(2024)]%
        {moeut2024}
\bibfield{author}{\bibinfo{person}{Matthias Bethge}, \bibinfo{person}{Sebastian
  Borgeaud}, \bibinfo{person}{Jean-Baptiste Alayrac},
  \bibinfo{person}{Jony~Almagro Armenteros}, \bibinfo{person}{Mikel Artetxe},
  \bibinfo{person}{Rachel Avram}, {and} \bibinfo{person}{Timo~Barth et al.}}
  \bibinfo{year}{2024}\natexlab{}.
\newblock \bibinfo{title}{MoEUT: Mixture-of-Experts Universal Transformers}.
\newblock
\showeprint[arxiv]{2405.16039}~[cs.LG]


\bibitem[Cao et~al\mbox{.}(2024)]%
        {cao2024learn}
\bibfield{author}{\bibinfo{person}{L. Cao} {et~al\mbox{.}}}
  \bibinfo{year}{2024}\natexlab{}.
\newblock \showarticletitle{Learn to Refuse: Making Large Language Models More
  Trustworthy}. In \bibinfo{booktitle}{\emph{Proceedings of the 2024 Conference
  on Empirical Methods in Natural Language Processing}}.
  \bibinfo{publisher}{Association for Computational Linguistics}.
\newblock


\bibitem[Chai et~al\mbox{.}(2024)]%
        {chai2024training}
\bibfield{author}{\bibinfo{person}{Yekun Chai}, \bibinfo{person}{Qingyi Liu},
  \bibinfo{person}{Shuohuan Wang}, \bibinfo{person}{Yu Sun},
  \bibinfo{person}{Qiwei Peng}, {and} \bibinfo{person}{Hua Wu}.}
  \bibinfo{year}{2024}\natexlab{}.
\newblock \showarticletitle{On training data influence of gpt models}.
\newblock \bibinfo{journal}{\emph{arXiv preprint arXiv:2404.07840}}
  (\bibinfo{year}{2024}).
\newblock


\bibitem[Chang et~al\mbox{.}(2024b)]%
        {chang2024scalable}
\bibfield{author}{\bibinfo{person}{Tyler~A Chang}, \bibinfo{person}{Dheeraj
  Rajagopal}, \bibinfo{person}{Tolga Bolukbasi}, \bibinfo{person}{Lucas Dixon},
  {and} \bibinfo{person}{Ian Tenney}.} \bibinfo{year}{2024}\natexlab{b}.
\newblock \showarticletitle{Scalable influence and fact tracing for large
  language model pretraining}.
\newblock \bibinfo{journal}{\emph{arXiv preprint arXiv:2410.17413}}
  (\bibinfo{year}{2024}).
\newblock


\bibitem[Chang et~al\mbox{.}(2024a)]%
        {Chang2024PostMarkAR}
\bibfield{author}{\bibinfo{person}{Yapei Chang}, \bibinfo{person}{Kalpesh
  Krishna}, \bibinfo{person}{Amir Houmansadr}, \bibinfo{person}{J. Wieting},
  {and} \bibinfo{person}{Mohit Iyyer}.} \bibinfo{year}{2024}\natexlab{a}.
\newblock \showarticletitle{PostMark: A Robust Blackbox Watermark for Large
  Language Models}.
\newblock \bibinfo{journal}{\emph{ArXiv}}  \bibinfo{volume}{abs/2406.14517}
  (\bibinfo{year}{2024}).
\newblock


\bibitem[Chao et~al\mbox{.}(2024)]%
        {chao2024jailbreakbench}
\bibfield{author}{\bibinfo{person}{Patrick Chao}, \bibinfo{person}{Edoardo
  Debenedetti}, \bibinfo{person}{Alexander Robey}, {et~al\mbox{.}}}
  \bibinfo{year}{2024}\natexlab{}.
\newblock \showarticletitle{JailbreakBench: An Open Robustness Benchmark for
  Jailbreaking Large Language Models}. In \bibinfo{booktitle}{\emph{Advances in
  Neural Information Processing Systems}}.
\newblock


\bibitem[Chen et~al\mbox{.}(2024b)]%
        {chen2024fasttrackfastaccuratefact}
\bibfield{author}{\bibinfo{person}{Si Chen}, \bibinfo{person}{Feiyang Kang},
  \bibinfo{person}{Ning Yu}, {and} \bibinfo{person}{Ruoxi Jia}.}
  \bibinfo{year}{2024}\natexlab{b}.
\newblock \bibinfo{title}{FASTTRACK: Fast and Accurate Fact Tracing for LLMs}.
\newblock
\showeprint[arxiv]{2404.15157}~[cs.CL]
\urldef\tempurl%
\url{https://arxiv.org/abs/2404.15157}
\showURL{%
\tempurl}


\bibitem[Chen et~al\mbox{.}(2024a)]%
        {DBLP:conf/icml/ChenH0LL000Z0SY24}
\bibfield{author}{\bibinfo{person}{Wei Chen}, \bibinfo{person}{Zhen Huang},
  \bibinfo{person}{Liang Xie}, {et~al\mbox{.}}}
  \bibinfo{year}{2024}\natexlab{a}.
\newblock \showarticletitle{From Yes-Men to Truth-Tellers: Addressing
  Sycophancy in Large Language Models with Pinpoint Tuning}. In
  \bibinfo{booktitle}{\emph{{ICML} 2024}}. \bibinfo{publisher}{OpenReview.net}.
\newblock


\bibitem[Cheng et~al\mbox{.}(2023)]%
        {Cheng2023TASER}
\bibfield{author}{\bibinfo{person}{Hao Cheng}, \bibinfo{person}{Hao Fang},
  \bibinfo{person}{Xiaodong Liu}, {and} \bibinfo{person}{Jianfeng Gao}.}
  \bibinfo{year}{2023}\natexlab{}.
\newblock \showarticletitle{Task-Aware Specialization for Efficient and Robust
  Dense Retrieval for Open-Domain Question Answering}.
\newblock \bibinfo{journal}{\emph{arXiv preprint arXiv:2210.05156}}
  (\bibinfo{year}{2023}).
\newblock
\href{https://doi.org/10.48550/arXiv.2210.05156}{doi:\nolinkurl{10.48550/arXiv.2210.05156}}


\bibitem[Chiang et~al\mbox{.}(2023)]%
        {chiang2023retrieval}
\bibfield{author}{\bibinfo{person}{Ting-Rui Chiang},
  \bibinfo{person}{Xinyan~Velocity Yu}, \bibinfo{person}{Joshua Robinson},
  \bibinfo{person}{Ollie Liu}, \bibinfo{person}{Isabelle Lee}, {and}
  \bibinfo{person}{Dani Yogatama}.} \bibinfo{year}{2023}\natexlab{}.
\newblock \showarticletitle{On Retrieval Augmentation and the Limitations of
  Language Model Training}.
\newblock \bibinfo{journal}{\emph{arXiv preprint arXiv:2311.09615}}
  (\bibinfo{year}{2023}).
\newblock


\bibitem[Christ et~al\mbox{.}(2024)]%
        {christ2024undetectable}
\bibfield{author}{\bibinfo{person}{Miranda Christ}, \bibinfo{person}{Sam Gunn},
  {and} \bibinfo{person}{Or Zamir}.} \bibinfo{year}{2024}\natexlab{}.
\newblock \showarticletitle{Undetectable watermarks for language models}. In
  \bibinfo{booktitle}{\emph{The Thirty Seventh Annual Conference on Learning
  Theory}}. PMLR, \bibinfo{pages}{1125--1139}.
\newblock


\bibitem[Conmy et~al\mbox{.}(2023)]%
        {conmy2023automated}
\bibfield{author}{\bibinfo{person}{Alex Conmy}, \bibinfo{person}{Eric
  Phillips}, \bibinfo{person}{Neel Nanda}, \bibinfo{person}{Rohin Shah}, {and}
  \bibinfo{person}{Nelson Elhage}.} \bibinfo{year}{2023}\natexlab{}.
\newblock \showarticletitle{Towards Automated Circuit Discovery for Mechanistic
  Interpretability}.
\newblock \bibinfo{journal}{\emph{arXiv preprint arXiv:2304.14997}}
  (\bibinfo{year}{2023}).
\newblock
\urldef\tempurl%
\url{https://arxiv.org/abs/2304.14997}
\showURL{%
\tempurl}


\bibitem[Cui et~al\mbox{.}(2025)]%
        {cui2025robust}
\bibfield{author}{\bibinfo{person}{Xinyue Cui}, \bibinfo{person}{Johnny
  Tian-Zheng Wei}, \bibinfo{person}{Swabha Swayamdipta}, {and}
  \bibinfo{person}{Robin Jia}.} \bibinfo{year}{2025}\natexlab{}.
\newblock \showarticletitle{Robust Data Watermarking in Language Models by
  Injecting Fictitious Knowledge}.
\newblock \bibinfo{journal}{\emph{arXiv preprint arXiv:2503.04036}}
  (\bibinfo{year}{2025}).
\newblock


\bibitem[Dai et~al\mbox{.}(2022)]%
        {DBLP:conf/acl/DaiDHSCW22}
\bibfield{author}{\bibinfo{person}{Damai Dai}, \bibinfo{person}{Li Dong},
  \bibinfo{person}{Yaru Hao}, \bibinfo{person}{Zhifang Sui},
  \bibinfo{person}{Baobao Chang}, {and} \bibinfo{person}{Furu Wei}.}
  \bibinfo{year}{2022}\natexlab{}.
\newblock \showarticletitle{Knowledge Neurons in Pretrained Transformers}. In
  \bibinfo{booktitle}{\emph{{ACL} 2022}},
  \bibfield{editor}{\bibinfo{person}{Smaranda Muresan},
  \bibinfo{person}{Preslav Nakov}, {and} \bibinfo{person}{Aline Villavicencio}}
  (Eds.). \bibinfo{pages}{8493--8502}.
\newblock


\bibitem[Dai et~al\mbox{.}(2024)]%
        {dai2024bias}
\bibfield{author}{\bibinfo{person}{Sunhao Dai}, \bibinfo{person}{Chen Xu},
  \bibinfo{person}{Shicheng Xu}, \bibinfo{person}{Liang Pang},
  \bibinfo{person}{Zhenhua Dong}, {and} \bibinfo{person}{Jun Xu}.}
  \bibinfo{year}{2024}\natexlab{}.
\newblock \showarticletitle{Bias and unfairness in information retrieval
  systems: New challenges in the llm era}. In
  \bibinfo{booktitle}{\emph{Proceedings of the 30th ACM SIGKDD Conference on
  Knowledge Discovery and Data Mining}}. \bibinfo{pages}{6437--6447}.
\newblock


\bibitem[Dathathri et~al\mbox{.}(2024)]%
        {SynthIDText}
\bibfield{author}{\bibinfo{person}{Sumanth Dathathri}, \bibinfo{person}{Abigail
  See}, \bibinfo{person}{Sumedh Ghaisas}, {et~al\mbox{.}}}
  \bibinfo{year}{2024}\natexlab{}.
\newblock \showarticletitle{Scalable watermarking for identifying large
  language model outputs}.
\newblock \bibinfo{journal}{\emph{Nature}}  \bibinfo{volume}{634}
  (\bibinfo{date}{10} \bibinfo{year}{2024}), \bibinfo{pages}{818--823}.
\newblock
\href{https://doi.org/10.1038/s41586-024-08025-4}{doi:\nolinkurl{10.1038/s41586-024-08025-4}}


\bibitem[DeepSeek-AI et~al\mbox{.}(2024)]%
        {deepseekmoe2024}
\bibfield{author}{\bibinfo{person}{DeepSeek-AI}, \bibinfo{person}{Damai Dai},
  {and} \bibinfo{person}{Chengqi~Deng and}.} \bibinfo{year}{2024}\natexlab{}.
\newblock \bibinfo{title}{DeepSeekMoE: Towards Ultimate Expert Specialization
  in Mixture-of-Experts Language Models}.
\newblock
\showeprint[arxiv]{2401.06066}~[cs.CL]
\href{https://doi.org/10.48550/arXiv.2401.06066}{doi:\nolinkurl{10.48550/arXiv.2401.06066}}


\bibitem[DeLorenzo et~al\mbox{.}(2024)]%
        {delorenzo2024creativeval}
\bibfield{author}{\bibinfo{person}{Matthew DeLorenzo}, \bibinfo{person}{Vasudev
  Gohil}, {and} \bibinfo{person}{Jeyavijayan Rajendran}.}
  \bibinfo{year}{2024}\natexlab{}.
\newblock \showarticletitle{CreativEval: Evaluating Creativity of LLM-Based
  Hardware Code Generation}.
\newblock \bibinfo{journal}{\emph{arXiv preprint arXiv:2404.08806}}
  (\bibinfo{year}{2024}).
\newblock


\bibitem[Deng et~al\mbox{.}(2024c)]%
        {deng2024webcites}
\bibfield{author}{\bibinfo{person}{Haolin Deng}, \bibinfo{person}{Chang Wang},
  \bibinfo{person}{Xin Li}, {et~al\mbox{.}}} \bibinfo{year}{2024}\natexlab{c}.
\newblock \showarticletitle{WebCiteS: Attributed Query-Focused Summarization on
  Chinese Web Search Results with Citations}.
\newblock \bibinfo{journal}{\emph{arXiv preprint arXiv:2403.01774}}
  (\bibinfo{year}{2024}).
\newblock


\bibitem[Deng et~al\mbox{.}(2024b)]%
        {deng2024neuron}
\bibfield{author}{\bibinfo{person}{Jia Deng}, \bibinfo{person}{Tianyi Tang},
  \bibinfo{person}{Yanbin Yin}, \bibinfo{person}{Wenhao Yang},
  \bibinfo{person}{Wayne~Xin Zhao}, {and} \bibinfo{person}{Ji-Rong Wen}.}
  \bibinfo{year}{2024}\natexlab{b}.
\newblock \showarticletitle{Neuron-based Personality Trait Induction in Large
  Language Models}.
\newblock \bibinfo{journal}{\emph{arXiv preprint arXiv:2410.12327}}
  (\bibinfo{year}{2024}).
\newblock


\bibitem[Deng et~al\mbox{.}(2024d)]%
        {deng2024everything}
\bibfield{author}{\bibinfo{person}{Jingcheng Deng}, \bibinfo{person}{Zihao
  Wei}, \bibinfo{person}{Liang Pang}, \bibinfo{person}{Hanxing Ding},
  \bibinfo{person}{Huawei Shen}, {and} \bibinfo{person}{Xueqi Cheng}.}
  \bibinfo{year}{2024}\natexlab{d}.
\newblock \showarticletitle{Everything is Editable: Extend Knowledge Editing to
  Unstructured Data in Large Language Models}.
\newblock \bibinfo{journal}{\emph{arXiv preprint arXiv:2405.15349}}
  (\bibinfo{year}{2024}).
\newblock


\bibitem[Deng et~al\mbox{.}(2024a)]%
        {deng2024pfme}
\bibfield{author}{\bibinfo{person}{Kunquan Deng}, \bibinfo{person}{Zeyu Huang},
  \bibinfo{person}{Chen Li}, \bibinfo{person}{Chenghua Lin}, {et~al\mbox{.}}}
  \bibinfo{year}{2024}\natexlab{a}.
\newblock \showarticletitle{PFME: A Modular Approach for Fine-grained
  Hallucination Detection and Editing of Large Language Models}.
\newblock \bibinfo{journal}{\emph{arXiv preprint arXiv:2407.00488}}
  (\bibinfo{year}{2024}).
\newblock


\bibitem[Ding et~al\mbox{.}(2024)]%
        {ding2024retrieve}
\bibfield{author}{\bibinfo{person}{Hanxing Ding}, \bibinfo{person}{Liang Pang},
  \bibinfo{person}{Zihao Wei}, \bibinfo{person}{Huawei Shen}, {and}
  \bibinfo{person}{Xueqi Cheng}.} \bibinfo{year}{2024}\natexlab{}.
\newblock \showarticletitle{Retrieve only when it needs: Adaptive retrieval
  augmentation for hallucination mitigation in large language models}.
\newblock \bibinfo{journal}{\emph{arXiv preprint arXiv:2402.10612}}
  (\bibinfo{year}{2024}).
\newblock


\bibitem[Djeddal et~al\mbox{.}(2024)]%
        {djeddal2024evaluation}
\bibfield{author}{\bibinfo{person}{Hanane Djeddal}, \bibinfo{person}{Pierre
  Erbacher}, \bibinfo{person}{Raouf Toukal}, \bibinfo{person}{Laure Soulier},
  \bibinfo{person}{Karen Pinel-Sauvagnat}, {et~al\mbox{.}}}
  \bibinfo{year}{2024}\natexlab{}.
\newblock \showarticletitle{An Evaluation Framework for Attributed Information
  Retrieval using Large Language Models}. In
  \bibinfo{booktitle}{\emph{{CIKM}}}. \bibinfo{pages}{5354--5359}.
\newblock


\bibitem[Duan et~al\mbox{.}(2025a)]%
        {duan2025related}
\bibfield{author}{\bibinfo{person}{Zenghao Duan}, \bibinfo{person}{Wenbin
  Duan}, \bibinfo{person}{Zhiyi Yin}, \bibinfo{person}{Yinghan Shen},
  {et~al\mbox{.}}} \bibinfo{year}{2025}\natexlab{a}.
\newblock \showarticletitle{Related Knowledge Perturbation Matters: Rethinking
  Multiple Pieces of Knowledge Editing in Same-Subject}.
\newblock \bibinfo{journal}{\emph{arXiv preprint arXiv:2502.06868}}
  (\bibinfo{year}{2025}).
\newblock


\bibitem[Duan et~al\mbox{.}(2025b)]%
        {duan2025gloss}
\bibfield{author}{\bibinfo{person}{Zenghao Duan}, \bibinfo{person}{Zhiyi Yin},
  \bibinfo{person}{Zhichao Shi}, \bibinfo{person}{Liang Pang},
  \bibinfo{person}{Shaoling Jing}, \bibinfo{person}{Jiayi Wu}, {et~al\mbox{.}}}
  \bibinfo{year}{2025}\natexlab{b}.
\newblock \showarticletitle{GloSS over Toxicity: Understanding and Mitigating
  Toxicity in LLMs via Global Toxic Subspace}.
\newblock \bibinfo{journal}{\emph{arXiv preprint arXiv:2505.17078}}
  (\bibinfo{year}{2025}).
\newblock


\bibitem[Elhage et~al\mbox{.}(2021)]%
        {elhage2021mathematical}
\bibfield{author}{\bibinfo{person}{Nelson Elhage}, \bibinfo{person}{Neel
  Nanda}, \bibinfo{person}{Catherine Olsson}, \bibinfo{person}{Tom Henighan},
  \bibinfo{person}{Nicholas Joseph}, \bibinfo{person}{Ben Mann},
  \bibinfo{person}{Amanda Askell}, {et~al\mbox{.}}}
  \bibinfo{year}{2021}\natexlab{}.
\newblock \showarticletitle{A mathematical framework for transformer circuits}.
\newblock \bibinfo{journal}{\emph{Transformer Circuits Thread}}
  \bibinfo{volume}{1}, \bibinfo{number}{1} (\bibinfo{year}{2021}),
  \bibinfo{pages}{12}.
\newblock


\bibitem[Engstrom et~al\mbox{.}(2024)]%
        {engstrom2024dsdm}
\bibfield{author}{\bibinfo{person}{Logan Engstrom}, \bibinfo{person}{Axel
  Feldmann}, {and} \bibinfo{person}{Aleksander Madry}.}
  \bibinfo{year}{2024}\natexlab{}.
\newblock \showarticletitle{Dsdm: Model-aware dataset selection with
  datamodels, 2024}.
\newblock \bibinfo{journal}{\emph{URL https://arxiv. org/abs/2401.12926}}
  (\bibinfo{year}{2024}).
\newblock


\bibitem[Fagni et~al\mbox{.}(2021)]%
        {Fagni_2021}
\bibfield{author}{\bibinfo{person}{Tiziano Fagni}, \bibinfo{person}{Fabrizio
  Falchi}, \bibinfo{person}{Margherita Gambini}, \bibinfo{person}{Antonio
  Martella}, {and} \bibinfo{person}{Maurizio Tesconi}.}
  \bibinfo{year}{2021}\natexlab{}.
\newblock \showarticletitle{TweepFake: About detecting deepfake tweets}.
\newblock \bibinfo{journal}{\emph{PLOS ONE}} \bibinfo{volume}{16},
  \bibinfo{number}{5} (\bibinfo{date}{May} \bibinfo{year}{2021}),
  \bibinfo{pages}{e0251415}.
\newblock
\showISSN{1932-6203}
\href{https://doi.org/10.1371/journal.pone.0251415}{doi:\nolinkurl{10.1371/journal.pone.0251415}}


\bibitem[Fedus et~al\mbox{.}(2021)]%
        {fedus2021switch}
\bibfield{author}{\bibinfo{person}{William Fedus}, \bibinfo{person}{Barret
  Zoph}, {and} \bibinfo{person}{Noam Shazeer}.}
  \bibinfo{year}{2021}\natexlab{}.
\newblock \showarticletitle{Switch Transformers: Scaling to Trillion Parameter
  Models with Simple and Efficient Sparsity}.
\newblock \bibinfo{journal}{\emph{arXiv preprint arXiv:2101.03961}}
  (\bibinfo{year}{2021}).
\newblock


\bibitem[Fierro et~al\mbox{.}(2024)]%
        {fierro2024learning}
\bibfield{author}{\bibinfo{person}{Constanza Fierro},
  \bibinfo{person}{Reinald~Kim Amplayo}, \bibinfo{person}{Fantine Huot},
  \bibinfo{person}{Nicola De~Cao}, \bibinfo{person}{Joshua Maynez},
  \bibinfo{person}{Shashi Narayan}, {and} \bibinfo{person}{Mirella Lapata}.}
  \bibinfo{year}{2024}\natexlab{}.
\newblock \showarticletitle{Learning to Plan and Generate Text with Citations}.
\newblock \bibinfo{journal}{\emph{arXiv preprint arXiv:2404.03381}}
  (\bibinfo{year}{2024}).
\newblock


\bibitem[Fotouhi et~al\mbox{.}(2025)]%
        {fotouhi2025fasttrainingdatasetattribution}
\bibfield{author}{\bibinfo{person}{Milad Fotouhi},
  \bibinfo{person}{Mohammad~Taha Bahadori}, \bibinfo{person}{Oluwaseyi
  Feyisetan}, \bibinfo{person}{Payman Arabshahi}, {and} \bibinfo{person}{David
  Heckerman}.} \bibinfo{year}{2025}\natexlab{}.
\newblock \bibinfo{title}{Fast Training Dataset Attribution via In-Context
  Learning}.
\newblock
\showeprint[arxiv]{2408.11852}~[cs.CL]
\urldef\tempurl%
\url{https://arxiv.org/abs/2408.11852}
\showURL{%
\tempurl}


\bibitem[Fr{\"o}hling and Zubiaga(2021)]%
        {frohling2021feature}
\bibfield{author}{\bibinfo{person}{Leon Fr{\"o}hling} {and}
  \bibinfo{person}{Arkaitz Zubiaga}.} \bibinfo{year}{2021}\natexlab{}.
\newblock \showarticletitle{Feature-based detection of automated language
  models: tackling GPT-2, GPT-3 and Grover}.
\newblock \bibinfo{journal}{\emph{PeerJ Computer Science}}  \bibinfo{volume}{7}
  (\bibinfo{year}{2021}), \bibinfo{pages}{e443}.
\newblock


\bibitem[Fu et~al\mbox{.}(2024)]%
        {fu2024transformers}
\bibfield{author}{\bibinfo{person}{Deqing Fu}, \bibinfo{person}{Tian-Qi Chen},
  \bibinfo{person}{Robin Jia}, {and} \bibinfo{person}{Vatsal Sharan}.}
  \bibinfo{year}{2024}\natexlab{}.
\newblock \showarticletitle{Transformers learn to achieve second-order
  convergence rates for in-context linear regression}.
\newblock \bibinfo{journal}{\emph{Advances in Neural Information Processing
  Systems}}  \bibinfo{volume}{37} (\bibinfo{year}{2024}),
  \bibinfo{pages}{98675--98716}.
\newblock


\bibitem[Gall{\'e} et~al\mbox{.}(2021)]%
        {galle2021unsupervised}
\bibfield{author}{\bibinfo{person}{Matthias Gall{\'e}}, \bibinfo{person}{Jos
  Rozen}, \bibinfo{person}{Germ{\'a}n Kruszewski}, {and} \bibinfo{person}{Hady
  Elsahar}.} \bibinfo{year}{2021}\natexlab{}.
\newblock \showarticletitle{Unsupervised and distributional detection of
  machine-generated text}.
\newblock \bibinfo{journal}{\emph{arXiv preprint arXiv:2111.02878}}
  (\bibinfo{year}{2021}).
\newblock


\bibitem[Gallegos et~al\mbox{.}(2024)]%
        {gallegos2024bias}
\bibfield{author}{\bibinfo{person}{Isabel~O Gallegos}, \bibinfo{person}{Ryan~A
  Rossi}, \bibinfo{person}{Joe Barrow}, \bibinfo{person}{Md~Mehrab Tanjim},
  \bibinfo{person}{Sungchul Kim}, \bibinfo{person}{Franck Dernoncourt},
  {et~al\mbox{.}}} \bibinfo{year}{2024}\natexlab{}.
\newblock \showarticletitle{Bias and fairness in large language models: A
  survey}.
\newblock \bibinfo{journal}{\emph{Computational Linguistics}}
  \bibinfo{volume}{50}, \bibinfo{number}{3} (\bibinfo{year}{2024}),
  \bibinfo{pages}{1097--1179}.
\newblock


\bibitem[Gao et~al\mbox{.}(2023a)]%
        {gao2023rarr}
\bibfield{author}{\bibinfo{person}{Luyu Gao}, \bibinfo{person}{Zhuyun Dai},
  \bibinfo{person}{Panupong Pasupat}, {et~al\mbox{.}}}
  \bibinfo{year}{2023}\natexlab{a}.
\newblock \showarticletitle{RARR: Researching and Revising What Language Models
  Say, Using Language Models}. In \bibinfo{booktitle}{\emph{Proceedings of the
  61st Annual Meeting of the Association for Computational Linguistics (Volume
  1: Long Papers)}}. \bibinfo{pages}{16477--16508}.
\newblock


\bibitem[Gao et~al\mbox{.}(2023b)]%
        {gao-etal-2023-enabling}
\bibfield{author}{\bibinfo{person}{Tianyu Gao}, \bibinfo{person}{Howard Yen},
  \bibinfo{person}{Jiatong Yu}, {and} \bibinfo{person}{Danqi Chen}.}
  \bibinfo{year}{2023}\natexlab{b}.
\newblock \showarticletitle{Enabling Large Language Models to Generate Text
  with Citations}. In \bibinfo{booktitle}{\emph{Proceedings of the 2023
  Conference on Empirical Methods in Natural Language Processing}}.
  \bibinfo{pages}{6465--6488}.
\newblock


\bibitem[Gehman et~al\mbox{.}(2020)]%
        {gehman2020realtoxicityprompts}
\bibfield{author}{\bibinfo{person}{Samuel Gehman}, \bibinfo{person}{Suchin
  Gururangan}, \bibinfo{person}{Maarten Sap}, {et~al\mbox{.}}}
  \bibinfo{year}{2020}\natexlab{}.
\newblock \showarticletitle{RealToxicityPrompts: Evaluating Neural Toxic
  Degeneration in Language Models}. In \bibinfo{booktitle}{\emph{Findings of
  the Association for Computational Linguistics: EMNLP 2020}}.
  \bibinfo{address}{Online}.
\newblock


\bibitem[Gehrmann et~al\mbox{.}(2019)]%
        {gehrmann2019gltr}
\bibfield{author}{\bibinfo{person}{Sebastian Gehrmann},
  \bibinfo{person}{Hendrik Strobelt}, {and} \bibinfo{person}{Alexander~M.
  Rush}.} \bibinfo{year}{2019}\natexlab{}.
\newblock \bibinfo{title}{GLTR: Statistical Detection and Visualization of
  Generated Text}.
\newblock
\showeprint[arxiv]{1906.04043}~[cs.CL]


\bibitem[Geiger et~al\mbox{.}(2021)]%
        {geiger2021causal}
\bibfield{author}{\bibinfo{person}{Atticus Geiger}, \bibinfo{person}{Zhengxuan
  Wu}, \bibinfo{person}{Han Lu}, \bibinfo{person}{Joshua Rozner},
  \bibinfo{person}{Thomas Icard}, \bibinfo{person}{Christopher Potts}, {and}
  \bibinfo{person}{Noah~D. Goodman}.} \bibinfo{year}{2021}\natexlab{}.
\newblock \showarticletitle{Inducing Causal Structure for Interpretable Neural
  Networks}. In \bibinfo{booktitle}{\emph{{ICML}}}. \bibinfo{publisher}{PMLR},
  \bibinfo{pages}{3821--3832}.
\newblock


\bibitem[Geva et~al\mbox{.}(2023)]%
        {DBLP:conf/emnlp/GevaBFG23}
\bibfield{author}{\bibinfo{person}{Mor Geva}, \bibinfo{person}{Jasmijn
  Bastings}, \bibinfo{person}{Katja Filippova}, {and} \bibinfo{person}{Amir
  Globerson}.} \bibinfo{year}{2023}\natexlab{}.
\newblock \showarticletitle{Dissecting Recall of Factual Associations in
  Auto-Regressive Language Models}. In \bibinfo{booktitle}{\emph{{EMNLP}
  2023}}. \bibinfo{publisher}{Association for Computational Linguistics},
  \bibinfo{pages}{12216--12235}.
\newblock


\bibitem[Geva et~al\mbox{.}(2022a)]%
        {geva2022lm}
\bibfield{author}{\bibinfo{person}{Mor Geva}, \bibinfo{person}{Avi Caciularu},
  \bibinfo{person}{Guy Dar}, {et~al\mbox{.}}} \bibinfo{year}{2022}\natexlab{a}.
\newblock \showarticletitle{Lm-debugger: An interactive tool for inspection and
  intervention in transformer-based language models}.
\newblock \bibinfo{journal}{\emph{arXiv preprint arXiv:2204.12130}}
  (\bibinfo{year}{2022}).
\newblock


\bibitem[Geva et~al\mbox{.}(2022b)]%
        {geva2022transformer}
\bibfield{author}{\bibinfo{person}{Mor Geva}, \bibinfo{person}{Avi Caciularu},
  \bibinfo{person}{Kevin~Ro Wang}, {and} \bibinfo{person}{Yoav Goldberg}.}
  \bibinfo{year}{2022}\natexlab{b}.
\newblock \showarticletitle{Transformer feed-forward layers build predictions
  by promoting concepts in the vocabulary space}.
\newblock \bibinfo{journal}{\emph{arXiv preprint arXiv:2203.14680}}
  (\bibinfo{year}{2022}).
\newblock


\bibitem[Geva et~al\mbox{.}(2020)]%
        {geva2020transformer}
\bibfield{author}{\bibinfo{person}{Mor Geva}, \bibinfo{person}{Roei Schuster},
  \bibinfo{person}{Jonathan Berant}, {and} \bibinfo{person}{Omer Levy}.}
  \bibinfo{year}{2020}\natexlab{}.
\newblock \showarticletitle{Transformer feed-forward layers are key-value
  memories}.
\newblock \bibinfo{journal}{\emph{arXiv preprint arXiv:2012.14913}}
  (\bibinfo{year}{2020}).
\newblock


\bibitem[Golany et~al\mbox{.}(2024)]%
        {golany2024efficient}
\bibfield{author}{\bibinfo{person}{Lotem Golany}, \bibinfo{person}{Filippo
  Galgani}, \bibinfo{person}{Maya Mamo}, \bibinfo{person}{Nimrod Parasol},
  {et~al\mbox{.}}} \bibinfo{year}{2024}\natexlab{}.
\newblock \showarticletitle{Efficient Data Generation for Source-grounded
  Information-seeking Dialogs: A Use Case for Meeting Transcripts}.
\newblock \bibinfo{journal}{\emph{arXiv preprint arXiv:2405.01121}}
  (\bibinfo{year}{2024}).
\newblock


\bibitem[G{\'o}mez-Rodr{\'\i}guez and Williams(2023)]%
        {gomez2023confederacy}
\bibfield{author}{\bibinfo{person}{Carlos G{\'o}mez-Rodr{\'\i}guez} {and}
  \bibinfo{person}{Paul Williams}.} \bibinfo{year}{2023}\natexlab{}.
\newblock \showarticletitle{A confederacy of models: A comprehensive evaluation
  of LLMs on creative writing}.
\newblock \bibinfo{journal}{\emph{arXiv preprint arXiv:2310.08433}}
  (\bibinfo{year}{2023}).
\newblock


\bibitem[Grattafiori et~al\mbox{.}(2024)]%
        {grattafiori2024llama}
\bibfield{author}{\bibinfo{person}{Aaron Grattafiori},
  \bibinfo{person}{Abhimanyu Dubey}, \bibinfo{person}{Abhinav Jauhri},
  \bibinfo{person}{Abhinav Pandey}, \bibinfo{person}{Abhishek Kadian},
  \bibinfo{person}{Ahmad Al-Dahle}, {et~al\mbox{.}}}
  \bibinfo{year}{2024}\natexlab{}.
\newblock \showarticletitle{The llama 3 herd of models}.
\newblock \bibinfo{journal}{\emph{arXiv preprint arXiv:2407.21783}}
  (\bibinfo{year}{2024}).
\newblock


\bibitem[Grosse et~al\mbox{.}(2023)]%
        {grosse2023studyinglargelanguagemodel}
\bibfield{author}{\bibinfo{person}{Roger Grosse}, \bibinfo{person}{Juhan Bae},
  \bibinfo{person}{Cem Anil}, \bibinfo{person}{Nelson Elhage},
  \bibinfo{person}{Alex Tamkin}, \bibinfo{person}{Amirhossein Tajdini},
  {et~al\mbox{.}}} \bibinfo{year}{2023}\natexlab{}.
\newblock \bibinfo{title}{Studying Large Language Model Generalization with
  Influence Functions}.
\newblock
\showeprint[arxiv]{2308.03296}~[cs.LG]


\bibitem[Gu et~al\mbox{.}(2023)]%
        {gu2023watermarkingpretrainedlanguagemodels}
\bibfield{author}{\bibinfo{person}{Chenxi Gu}, \bibinfo{person}{Chengsong
  Huang}, \bibinfo{person}{Xiaoqing Zheng}, \bibinfo{person}{Kai-Wei Chang},
  {and} \bibinfo{person}{Cho-Jui Hsieh}.} \bibinfo{year}{2023}\natexlab{}.
\newblock \bibinfo{title}{Watermarking Pre-trained Language Models with
  Backdooring}.
\newblock
\showeprint[arxiv]{2210.07543}~[cs.CL]
\urldef\tempurl%
\url{https://arxiv.org/abs/2210.07543}
\showURL{%
\tempurl}


\bibitem[Guo et~al\mbox{.}(2023)]%
        {guo2023close}
\bibfield{author}{\bibinfo{person}{Biyang Guo}, \bibinfo{person}{Xin Zhang},
  \bibinfo{person}{Ziyuan Wang}, \bibinfo{person}{Minqi Jiang},
  \bibinfo{person}{Jinran Nie}, \bibinfo{person}{Yuxuan Ding},
  \bibinfo{person}{Jianwei Yue}, {and} \bibinfo{person}{Yupeng Wu}.}
  \bibinfo{year}{2023}\natexlab{}.
\newblock \bibinfo{title}{How Close is ChatGPT to Human Experts? Comparison
  Corpus, Evaluation, and Detection}.
\newblock
\showeprint[arxiv]{2301.07597}~[cs.CL]


\bibitem[Guo et~al\mbox{.}(2025)]%
        {guo2025deepseek}
\bibfield{author}{\bibinfo{person}{Daya Guo}, \bibinfo{person}{Dejian Yang},
  \bibinfo{person}{Haowei Zhang}, \bibinfo{person}{Junxiao Song},
  \bibinfo{person}{Ruoyu Zhang}, \bibinfo{person}{Runxin Xu}, {et~al\mbox{.}}}
  \bibinfo{year}{2025}\natexlab{}.
\newblock \showarticletitle{Deepseek-r1: Incentivizing reasoning capability in
  llms via reinforcement learning}.
\newblock \bibinfo{journal}{\emph{arXiv preprint arXiv:2501.12948}}
  (\bibinfo{year}{2025}).
\newblock


\bibitem[Guo et~al\mbox{.}(2021)]%
        {guo2021fastifscalableinfluencefunctions}
\bibfield{author}{\bibinfo{person}{Han Guo}, \bibinfo{person}{Nazneen~Fatema
  Rajani}, \bibinfo{person}{Peter Hase}, \bibinfo{person}{Mohit Bansal}, {and}
  \bibinfo{person}{Caiming Xiong}.} \bibinfo{year}{2021}\natexlab{}.
\newblock \bibinfo{title}{FastIF: Scalable Influence Functions for Efficient
  Model Interpretation and Debugging}.
\newblock
\showeprint[arxiv]{2012.15781}~[cs.LG]
\urldef\tempurl%
\url{https://arxiv.org/abs/2012.15781}
\showURL{%
\tempurl}


\bibitem[Guo et~al\mbox{.}(2019)]%
        {guo2019matchzoo}
\bibfield{author}{\bibinfo{person}{Jiafeng Guo}, \bibinfo{person}{Yixing Fan},
  \bibinfo{person}{Xiang Ji}, {and} \bibinfo{person}{Xueqi Cheng}.}
  \bibinfo{year}{2019}\natexlab{}.
\newblock \showarticletitle{Matchzoo: A learning, practicing, and developing
  system for neural text matching}. In \bibinfo{booktitle}{\emph{Proceedings of
  the 42Nd international ACM SIGIR conference on research and development in
  information retrieval}}. \bibinfo{pages}{1297--1300}.
\newblock


\bibitem[Gupta et~al\mbox{.}(2024)]%
        {gupta2024interpbench}
\bibfield{author}{\bibinfo{person}{Rohan Gupta}, \bibinfo{person}{Iv{\'a}n
  Arcuschin}, \bibinfo{person}{Thomas Kwa}, {and} \bibinfo{person}{Adri{\`a}
  Garriga-Alonso}.} \bibinfo{year}{2024}\natexlab{}.
\newblock \showarticletitle{InterpBench: Semi-Synthetic Transformers for
  Evaluating Mechanistic Interpretability Techniques}. In
  \bibinfo{booktitle}{\emph{{NeurIPS} 2024}}.
\newblock


\bibitem[Guu et~al\mbox{.}(2023)]%
        {guu2023simfluence}
\bibfield{author}{\bibinfo{person}{Kelvin Guu}, \bibinfo{person}{Albert
  Webson}, \bibinfo{person}{Ellie Pavlick}, \bibinfo{person}{Lucas Dixon},
  \bibinfo{person}{Ian Tenney}, {and} \bibinfo{person}{Tolga Bolukbasi}.}
  \bibinfo{year}{2023}\natexlab{}.
\newblock \showarticletitle{Simfluence: Modeling the influence of individual
  training examples by simulating training runs}.
\newblock \bibinfo{journal}{\emph{arXiv preprint arXiv:2303.08114}}
  (\bibinfo{year}{2023}).
\newblock


\bibitem[Hammoudeh and Lowd(2022)]%
        {hammoudeh2022identifying}
\bibfield{author}{\bibinfo{person}{Zayd Hammoudeh} {and}
  \bibinfo{person}{Daniel Lowd}.} \bibinfo{year}{2022}\natexlab{}.
\newblock \showarticletitle{Identifying a training-set attack's target using
  renormalized influence estimation}. In \bibinfo{booktitle}{\emph{Proceedings
  of the 2022 ACM SIGSAC Conference on Computer and Communications Security}}.
  \bibinfo{pages}{1367--1381}.
\newblock


\bibitem[Hanawa et~al\mbox{.}(2021)]%
        {hanawa2021evaluationsimilaritybasedexplanations}
\bibfield{author}{\bibinfo{person}{Kazuaki Hanawa}, \bibinfo{person}{Sho
  Yokoi}, \bibinfo{person}{Satoshi Hara}, {and} \bibinfo{person}{Kentaro
  Inui}.} \bibinfo{year}{2021}\natexlab{}.
\newblock \bibinfo{title}{Evaluation of Similarity-based Explanations}.
\newblock
\showeprint[arxiv]{2006.04528}~[cs.LG]
\urldef\tempurl%
\url{https://arxiv.org/abs/2006.04528}
\showURL{%
\tempurl}


\bibitem[Hazra et~al\mbox{.}(2024)]%
        {hazra-etal-2024-safety}
\bibfield{author}{\bibinfo{person}{Rima Hazra}, \bibinfo{person}{Sayan Layek},
  \bibinfo{person}{Somnath Banerjee}, {and} \bibinfo{person}{Soujanya Poria}.}
  \bibinfo{year}{2024}\natexlab{}.
\newblock \showarticletitle{Safety Arithmetic: A Framework for Test-time Safety
  Alignment of Language Models by Steering Parameters and Activations}. In
  \bibinfo{booktitle}{\emph{{EMNLP} 2024}}.
\newblock


\bibitem[He et~al\mbox{.}(2023)]%
        {he2023mgtbench}
\bibfield{author}{\bibinfo{person}{Xinlei He}, \bibinfo{person}{Xinyue Shen},
  \bibinfo{person}{Zeyuan Chen}, \bibinfo{person}{Michael Backes}, {and}
  \bibinfo{person}{Yang Zhang}.} \bibinfo{year}{2023}\natexlab{}.
\newblock \bibinfo{title}{MGTBench: Benchmarking Machine-Generated Text
  Detection}.
\newblock
\showeprint[arxiv]{2303.14822}~[cs.CR]


\bibitem[Hendrycks et~al\mbox{.}(2021)]%
        {hendrycks2021measuring}
\bibfield{author}{\bibinfo{person}{Dan Hendrycks}, \bibinfo{person}{Steven
  Basart}, \bibinfo{person}{Saurav Kadavath}, \bibinfo{person}{Mantas Mazeika},
  \bibinfo{person}{Akul Arora}, \bibinfo{person}{Ethan Guo},
  \bibinfo{person}{Collin Burns}, \bibinfo{person}{Samir Puranik},
  {et~al\mbox{.}}} \bibinfo{year}{2021}\natexlab{}.
\newblock \bibinfo{title}{Measuring Coding Challenge Competence With APPS}.
\newblock
\showeprint[arxiv]{2105.09938}~[cs.SE]


\bibitem[Honovich et~al\mbox{.}(2022)]%
        {honovich2022true}
\bibfield{author}{\bibinfo{person}{Or Honovich}, \bibinfo{person}{Roee
  Aharoni}, \bibinfo{person}{Jonathan Herzig}, \bibinfo{person}{Hagai
  Taitelbaum}, \bibinfo{person}{Doron Kukliansy}, \bibinfo{person}{Vered
  Cohen}, \bibinfo{person}{Thomas Scialom}, {et~al\mbox{.}}}
  \bibinfo{year}{2022}\natexlab{}.
\newblock \showarticletitle{TRUE: Re-evaluating factual consistency
  evaluation}.
\newblock \bibinfo{journal}{\emph{arXiv preprint arXiv:2204.04991}}
  (\bibinfo{year}{2022}).
\newblock


\bibitem[Hou et~al\mbox{.}(2024a)]%
        {hou2024semstampsemanticwatermarkparaphrastic}
\bibfield{author}{\bibinfo{person}{Abe~Bohan Hou}, \bibinfo{person}{Jingyu
  Zhang}, \bibinfo{person}{Tianxing He}, \bibinfo{person}{Yichen Wang},
  \bibinfo{person}{Yung-Sung Chuang}, \bibinfo{person}{Hongwei Wang},
  \bibinfo{person}{Lingfeng Shen}, \bibinfo{person}{Benjamin~Van Durme},
  \bibinfo{person}{Daniel Khashabi}, {and} \bibinfo{person}{Yulia Tsvetkov}.}
  \bibinfo{year}{2024}\natexlab{a}.
\newblock \bibinfo{title}{SemStamp: A Semantic Watermark with Paraphrastic
  Robustness for Text Generation}.
\newblock
\showeprint[arxiv]{2310.03991}~[cs.CL]
\urldef\tempurl%
\url{https://arxiv.org/abs/2310.03991}
\showURL{%
\tempurl}


\bibitem[Hou et~al\mbox{.}(2024b)]%
        {hou2024ksemstampclusteringbasedsemanticwatermark}
\bibfield{author}{\bibinfo{person}{Abe~Bohan Hou}, \bibinfo{person}{Jingyu
  Zhang}, \bibinfo{person}{Yichen Wang}, \bibinfo{person}{Daniel Khashabi},
  {and} \bibinfo{person}{Tianxing He}.} \bibinfo{year}{2024}\natexlab{b}.
\newblock \bibinfo{title}{k-SemStamp: A Clustering-Based Semantic Watermark for
  Detection of Machine-Generated Text}.
\newblock
\showeprint[arxiv]{2402.11399}~[cs.CL]


\bibitem[Houvardas and Stamatatos(2006)]%
        {houvardas2006n}
\bibfield{author}{\bibinfo{person}{John Houvardas} {and}
  \bibinfo{person}{Efstathios Stamatatos}.} \bibinfo{year}{2006}\natexlab{}.
\newblock \showarticletitle{N-gram feature selection for authorship
  identification}. In \bibinfo{booktitle}{\emph{International conference on
  artificial intelligence: Methodology, systems, and applications}}. Springer,
  \bibinfo{pages}{77--86}.
\newblock


\bibitem[Hu et~al\mbox{.}(2024)]%
        {hu2024benchmarking}
\bibfield{author}{\bibinfo{person}{Nan Hu}, \bibinfo{person}{Jiaoyan Chen},
  \bibinfo{person}{Yike Wu}, \bibinfo{person}{Guilin Qi},
  \bibinfo{person}{Sheng Bi}, \bibinfo{person}{Tongtong Wu}, {et~al\mbox{.}}}
  \bibinfo{year}{2024}\natexlab{}.
\newblock \showarticletitle{Benchmarking large language models in complex
  question answering attribution using knowledge graphs}.
\newblock \bibinfo{journal}{\emph{arXiv preprint arXiv:2401.14640}}
  (\bibinfo{year}{2024}).
\newblock


\bibitem[Hu et~al\mbox{.}(2023)]%
        {hu2023unbiased}
\bibfield{author}{\bibinfo{person}{Zhengmian Hu}, \bibinfo{person}{Lichang
  Chen}, \bibinfo{person}{Xidong Wu}, \bibinfo{person}{Yihan Wu},
  \bibinfo{person}{Hongyang Zhang}, {and} \bibinfo{person}{Heng Huang}.}
  \bibinfo{year}{2023}\natexlab{}.
\newblock \showarticletitle{Unbiased watermark for large language models}.
\newblock \bibinfo{journal}{\emph{arXiv preprint arXiv:2310.10669}}
  (\bibinfo{year}{2023}).
\newblock


\bibitem[Huang et~al\mbox{.}(2024b)]%
        {huang2024training}
\bibfield{author}{\bibinfo{person}{Chengyu Huang}, \bibinfo{person}{Zeqiu Wu},
  \bibinfo{person}{Yushi Hu}, {and} \bibinfo{person}{Wenya Wang}.}
  \bibinfo{year}{2024}\natexlab{b}.
\newblock \showarticletitle{Training language models to generate text with
  citations via fine-grained rewards}.
\newblock \bibinfo{journal}{\emph{arXiv preprint arXiv:2402.04315}}
  (\bibinfo{year}{2024}).
\newblock


\bibitem[Huang et~al\mbox{.}(2024a)]%
        {huang2024learning}
\bibfield{author}{\bibinfo{person}{Lei Huang}, \bibinfo{person}{Xiaocheng
  Feng}, \bibinfo{person}{Weitao Ma}, \bibinfo{person}{Yuxuan Gu},
  \bibinfo{person}{Weihong Zhong}, \bibinfo{person}{Xiachong Feng},
  {et~al\mbox{.}}} \bibinfo{year}{2024}\natexlab{a}.
\newblock \showarticletitle{Learning fine-grained grounded citations for
  attributed large language models}.
\newblock \bibinfo{journal}{\emph{arXiv preprint arXiv:2408.04568}}
  (\bibinfo{year}{2024}).
\newblock


\bibitem[Huang et~al\mbox{.}(2025)]%
        {huang2025survey}
\bibfield{author}{\bibinfo{person}{Lei Huang}, \bibinfo{person}{Weijiang Yu},
  \bibinfo{person}{Weitao Ma}, \bibinfo{person}{Weihong Zhong},
  {et~al\mbox{.}}} \bibinfo{year}{2025}\natexlab{}.
\newblock \showarticletitle{A survey on hallucination in large language models:
  Principles, taxonomy, challenges, and open questions}.
\newblock \bibinfo{journal}{\emph{ACM Transactions on Information Systems}}
  \bibinfo{volume}{43}, \bibinfo{number}{2} (\bibinfo{year}{2025}),
  \bibinfo{pages}{1--55}.
\newblock


\bibitem[Ilyas et~al\mbox{.}(2022)]%
        {ilyas2022datamodelspredictingpredictionstraining}
\bibfield{author}{\bibinfo{person}{Andrew Ilyas}, \bibinfo{person}{Sung~Min
  Park}, \bibinfo{person}{Logan Engstrom}, \bibinfo{person}{Guillaume Leclerc},
  {and} \bibinfo{person}{Aleksander Madry}.} \bibinfo{year}{2022}\natexlab{}.
\newblock \bibinfo{title}{Datamodels: Predicting Predictions from Training
  Data}.
\newblock
\showeprint[arxiv]{2202.00622}~[stat.ML]
\urldef\tempurl%
\url{https://arxiv.org/abs/2202.00622}
\showURL{%
\tempurl}


\bibitem[Ippolito et~al\mbox{.}(2019)]%
        {ippolito2019automatic}
\bibfield{author}{\bibinfo{person}{Daphne Ippolito}, \bibinfo{person}{Daniel
  Duckworth}, \bibinfo{person}{Chris Callison-Burch}, {and}
  \bibinfo{person}{Douglas Eck}.} \bibinfo{year}{2019}\natexlab{}.
\newblock \showarticletitle{Automatic detection of generated text is easiest
  when humans are fooled}.
\newblock \bibinfo{journal}{\emph{arXiv preprint arXiv:1911.00650}}
  (\bibinfo{year}{2019}).
\newblock


\bibitem[Ji et~al\mbox{.}(2024)]%
        {ji2024chain}
\bibfield{author}{\bibinfo{person}{Bin Ji}, \bibinfo{person}{Huijun Liu},
  \bibinfo{person}{Mingzhe Du}, {and} \bibinfo{person}{See-Kiong Ng}.}
  \bibinfo{year}{2024}\natexlab{}.
\newblock \showarticletitle{Chain-of-Thought Improves Text Generation with
  Citations in Large Language Models}. In \bibinfo{booktitle}{\emph{Proceedings
  of the AAAI Conference on Artificial Intelligence}},
  Vol.~\bibinfo{volume}{38}. \bibinfo{pages}{18345--18353}.
\newblock


\bibitem[Jiang et~al\mbox{.}(2025)]%
        {DBLP:journals/corr/abs-2505-22311}
\bibfield{author}{\bibinfo{person}{Feibo Jiang}, \bibinfo{person}{Cunhua Pan},
  \bibinfo{person}{Li Dong}, \bibinfo{person}{Kezhi Wang},
  \bibinfo{person}{Octavia~A. Dobre}, {and} \bibinfo{person}{M{\'{e}}rouane
  Debbah}.} \bibinfo{year}{2025}\natexlab{}.
\newblock \showarticletitle{From Large {AI} Models to Agentic {AI:} {A}
  Tutorial on Future Intelligent Communications}.
\newblock \bibinfo{journal}{\emph{CoRR}}  \bibinfo{volume}{abs/2505.22311}
  (\bibinfo{year}{2025}).
\newblock
\showeprint[arXiv]{2505.22311}


\bibitem[Jiang et~al\mbox{.}(2024)]%
        {Jiang2024MedMoE}
\bibfield{author}{\bibinfo{person}{Songtao Jiang}, \bibinfo{person}{Tuo Zheng},
  \bibinfo{person}{Yan Zhang}, \bibinfo{person}{Yeying Jin},
  \bibinfo{person}{Li Yuan}, {and} \bibinfo{person}{Zuozhu Liu}.}
  \bibinfo{year}{2024}\natexlab{}.
\newblock \showarticletitle{Med-MoE: Mixture of Domain-Specific Experts for
  Lightweight Medical Vision-Language Models}.
\newblock \bibinfo{journal}{\emph{arXiv preprint arXiv:2404.10237}}
  (\bibinfo{year}{2024}).
\newblock


\bibitem[Jones and Bergen(2024)]%
        {jones2024liesdamnedliesdistributional}
\bibfield{author}{\bibinfo{person}{Cameron~R. Jones} {and}
  \bibinfo{person}{Benjamin~K. Bergen}.} \bibinfo{year}{2024}\natexlab{}.
\newblock \bibinfo{title}{Lies, Damned Lies, and Distributional Language
  Statistics: Persuasion and Deception with Large Language Models}.
\newblock
\showeprint[arxiv]{2412.17128}~[cs.CL]
\urldef\tempurl%
\url{https://arxiv.org/abs/2412.17128}
\showURL{%
\tempurl}


\bibitem[Kamalloo et~al\mbox{.}(2023)]%
        {kamalloo2023hagrid}
\bibfield{author}{\bibinfo{person}{Ehsan Kamalloo}, \bibinfo{person}{Aref
  Jafari}, \bibinfo{person}{Xinyu Zhang}, \bibinfo{person}{Nandan Thakur},
  {and} \bibinfo{person}{Jimmy Lin}.} \bibinfo{year}{2023}\natexlab{}.
\newblock \showarticletitle{Hagrid: A human-llm collaborative dataset for
  generative information-seeking with attribution}.
\newblock \bibinfo{journal}{\emph{arXiv preprint arXiv:2307.16883}}
  (\bibinfo{year}{2023}).
\newblock


\bibitem[Khaddaj et~al\mbox{.}(2025)]%
        {khaddaj2025smalltolargegeneralizationdatainfluences}
\bibfield{author}{\bibinfo{person}{Alaa Khaddaj}, \bibinfo{person}{Logan
  Engstrom}, {and} \bibinfo{person}{Aleksander Madry}.}
  \bibinfo{year}{2025}\natexlab{}.
\newblock \bibinfo{title}{Small-to-Large Generalization: Data Influences Models
  Consistently Across Scale}.
\newblock
\showeprint[arxiv]{2505.16260}~[cs.LG]
\urldef\tempurl%
\url{https://arxiv.org/abs/2505.16260}
\showURL{%
\tempurl}


\bibitem[Khalifa et~al\mbox{.}(2024)]%
        {khalifa2024source}
\bibfield{author}{\bibinfo{person}{Muhammad Khalifa}, \bibinfo{person}{David
  Wadden}, \bibinfo{person}{Emma Strubell}, \bibinfo{person}{Honglak Lee},
  \bibinfo{person}{Lu Wang}, \bibinfo{person}{Iz Beltagy}, {and}
  \bibinfo{person}{Hao Peng}.} \bibinfo{year}{2024}\natexlab{}.
\newblock \showarticletitle{Source-aware training enables knowledge attribution
  in language models}.
\newblock \bibinfo{journal}{\emph{arXiv preprint arXiv:2404.01019}}
  (\bibinfo{year}{2024}).
\newblock


\bibitem[Kirchenbauer et~al\mbox{.}(2023)]%
        {kirchenbauer2023watermark}
\bibfield{author}{\bibinfo{person}{John Kirchenbauer}, \bibinfo{person}{Jonas
  Geiping}, \bibinfo{person}{Yuxin Wen}, \bibinfo{person}{Jonathan Katz},
  \bibinfo{person}{Ian Miers}, {and} \bibinfo{person}{Tom Goldstein}.}
  \bibinfo{year}{2023}\natexlab{}.
\newblock \bibinfo{title}{A Watermark for Large Language Models}.
\newblock
\showeprint[arxiv]{2301.10226}~[cs.LG]


\bibitem[Kirchenbauer et~al\mbox{.}(2024)]%
        {kirchenbauer2024on}
\bibfield{author}{\bibinfo{person}{John Kirchenbauer}, \bibinfo{person}{Jonas
  Geiping}, \bibinfo{person}{Yuxin Wen}, \bibinfo{person}{Manli Shu},
  {et~al\mbox{.}}} \bibinfo{year}{2024}\natexlab{}.
\newblock \showarticletitle{On the Reliability of Watermarks for Large Language
  Models}. In \bibinfo{booktitle}{\emph{The Twelfth International Conference on
  Learning Representations}}.
\newblock


\bibitem[Koh and Liang(2017)]%
        {koh2017understanding}
\bibfield{author}{\bibinfo{person}{Pang~Wei Koh} {and} \bibinfo{person}{Percy
  Liang}.} \bibinfo{year}{2017}\natexlab{}.
\newblock \showarticletitle{Understanding black-box predictions via influence
  functions}. In \bibinfo{booktitle}{\emph{International conference on machine
  learning}}. PMLR, \bibinfo{pages}{1885--1894}.
\newblock


\bibitem[Koh et~al\mbox{.}(2019)]%
        {koh2019accuracy}
\bibfield{author}{\bibinfo{person}{Pang Wei~W Koh}, \bibinfo{person}{Kai-Siang
  Ang}, \bibinfo{person}{Hubert Teo}, {and} \bibinfo{person}{Percy~S Liang}.}
  \bibinfo{year}{2019}\natexlab{}.
\newblock \showarticletitle{On the accuracy of influence functions for
  measuring group effects}.
\newblock \bibinfo{journal}{\emph{Advances in neural information processing
  systems}}  \bibinfo{volume}{32} (\bibinfo{year}{2019}).
\newblock


\bibitem[Konz et~al\mbox{.}(2023)]%
        {konz2023attributinglearnedconceptsneural}
\bibfield{author}{\bibinfo{person}{Nicholas Konz}, \bibinfo{person}{Charles
  Godfrey}, \bibinfo{person}{Madelyn Shapiro}, \bibinfo{person}{Jonathan Tu},
  \bibinfo{person}{Henry Kvinge}, {and} \bibinfo{person}{Davis Brown}.}
  \bibinfo{year}{2023}\natexlab{}.
\newblock \bibinfo{title}{Attributing Learned Concepts in Neural Networks to
  Training Data}.
\newblock
\showeprint[arxiv]{2310.03149}~[cs.LG]
\urldef\tempurl%
\url{https://arxiv.org/abs/2310.03149}
\showURL{%
\tempurl}


\bibitem[Krakauer et~al\mbox{.}(2025)]%
        {DBLP:journals/corr/abs-2506-11135}
\bibfield{author}{\bibinfo{person}{David~C. Krakauer}, \bibinfo{person}{John~W.
  Krakauer}, {and} \bibinfo{person}{Melanie Mitchell}.}
  \bibinfo{year}{2025}\natexlab{}.
\newblock \showarticletitle{Large Language Models and Emergence: {A} Complex
  Systems Perspective}.
\newblock \bibinfo{journal}{\emph{CoRR}}  \bibinfo{volume}{abs/2506.11135}
  (\bibinfo{year}{2025}).
\newblock
\showeprint[arXiv]{2506.11135}
\href{https://doi.org/10.48550/ARXIV.2506.11135}{doi:\nolinkurl{10.48550/ARXIV.2506.11135}}


\bibitem[Kwon et~al\mbox{.}(2024)]%
        {kwon2024datainfefficientlyestimatingdata}
\bibfield{author}{\bibinfo{person}{Yongchan Kwon}, \bibinfo{person}{Eric Wu},
  \bibinfo{person}{Kevin Wu}, {and} \bibinfo{person}{James Zou}.}
  \bibinfo{year}{2024}\natexlab{}.
\newblock \bibinfo{title}{DataInf: Efficiently Estimating Data Influence in
  LoRA-tuned LLMs and Diffusion Models}.
\newblock
\showeprint[arxiv]{2310.00902}~[cs.LG]
\urldef\tempurl%
\url{https://arxiv.org/abs/2310.00902}
\showURL{%
\tempurl}


\bibitem[Ladhak et~al\mbox{.}(2023)]%
        {ladhak-etal-2023-contrastive}
\bibfield{author}{\bibinfo{person}{Faisal Ladhak}, \bibinfo{person}{Esin
  Durmus}, {and} \bibinfo{person}{Tatsunori Hashimoto}.}
  \bibinfo{year}{2023}\natexlab{}.
\newblock \showarticletitle{Contrastive Error Attribution for Finetuned
  Language Models}. In \bibinfo{booktitle}{\emph{Proceedings of the 61st Annual
  Meeting of the Association for Computational Linguistics}}.
\newblock


\bibitem[Lai et~al\mbox{.}(2023)]%
        {lai2023large}
\bibfield{author}{\bibinfo{person}{Jinqi Lai}, \bibinfo{person}{Wensheng Gan},
  \bibinfo{person}{Jiayang Wu}, \bibinfo{person}{Zhenlian Qi}, {and}
  \bibinfo{person}{Philip~S. Yu}.} \bibinfo{year}{2023}\natexlab{}.
\newblock \bibinfo{title}{Large Language Models in Law: A Survey}.
\newblock
\showeprint[arxiv]{2312.03718}~[cs.CL]


\bibitem[Lan et~al\mbox{.}(2024)]%
        {lan2024towards}
\bibfield{author}{\bibinfo{person}{Michael Lan}, \bibinfo{person}{Philip Torr},
  {and} \bibinfo{person}{Fazl Barez}.} \bibinfo{year}{2024}\natexlab{}.
\newblock \showarticletitle{Towards Interpretable Sequence Continuation:
  Analyzing Shared Circuits in Large Language Models}. In
  \bibinfo{booktitle}{\emph{EMNLP 2024}}. \bibinfo{pages}{12576--12601}.
\newblock


\bibitem[Levy et~al\mbox{.}(2017)]%
        {levy2017zero}
\bibfield{author}{\bibinfo{person}{Omer Levy}, \bibinfo{person}{Minjoon Seo},
  \bibinfo{person}{Eunsol Choi}, {and} \bibinfo{person}{Luke Zettlemoyer}.}
  \bibinfo{year}{2017}\natexlab{}.
\newblock \showarticletitle{Zero-shot relation extraction via reading
  comprehension}.
\newblock \bibinfo{journal}{\emph{arXiv preprint arXiv:1706.04115}}
  (\bibinfo{year}{2017}).
\newblock


\bibitem[Li et~al\mbox{.}(2024c)]%
        {li2024improving}
\bibfield{author}{\bibinfo{person}{Dongfang Li}, \bibinfo{person}{Zetian Sun},
  \bibinfo{person}{Baotian Hu}, \bibinfo{person}{Zhenyu Liu},
  \bibinfo{person}{Xinshuo Hu}, \bibinfo{person}{Xuebo Liu}, {et~al\mbox{.}}}
  \bibinfo{year}{2024}\natexlab{c}.
\newblock \showarticletitle{Improving Attributed Text Generation of Large
  Language Models via Preference Learning}.
\newblock \bibinfo{journal}{\emph{arXiv preprint arXiv:2403.18381}}
  (\bibinfo{year}{2024}).
\newblock


\bibitem[Li et~al\mbox{.}(2024a)]%
        {li2024nearest}
\bibfield{author}{\bibinfo{person}{Minghan Li}, \bibinfo{person}{Xilun Chen},
  \bibinfo{person}{Ari Holtzman}, \bibinfo{person}{Beidi Chen},
  \bibinfo{person}{Jimmy Lin}, {et~al\mbox{.}}}
  \bibinfo{year}{2024}\natexlab{a}.
\newblock \showarticletitle{Nearest neighbor speculative decoding for llm
  generation and attribution}.
\newblock \bibinfo{journal}{\emph{Advances in Neural Information Processing
  Systems}}  \bibinfo{volume}{37} (\bibinfo{year}{2024}),
  \bibinfo{pages}{80987--81015}.
\newblock


\bibitem[Li and Fung(2025)]%
        {li2025security}
\bibfield{author}{\bibinfo{person}{Miles~Q Li} {and} \bibinfo{person}{Benjamin
  Fung}.} \bibinfo{year}{2025}\natexlab{}.
\newblock \showarticletitle{Security Concerns for Large Language Models: A
  Survey}.
\newblock \bibinfo{journal}{\emph{arXiv preprint arXiv:2505.18889}}
  (\bibinfo{year}{2025}).
\newblock


\bibitem[Li et~al\mbox{.}(2024b)]%
        {li2024citation}
\bibfield{author}{\bibinfo{person}{Weitao Li}, \bibinfo{person}{Junkai Li},
  \bibinfo{person}{Weizhi Ma}, {and} \bibinfo{person}{Yang Liu}.}
  \bibinfo{year}{2024}\natexlab{b}.
\newblock \showarticletitle{Citation-Enhanced Generation for LLM-based
  Chatbot}.
\newblock \bibinfo{journal}{\emph{arXiv preprint arXiv:2402.16063}}
  (\bibinfo{year}{2024}).
\newblock


\bibitem[Li et~al\mbox{.}(2023)]%
        {li2023towards}
\bibfield{author}{\bibinfo{person}{Xinze Li}, \bibinfo{person}{Yixin Cao},
  \bibinfo{person}{Liangming Pan}, \bibinfo{person}{Yubo Ma}, {and}
  \bibinfo{person}{Aixin Sun}.} \bibinfo{year}{2023}\natexlab{}.
\newblock \showarticletitle{Towards verifiable generation: A benchmark for
  knowledge-aware language model attribution}.
\newblock \bibinfo{journal}{\emph{arXiv preprint arXiv:2310.05634}}
  (\bibinfo{year}{2023}).
\newblock


\bibitem[Li et~al\mbox{.}(2024g)]%
        {li2024llatrieval}
\bibfield{author}{\bibinfo{person}{Xiaonan Li}, \bibinfo{person}{Changtai Zhu},
  \bibinfo{person}{Linyang Li}, {et~al\mbox{.}}}
  \bibinfo{year}{2024}\natexlab{g}.
\newblock \showarticletitle{LLatrieval: LLM-Verified Retrieval for Verifiable
  Generation}. In \bibinfo{booktitle}{\emph{Proceedings of the 2024 Conference
  of the North American Chapter of the Association for Computational
  Linguistics}}.
\newblock


\bibitem[Li et~al\mbox{.}(2022)]%
        {Li_2022}
\bibfield{author}{\bibinfo{person}{Yujia Li}, \bibinfo{person}{David Choi},
  \bibinfo{person}{Junyoung Chung}, \bibinfo{person}{Nate Kushman},
  \bibinfo{person}{Julian Schrittwieser}, \bibinfo{person}{Rémi Leblond},
  {et~al\mbox{.}}} \bibinfo{year}{2022}\natexlab{}.
\newblock \showarticletitle{Competition-level code generation with AlphaCode}.
\newblock \bibinfo{journal}{\emph{Science}} \bibinfo{volume}{378},
  \bibinfo{number}{6624} (\bibinfo{date}{Dec.} \bibinfo{year}{2022}),
  \bibinfo{pages}{1092–1097}.
\newblock
\showISSN{1095-9203}


\bibitem[Li et~al\mbox{.}(2024e)]%
        {li2024attributionbench}
\bibfield{author}{\bibinfo{person}{Yifei Li}, \bibinfo{person}{Xiang Yue},
  \bibinfo{person}{Zeyi Liao}, {and} \bibinfo{person}{Huan Sun}.}
  \bibinfo{year}{2024}\natexlab{e}.
\newblock \showarticletitle{AttributionBench: How Hard is Automatic Attribution
  Evaluation?}
\newblock \bibinfo{journal}{\emph{arXiv preprint arXiv:2402.15089}}
  (\bibinfo{year}{2024}).
\newblock


\bibitem[Li et~al\mbox{.}(2024d)]%
        {dynmoe2024}
\bibfield{author}{\bibinfo{person}{Zefei Li}, \bibinfo{person}{Xin Wang},
  \bibinfo{person}{Yusen Zhang}, \bibinfo{person}{Boning Zhang},
  \bibinfo{person}{Ningyu Zhang}, \bibinfo{person}{Siyuan Cheng},
  \bibinfo{person}{Runxin Xu}, {et~al\mbox{.}}}
  \bibinfo{year}{2024}\natexlab{d}.
\newblock \bibinfo{title}{Dynamic Mixture of Experts: An Auto-Tuning Approach
  for Efficient Transformer Models}.
\newblock
\showeprint[arxiv]{2402.06305}~[cs.CV]


\bibitem[Li et~al\mbox{.}(2024f)]%
        {li2024influence}
\bibfield{author}{\bibinfo{person}{Zhe Li}, \bibinfo{person}{Wei Zhao},
  \bibinfo{person}{Yige Li}, {and} \bibinfo{person}{Jun Sun}.}
  \bibinfo{year}{2024}\natexlab{f}.
\newblock \showarticletitle{Do Influence Functions Work on Large Language
  Models?}
\newblock \bibinfo{journal}{\emph{arXiv preprint arXiv:2409.19998}}
  (\bibinfo{year}{2024}).
\newblock


\bibitem[Lin et~al\mbox{.}(2022)]%
        {lin2022truthfulqa}
\bibfield{author}{\bibinfo{person}{Stephanie Lin}, \bibinfo{person}{Jacob
  Hilton}, {and} \bibinfo{person}{Owain Evans}.}
  \bibinfo{year}{2022}\natexlab{}.
\newblock \showarticletitle{TruthfulQA: Measuring How Models Mimic Human
  Falsehoods}. In \bibinfo{booktitle}{\emph{Proceedings of the 60th Annual
  Meeting of the Association for Computational Linguistics}}.
  \bibinfo{pages}{3214--3252}.
\newblock


\bibitem[Lindsey et~al\mbox{.}(2025)]%
        {lindsey2025biology}
\bibfield{author}{\bibinfo{person}{Jack Lindsey}, \bibinfo{person}{Wes Gurnee},
  \bibinfo{person}{Emmanuel Ameisen}, \bibinfo{person}{Brian Chen},
  \bibinfo{person}{Adam Pearce}, \bibinfo{person}{Nicholas~L. Turner},
  {et~al\mbox{.}}} \bibinfo{year}{2025}\natexlab{}.
\newblock \showarticletitle{On the Biology of a Large Language Model}.
\newblock \bibinfo{journal}{\emph{Transformer Circuits Thread}}
  (\bibinfo{year}{2025}).
\newblock


\bibitem[Liu et~al\mbox{.}(2024)]%
        {liu2024unforgeable}
\bibfield{author}{\bibinfo{person}{Aiwei Liu}, \bibinfo{person}{Leyi Pan},
  \bibinfo{person}{Xuming Hu}, \bibinfo{person}{Shu'ang Li},
  \bibinfo{person}{Lijie Wen}, \bibinfo{person}{Irwin King}, {and}
  \bibinfo{person}{Philip~S. Yu}.} \bibinfo{year}{2024}\natexlab{}.
\newblock \bibinfo{title}{An Unforgeable Publicly Verifiable Watermark for
  Large Language Models}.
\newblock
\showeprint[arxiv]{2307.16230}~[cs.CL]


\bibitem[Liu et~al\mbox{.}(2025)]%
        {liu2025attribotbagtricksefficiently}
\bibfield{author}{\bibinfo{person}{Fengyuan Liu}, \bibinfo{person}{Nikhil
  Kandpal}, {and} \bibinfo{person}{Colin Raffel}.}
  \bibinfo{year}{2025}\natexlab{}.
\newblock \bibinfo{title}{AttriBoT: A Bag of Tricks for Efficiently
  Approximating Leave-One-Out Context Attribution}.
\newblock
\showeprint[arxiv]{2411.15102}~[cs.LG]
\urldef\tempurl%
\url{https://arxiv.org/abs/2411.15102}
\showURL{%
\tempurl}


\bibitem[Liu et~al\mbox{.}(2023a)]%
        {liu2023cococoherenceenhancedmachinegeneratedtext}
\bibfield{author}{\bibinfo{person}{Xiaoming Liu}, \bibinfo{person}{Zhaohan
  Zhang}, \bibinfo{person}{Yichen Wang}, \bibinfo{person}{Hang Pu},
  \bibinfo{person}{Yu Lan}, {and} \bibinfo{person}{Chao Shen}.}
  \bibinfo{year}{2023}\natexlab{a}.
\newblock \bibinfo{title}{CoCo: Coherence-Enhanced Machine-Generated Text
  Detection Under Data Limitation With Contrastive Learning}.
\newblock
\showeprint[arxiv]{2212.10341}~[cs.CL]


\bibitem[Liu and Bu(2024)]%
        {liu2024adaptivetextwatermarklarge}
\bibfield{author}{\bibinfo{person}{Yepeng Liu} {and} \bibinfo{person}{Yuheng
  Bu}.} \bibinfo{year}{2024}\natexlab{}.
\newblock \bibinfo{title}{Adaptive Text Watermark for Large Language Models}.
\newblock
\showeprint[arxiv]{2401.13927}~[cs.CL]
\urldef\tempurl%
\url{https://arxiv.org/abs/2401.13927}
\showURL{%
\tempurl}


\bibitem[Liu et~al\mbox{.}(2023b)]%
        {liu2023argugpt}
\bibfield{author}{\bibinfo{person}{Yikang Liu}, \bibinfo{person}{Ziyin Zhang},
  \bibinfo{person}{Wanyang Zhang}, \bibinfo{person}{Shisen Yue},
  \bibinfo{person}{Xiaojing Zhao}, \bibinfo{person}{Xinyuan Cheng},
  {et~al\mbox{.}}} \bibinfo{year}{2023}\natexlab{b}.
\newblock \bibinfo{title}{ArguGPT: evaluating, understanding and identifying
  argumentative essays generated by GPT models}.
\newblock
\showeprint[arxiv]{2304.07666}~[cs.CL]


\bibitem[Luong et~al\mbox{.}(2024)]%
        {luong2024realistic}
\bibfield{author}{\bibinfo{person}{Tinh~Son Luong},
  \bibinfo{person}{Thanh‐Thien Le}, \bibinfo{person}{Linh Ngo~Van}, {and}
  \bibinfo{person}{Thien~Huu Nguyen}.} \bibinfo{year}{2024}\natexlab{}.
\newblock \showarticletitle{Realistic Evaluation of Toxicity in Large Language
  Models}.
\newblock \bibinfo{journal}{\emph{arXiv preprint arXiv:2405.10659}}
  (\bibinfo{year}{2024}).
\newblock
\urldef\tempurl%
\url{https://arxiv.org/abs/2405.10659}
\showURL{%
\tempurl}


\bibitem[Lv et~al\mbox{.}(2024)]%
        {DBLP:journals/corr/abs-2403-19521}
\bibfield{author}{\bibinfo{person}{Ang Lv}, \bibinfo{person}{Kaiyi Zhang},
  \bibinfo{person}{Yuhan Chen}, \bibinfo{person}{Yulong Wang},
  \bibinfo{person}{Lifeng Liu}, \bibinfo{person}{Ji{-}Rong Wen},
  \bibinfo{person}{Jian Xie}, {and} \bibinfo{person}{Rui Yan}.}
  \bibinfo{year}{2024}\natexlab{}.
\newblock \showarticletitle{Interpreting Key Mechanisms of Factual Recall in
  Transformer-Based Language Models}.
\newblock \bibinfo{journal}{\emph{CoRR}}  \bibinfo{volume}{abs/2403.19521}
  (\bibinfo{year}{2024}).
\newblock


\bibitem[Lyu et~al\mbox{.}(2024)]%
        {lyu2024crud}
\bibfield{author}{\bibinfo{person}{Yuanjie Lyu}, \bibinfo{person}{Zhiyu Li},
  \bibinfo{person}{Simin Niu}, \bibinfo{person}{Feiyu Xiong}, {et~al\mbox{.}}}
  \bibinfo{year}{2024}\natexlab{}.
\newblock \showarticletitle{Crud-rag: A comprehensive chinese benchmark for
  retrieval-augmented generation of large language models}.
\newblock \bibinfo{journal}{\emph{ACM Transactions on Information Systems}}
  (\bibinfo{year}{2024}).
\newblock


\bibitem[Mao et~al\mbox{.}(2024)]%
        {mao2024raidar}
\bibfield{author}{\bibinfo{person}{Chengzhi Mao}, \bibinfo{person}{Carl
  Vondrick}, \bibinfo{person}{Hao Wang}, {and} \bibinfo{person}{Junfeng Yang}.}
  \bibinfo{year}{2024}\natexlab{}.
\newblock \showarticletitle{Raidar: generative ai detection via rewriting}.
\newblock \bibinfo{journal}{\emph{arXiv preprint arXiv:2401.12970}}
  (\bibinfo{year}{2024}).
\newblock


\bibitem[Meng et~al\mbox{.}(2022)]%
        {meng2022locating_neurips}
\bibfield{author}{\bibinfo{person}{Kevin Meng}, \bibinfo{person}{David Bau},
  \bibinfo{person}{Alex Andonian}, {and} \bibinfo{person}{Yonatan Belinkov}.}
  \bibinfo{year}{2022}\natexlab{}.
\newblock \showarticletitle{Locating and Editing Factual Associations in GPT}.
  In \bibinfo{booktitle}{\emph{Advances in Neural Information Processing
  Systems}}, Vol.~\bibinfo{volume}{35}.
\newblock


\bibitem[Meng et~al\mbox{.}(2023)]%
        {DBLP:conf/iclr/MengSABB23}
\bibfield{author}{\bibinfo{person}{Kevin Meng}, \bibinfo{person}{Arnab~Sen
  Sharma}, \bibinfo{person}{Alex~J. Andonian}, \bibinfo{person}{Yonatan
  Belinkov}, {and} \bibinfo{person}{David Bau}.}
  \bibinfo{year}{2023}\natexlab{}.
\newblock \showarticletitle{Mass-Editing Memory in a Transformer}. In
  \bibinfo{booktitle}{\emph{{ICLR} 2023, Kigali, Rwanda, May 1-5, 2023}}.
  \bibinfo{publisher}{OpenReview.net}.
\newblock


\bibitem[Merullo et~al\mbox{.}({[n.\,d.]})]%
        {merullocircuit}
\bibfield{author}{\bibinfo{person}{Jack Merullo}, \bibinfo{person}{Carsten
  Eickhoff}, {and} \bibinfo{person}{Ellie Pavlick}.}
  \bibinfo{year}{[n.\,d.]}\natexlab{}.
\newblock \showarticletitle{Circuit Component Reuse Across Tasks in Transformer
  Language Models}. In \bibinfo{booktitle}{\emph{The Twelfth International
  Conference on Learning Representations}}.
\newblock


\bibitem[Methuku and Myakala(2025)]%
        {methuku2025digital}
\bibfield{author}{\bibinfo{person}{Vijayalaxmi Methuku} {and}
  \bibinfo{person}{Praveen~Kumar Myakala}.} \bibinfo{year}{2025}\natexlab{}.
\newblock \showarticletitle{Digital doppelgangers: Ethical and societal
  implications of pre-mortem ai clones}.
\newblock \bibinfo{journal}{\emph{arXiv preprint arXiv:2502.21248}}
  (\bibinfo{year}{2025}).
\newblock


\bibitem[Min et~al\mbox{.}(2025)]%
        {min2025understanding}
\bibfield{author}{\bibinfo{person}{Taywon Min}, \bibinfo{person}{Haeone Lee},
  \bibinfo{person}{Yongchan Kwon}, {and} \bibinfo{person}{Kimin Lee}.}
  \bibinfo{year}{2025}\natexlab{}.
\newblock \showarticletitle{Understanding Impact of Human Feedback via
  Influence Functions}.
\newblock \bibinfo{journal}{\emph{arXiv preprint arXiv:2501.05790}}
  (\bibinfo{year}{2025}).
\newblock


\bibitem[Mitchell et~al\mbox{.}(2023)]%
        {mitchell2023detectgpt}
\bibfield{author}{\bibinfo{person}{Eric Mitchell}, \bibinfo{person}{Yoonho
  Lee}, \bibinfo{person}{Alexander Khazatsky}, \bibinfo{person}{Christopher~D.
  Manning}, {and} \bibinfo{person}{Chelsea Finn}.}
  \bibinfo{year}{2023}\natexlab{}.
\newblock \bibinfo{title}{DetectGPT: Zero-Shot Machine-Generated Text Detection
  using Probability Curvature}.
\newblock
\showeprint[arxiv]{2301.11305}~[cs.CL]


\bibitem[Mitrović et~al\mbox{.}(2023)]%
        {mitrović2023chatgpt}
\bibfield{author}{\bibinfo{person}{Sandra Mitrović}, \bibinfo{person}{Davide
  Andreoletti}, {and} \bibinfo{person}{Omran Ayoub}.}
  \bibinfo{year}{2023}\natexlab{}.
\newblock \bibinfo{title}{ChatGPT or Human? Detect and Explain. Explaining
  Decisions of Machine Learning Model for Detecting Short ChatGPT-generated
  Text}.
\newblock
\showeprint[arxiv]{2301.13852}~[cs.CL]


\bibitem[Nguyen et~al\mbox{.}(2023)]%
        {nguyen2023a}
\bibfield{author}{\bibinfo{person}{Elisa Nguyen}, \bibinfo{person}{Minjoon
  Seo}, {and} \bibinfo{person}{Seong~Joon Oh}.}
  \bibinfo{year}{2023}\natexlab{}.
\newblock \showarticletitle{A Bayesian Approach To Analysing Training Data
  Attribution In Deep Learning}. In \bibinfo{booktitle}{\emph{Thirty-seventh
  Conference on Neural Information Processing Systems}}.
\newblock


\bibitem[Ni et~al\mbox{.}(2021)]%
        {ni2021large}
\bibfield{author}{\bibinfo{person}{Jianmo Ni}, \bibinfo{person}{Chen Qu},
  \bibinfo{person}{Jing Lu}, \bibinfo{person}{Zhuyun Dai},
  \bibinfo{person}{Gustavo~Hern{\'a}ndez {\'A}brego}, \bibinfo{person}{Ji Ma},
  \bibinfo{person}{Vincent~Y Zhao}, \bibinfo{person}{Yi Luan},
  \bibinfo{person}{Keith~B Hall}, \bibinfo{person}{Ming-Wei Chang},
  {et~al\mbox{.}}} \bibinfo{year}{2021}\natexlab{}.
\newblock \showarticletitle{Large dual encoders are generalizable retrievers}.
\newblock \bibinfo{journal}{\emph{arXiv preprint arXiv:2112.07899}}
  (\bibinfo{year}{2021}).
\newblock


\bibitem[Nikankin et~al\mbox{.}(2024)]%
        {nikankin2024arithmetic}
\bibfield{author}{\bibinfo{person}{Yaniv Nikankin}, \bibinfo{person}{Anja
  Reusch}, \bibinfo{person}{Aaron Mueller}, {and} \bibinfo{person}{Yonatan
  Belinkov}.} \bibinfo{year}{2024}\natexlab{}.
\newblock \showarticletitle{Arithmetic Without Algorithms: Language Models
  Solve Math With a Bag of Heuristics}.
\newblock \bibinfo{journal}{\emph{arXiv preprint arXiv:2410.21272}}
  (\bibinfo{year}{2024}).
\newblock


\bibitem[Olah et~al\mbox{.}(2020)]%
        {olah2020zoom}
\bibfield{author}{\bibinfo{person}{Chris Olah}, \bibinfo{person}{Nick
  Cammarata}, \bibinfo{person}{Ludwig Schubert}, \bibinfo{person}{Gabriel Goh},
  \bibinfo{person}{Michael Petrov}, {and} \bibinfo{person}{Shan Carter}.}
  \bibinfo{year}{2020}\natexlab{}.
\newblock \showarticletitle{Zoom in: An introduction to circuits}.
\newblock \bibinfo{journal}{\emph{Distill}} \bibinfo{volume}{5},
  \bibinfo{number}{3} (\bibinfo{year}{2020}), \bibinfo{pages}{e00024--001}.
\newblock


\bibitem[OpenAI(2023)]%
        {openai2023gpt4}
\bibfield{author}{\bibinfo{person}{OpenAI}.} \bibinfo{year}{2023}\natexlab{}.
\newblock \bibinfo{title}{GPT-4 Technical Report}.
\newblock
\showeprint[arxiv]{2303.08774}~[cs.CL]


\bibitem[Pan et~al\mbox{.}(2023)]%
        {pan2023unifying}
\bibfield{author}{\bibinfo{person}{Shirui Pan}, \bibinfo{person}{Linhao Luo},
  \bibinfo{person}{Yufei Wang}, \bibinfo{person}{Chen Chen},
  \bibinfo{person}{Jiapu Wang}, {and} \bibinfo{person}{Xindong Wu}.}
  \bibinfo{year}{2023}\natexlab{}.
\newblock \showarticletitle{Unifying Large Language Models and Knowledge
  Graphs: A Roadmap}.
\newblock \bibinfo{journal}{\emph{arXiv preprint arXiv:2306.08302}}
  (\bibinfo{year}{2023}).
\newblock


\bibitem[Panaitescu-Liess et~al\mbox{.}(2025)]%
        {panaitescu2025can}
\bibfield{author}{\bibinfo{person}{Michael-Andrei Panaitescu-Liess},
  \bibinfo{person}{Zora Che}, \bibinfo{person}{Bang An}, {et~al\mbox{.}}}
  \bibinfo{year}{2025}\natexlab{}.
\newblock \showarticletitle{Can watermarking large language models prevent
  copyrighted text generation and hide training data?}. In
  \bibinfo{booktitle}{\emph{Proceedings of the AAAI Conference on Artificial
  Intelligence}}.
\newblock


\bibitem[Pang et~al\mbox{.}(2016)]%
        {pang2016text}
\bibfield{author}{\bibinfo{person}{Liang Pang}, \bibinfo{person}{Yanyan Lan},
  \bibinfo{person}{Jiafeng Guo}, \bibinfo{person}{Jun Xu},
  \bibinfo{person}{Shengxian Wan}, {and} \bibinfo{person}{Xueqi Cheng}.}
  \bibinfo{year}{2016}\natexlab{}.
\newblock \showarticletitle{Text matching as image recognition}. In
  \bibinfo{booktitle}{\emph{Proceedings of the AAAI conference on artificial
  intelligence}}, Vol.~\bibinfo{volume}{30}.
\newblock


\bibitem[Pang et~al\mbox{.}(2024)]%
        {pang2024attacking}
\bibfield{author}{\bibinfo{person}{Qi Pang}, \bibinfo{person}{Shengyuan Hu},
  \bibinfo{person}{Wenting Zheng}, {and} \bibinfo{person}{Virginia Smith}.}
  \bibinfo{year}{2024}\natexlab{}.
\newblock \showarticletitle{Attacking LLM Watermarks by Exploiting Their
  Strengths}.
\newblock \bibinfo{journal}{\emph{arXiv preprint arXiv:2402.16187}}
  (\bibinfo{year}{2024}).
\newblock


\bibitem[Park et~al\mbox{.}(2023)]%
        {Park2023trak}
\bibfield{author}{\bibinfo{person}{Sung~Min Park}, \bibinfo{person}{Kristian
  Georgiev}, \bibinfo{person}{Andrew Ilyas}, \bibinfo{person}{Guillaume
  Leclerc}, {and} \bibinfo{person}{Aleksander Madry}.}
  \bibinfo{year}{2023}\natexlab{}.
\newblock \showarticletitle{TRAK: attributing model behavior at scale}. In
  \bibinfo{booktitle}{\emph{Proceedings of the 40th International Conference on
  Machine Learning}}.
\newblock


\bibitem[Parshakov et~al\mbox{.}(2025)]%
        {parshakov2025usersfavorllmgeneratedcontent}
\bibfield{author}{\bibinfo{person}{Petr Parshakov}, \bibinfo{person}{Iuliia
  Naidenova}, \bibinfo{person}{Sofia Paklina}, \bibinfo{person}{Nikita Matkin},
  {and} \bibinfo{person}{Cornel Nesseler}.} \bibinfo{year}{2025}\natexlab{}.
\newblock \bibinfo{title}{Users Favor LLM-Generated Content -- Until They Know
  It's AI}.
\newblock
\showeprint[arxiv]{2503.16458}~[cs.HC]
\urldef\tempurl%
\url{https://arxiv.org/abs/2503.16458}
\showURL{%
\tempurl}


\bibitem[Petroni et~al\mbox{.}(2019)]%
        {petroni2019language}
\bibfield{author}{\bibinfo{person}{Fabio Petroni}, \bibinfo{person}{Tim
  Rockt{\"a}schel}, \bibinfo{person}{Patrick Lewis}, \bibinfo{person}{Anton
  Bakhtin}, \bibinfo{person}{Yuxiang Wu}, \bibinfo{person}{Alexander~H Miller},
  {and} \bibinfo{person}{Sebastian Riedel}.} \bibinfo{year}{2019}\natexlab{}.
\newblock \showarticletitle{Language models as knowledge bases?}
\newblock \bibinfo{journal}{\emph{arXiv preprint arXiv:1909.01066}}
  (\bibinfo{year}{2019}).
\newblock


\bibitem[Pezeshkpour et~al\mbox{.}(2021)]%
        {pezeshkpour2021empirical}
\bibfield{author}{\bibinfo{person}{Pouya Pezeshkpour}, \bibinfo{person}{Sarthak
  Jain}, \bibinfo{person}{Byron~C Wallace}, {and} \bibinfo{person}{Sameer
  Singh}.} \bibinfo{year}{2021}\natexlab{}.
\newblock \showarticletitle{An empirical comparison of instance attribution
  methods for NLP}.
\newblock \bibinfo{journal}{\emph{arXiv preprint arXiv:2104.04128}}
  (\bibinfo{year}{2021}).
\newblock


\bibitem[Piet et~al\mbox{.}(2023)]%
        {piet2023mark}
\bibfield{author}{\bibinfo{person}{Julien Piet}, \bibinfo{person}{Chawin
  Sitawarin}, \bibinfo{person}{Vivian Fang}, \bibinfo{person}{Norman Mu}, {and}
  \bibinfo{person}{David Wagner}.} \bibinfo{year}{2023}\natexlab{}.
\newblock \showarticletitle{Mark my words: Analyzing and evaluating language
  model watermarks}.
\newblock \bibinfo{journal}{\emph{arXiv preprint arXiv:2312.00273}}
  (\bibinfo{year}{2023}).
\newblock


\bibitem[Press et~al\mbox{.}(2024)]%
        {press2024citeme}
\bibfield{author}{\bibinfo{person}{Ori Press}, \bibinfo{person}{Andreas
  Hochlehnert}, \bibinfo{person}{Ameya Prabhu}, \bibinfo{person}{Vishaal
  Udandarao}, \bibinfo{person}{Ofir Press}, {and} \bibinfo{person}{Matthias
  Bethge}.} \bibinfo{year}{2024}\natexlab{}.
\newblock \showarticletitle{CiteME: Can Language Models Accurately Cite
  Scientific Claims?}
\newblock \bibinfo{journal}{\emph{arXiv preprint arXiv:2407.12861}}
  (\bibinfo{year}{2024}).
\newblock


\bibitem[Pruthi et~al\mbox{.}(2020)]%
        {pruthi2020estimating}
\bibfield{author}{\bibinfo{person}{Garima Pruthi}, \bibinfo{person}{Frederick
  Liu}, \bibinfo{person}{Satyen Kale}, {and} \bibinfo{person}{Mukund
  Sundararajan}.} \bibinfo{year}{2020}\natexlab{}.
\newblock \showarticletitle{Estimating training data influence by tracing
  gradient descent}.
\newblock \bibinfo{journal}{\emph{Advances in Neural Information Processing
  Systems}}  \bibinfo{volume}{33} (\bibinfo{year}{2020}),
  \bibinfo{pages}{19920--19930}.
\newblock


\bibitem[Prystawski et~al\mbox{.}(2023)]%
        {prystawski2023think}
\bibfield{author}{\bibinfo{person}{Ben Prystawski}, \bibinfo{person}{Michael
  Li}, {and} \bibinfo{person}{Noah Goodman}.} \bibinfo{year}{2023}\natexlab{}.
\newblock \showarticletitle{Why think step by step? reasoning emerges from the
  locality of experience}.
\newblock \bibinfo{journal}{\emph{Advances in Neural Information Processing
  Systems}}  \bibinfo{volume}{36} (\bibinfo{year}{2023}),
  \bibinfo{pages}{70926--70947}.
\newblock


\bibitem[Rafailov et~al\mbox{.}(2024)]%
        {rafailov2024direct}
\bibfield{author}{\bibinfo{person}{Rafael Rafailov}, \bibinfo{person}{Archit
  Sharma}, \bibinfo{person}{Eric Mitchell}, \bibinfo{person}{Christopher~D
  Manning}, {et~al\mbox{.}}} \bibinfo{year}{2024}\natexlab{}.
\newblock \showarticletitle{Direct preference optimization: Your language model
  is secretly a reward model}.
\newblock \bibinfo{journal}{\emph{Advances in Neural Information Processing
  Systems}}  \bibinfo{volume}{36} (\bibinfo{year}{2024}).
\newblock


\bibitem[Rastogi et~al\mbox{.}(2025)]%
        {rastogi2025stamp}
\bibfield{author}{\bibinfo{person}{Saksham Rastogi}, \bibinfo{person}{Pratyush
  Maini}, {and} \bibinfo{person}{Danish Pruthi}.}
  \bibinfo{year}{2025}\natexlab{}.
\newblock \showarticletitle{STAMP Your Content: Proving Dataset Membership via
  Watermarked Rephrasings}.
\newblock \bibinfo{journal}{\emph{arXiv preprint arXiv:2504.13416}}
  (\bibinfo{year}{2025}).
\newblock


\bibitem[Reddy et~al\mbox{.}(2023)]%
        {reddy2023smartbook}
\bibfield{author}{\bibinfo{person}{Revanth~Gangi Reddy},
  \bibinfo{person}{Daniel Lee}, \bibinfo{person}{Yi~R Fung},
  \bibinfo{person}{Khanh~Duy Nguyen}, {et~al\mbox{.}}}
  \bibinfo{year}{2023}\natexlab{}.
\newblock \showarticletitle{Smartbook: Ai-assisted situation report generation
  for intelligence analysts}.
\newblock \bibinfo{journal}{\emph{arXiv preprint arXiv:2303.14337}}
  (\bibinfo{year}{2023}).
\newblock


\bibitem[Rodriguez et~al\mbox{.}(2022)]%
        {rodriguez-etal-2022-cross}
\bibfield{author}{\bibinfo{person}{Juan~Diego Rodriguez}, \bibinfo{person}{Todd
  Hay}, \bibinfo{person}{David Gros}, \bibinfo{person}{Zain Shamsi}, {and}
  \bibinfo{person}{Ravi Srinivasan}.} \bibinfo{year}{2022}\natexlab{}.
\newblock \showarticletitle{Cross-Domain Detection of {GPT}-2-Generated
  Technical Text}. In \bibinfo{booktitle}{\emph{Proceedings of the 2022
  Conference of the North American Chapter of the Association for Computational
  Linguistics: Human Language Technologies}}.
\newblock


\bibitem[Roziere et~al\mbox{.}(2023)]%
        {roziere2023code}
\bibfield{author}{\bibinfo{person}{Baptiste Roziere}, \bibinfo{person}{Jonas
  Gehring}, \bibinfo{person}{Fabian Gloeckle}, \bibinfo{person}{Sten Sootla},
  \bibinfo{person}{Itai Gat}, \bibinfo{person}{Xiaoqing~Ellen Tan},
  \bibinfo{person}{Yossi Adi}, \bibinfo{person}{Jingyu Liu}, {et~al\mbox{.}}}
  \bibinfo{year}{2023}\natexlab{}.
\newblock \showarticletitle{Code llama: Open foundation models for code}.
\newblock \bibinfo{journal}{\emph{arXiv preprint arXiv:2308.12950}}
  (\bibinfo{year}{2023}).
\newblock


\bibitem[Sadasivan et~al\mbox{.}(2023)]%
        {sadasivan2023can}
\bibfield{author}{\bibinfo{person}{Vinu~Sankar Sadasivan},
  \bibinfo{person}{Aounon Kumar}, \bibinfo{person}{Sriram Balasubramanian},
  \bibinfo{person}{Wenxiao Wang}, {and} \bibinfo{person}{Soheil Feizi}.}
  \bibinfo{year}{2023}\natexlab{}.
\newblock \showarticletitle{Can ai-generated text be reliably detected?}
\newblock \bibinfo{journal}{\emph{arXiv preprint arXiv:2303.11156}}
  (\bibinfo{year}{2023}).
\newblock


\bibitem[Sancheti et~al\mbox{.}(2024)]%
        {sancheti2024post}
\bibfield{author}{\bibinfo{person}{Abhilasha Sancheti},
  \bibinfo{person}{Koustava Goswami}, {and} \bibinfo{person}{Balaji~Vasan
  Srinivasan}.} \bibinfo{year}{2024}\natexlab{}.
\newblock \showarticletitle{Post-Hoc Answer Attribution for Grounded and
  Trustworthy Long Document Comprehension: Task, Insights, and Challenges}.
\newblock \bibinfo{journal}{\emph{arXiv:2406.06938}} (\bibinfo{year}{2024}).
\newblock


\bibitem[Sander et~al\mbox{.}(2024)]%
        {sander2024watermarking}
\bibfield{author}{\bibinfo{person}{Tom Sander}, \bibinfo{person}{Pierre
  Fernandez}, \bibinfo{person}{Alain Durmus}, \bibinfo{person}{Matthijs Douze},
  {and} \bibinfo{person}{Teddy Furon}.} \bibinfo{year}{2024}\natexlab{}.
\newblock \showarticletitle{Watermarking Makes Language Models Radioactive}.
\newblock \bibinfo{journal}{\emph{arXiv preprint arXiv:2402.14904}}
  (\bibinfo{year}{2024}).
\newblock


\bibitem[Saxena et~al\mbox{.}(2024)]%
        {saxena2024attribution}
\bibfield{author}{\bibinfo{person}{Yash Saxena}, \bibinfo{person}{Deepa
  Tilwani}, \bibinfo{person}{Ali Mohammadi}, \bibinfo{person}{Edward Raff},
  \bibinfo{person}{Amit Sheth}, \bibinfo{person}{Srinivasan Parthasarathy},
  {and} \bibinfo{person}{Manas Gaur}.} \bibinfo{year}{2024}\natexlab{}.
\newblock \showarticletitle{Attribution in Scientific Literature: New Benchmark
  and Methods}.
\newblock \bibinfo{journal}{\emph{arXiv preprint arXiv:2405.02228}}
  (\bibinfo{year}{2024}).
\newblock


\bibitem[Schioppa et~al\mbox{.}(2021)]%
        {schioppa2021scalinginfluencefunctions}
\bibfield{author}{\bibinfo{person}{Andrea Schioppa}, \bibinfo{person}{Polina
  Zablotskaia}, \bibinfo{person}{David Vilar}, {and} \bibinfo{person}{Artem
  Sokolov}.} \bibinfo{year}{2021}\natexlab{}.
\newblock \bibinfo{title}{Scaling Up Influence Functions}.
\newblock
\showeprint[arxiv]{2112.03052}~[cs.LG]
\urldef\tempurl%
\url{https://arxiv.org/abs/2112.03052}
\showURL{%
\tempurl}


\bibitem[Shazeer et~al\mbox{.}(2017)]%
        {shazeer2017outrageously}
\bibfield{author}{\bibinfo{person}{Noam Shazeer}, \bibinfo{person}{Azalia
  Mirhoseini}, \bibinfo{person}{Krzysztof Maziarz}, \bibinfo{person}{Andy
  Davis}, {et~al\mbox{.}}} \bibinfo{year}{2017}\natexlab{}.
\newblock \showarticletitle{Outrageously Large Neural Networks: The
  Sparsely-Gated Mixture-of-Experts Layer}.
\newblock \bibinfo{journal}{\emph{arXiv preprint arXiv:1701.06538}}
  (\bibinfo{year}{2017}).
\newblock


\bibitem[Shi et~al\mbox{.}(2023)]%
        {shi2023detecting}
\bibfield{author}{\bibinfo{person}{Weijia Shi}, \bibinfo{person}{Anirudh
  Ajith}, \bibinfo{person}{Mengzhou Xia}, \bibinfo{person}{Yangsibo Huang},
  \bibinfo{person}{Daogao Liu}, \bibinfo{person}{Terra Blevins},
  \bibinfo{person}{Danqi Chen}, {and} \bibinfo{person}{Luke Zettlemoyer}.}
  \bibinfo{year}{2023}\natexlab{}.
\newblock \showarticletitle{Detecting pretraining data from large language
  models}.
\newblock \bibinfo{journal}{\emph{arXiv preprint arXiv:2310.16789}}
  (\bibinfo{year}{2023}).
\newblock


\bibitem[Simonds et~al\mbox{.}(2024)]%
        {Simonds2024MoDEM}
\bibfield{author}{\bibinfo{person}{Toby Simonds}, \bibinfo{person}{Kemal
  Kurniawan}, {and} \bibinfo{person}{Jey~Han Lau}.}
  \bibinfo{year}{2024}\natexlab{}.
\newblock \showarticletitle{MoDEM: Mixture of Domain Expert Models}.
\newblock \bibinfo{journal}{\emph{arXiv preprint arXiv:2410.07490}}
  (\bibinfo{year}{2024}).
\newblock
\href{https://doi.org/10.48550/arXiv.2410.07490}{doi:\nolinkurl{10.48550/arXiv.2410.07490}}


\bibitem[Singh et~al\mbox{.}(2024)]%
        {singh2024rethinking}
\bibfield{author}{\bibinfo{person}{Chandan Singh},
  \bibinfo{person}{Jeevana~Priya Inala}, \bibinfo{person}{Michel Galley},
  \bibinfo{person}{Rich Caruana}, {and} \bibinfo{person}{Jianfeng Gao}.}
  \bibinfo{year}{2024}\natexlab{}.
\newblock \showarticletitle{Rethinking interpretability in the era of large
  language models}.
\newblock \bibinfo{journal}{\emph{arXiv preprint arXiv:2402.01761}}
  (\bibinfo{year}{2024}).
\newblock


\bibitem[Slobodkin et~al\mbox{.}(2024)]%
        {slobodkin2024attribute}
\bibfield{author}{\bibinfo{person}{Aviv Slobodkin}, \bibinfo{person}{Eran
  Hirsch}, \bibinfo{person}{Arie Cattan}, \bibinfo{person}{Tal Schuster}, {and}
  \bibinfo{person}{Ido Dagan}.} \bibinfo{year}{2024}\natexlab{}.
\newblock \showarticletitle{Attribute First, then Generate:
  Locally-attributable Grounded Text Generation}.
\newblock \bibinfo{journal}{\emph{arXiv preprint arXiv:2403.17104}}
  (\bibinfo{year}{2024}).
\newblock


\bibitem[Sun et~al\mbox{.}(2023)]%
        {sun2023towards}
\bibfield{author}{\bibinfo{person}{Hao Sun}, \bibinfo{person}{Hengyi Cai},
  \bibinfo{person}{Bo Wang}, \bibinfo{person}{Yingyan Hou},
  \bibinfo{person}{Xiaochi Wei}, \bibinfo{person}{Shuaiqiang Wang},
  \bibinfo{person}{Yan Zhang}, {and} \bibinfo{person}{Dawei Yin}.}
  \bibinfo{year}{2023}\natexlab{}.
\newblock \showarticletitle{Towards verifiable text generation with evolving
  memory and self-reflection}.
\newblock \bibinfo{journal}{\emph{arXiv preprint arXiv:2312.09075}}
  (\bibinfo{year}{2023}).
\newblock


\bibitem[Tian et~al\mbox{.}(2024)]%
        {tian2024forget}
\bibfield{author}{\bibinfo{person}{Bozhong Tian}, \bibinfo{person}{Xiaozhuan
  Liang}, \bibinfo{person}{Siyuan Cheng}, \bibinfo{person}{Qingbin Liu},
  {et~al\mbox{.}}} \bibinfo{year}{2024}\natexlab{}.
\newblock \showarticletitle{To forget or not? towards practical knowledge
  unlearning for large language models}.
\newblock \bibinfo{journal}{\emph{arXiv preprint arXiv:2407.01920}}
  (\bibinfo{year}{2024}).
\newblock


\bibitem[Tilwani et~al\mbox{.}(2024)]%
        {tilwani2024reasons}
\bibfield{author}{\bibinfo{person}{Deepa Tilwani}, \bibinfo{person}{Yash
  Saxena}, \bibinfo{person}{Ali Mohammadi}, \bibinfo{person}{Edward Raff},
  {et~al\mbox{.}}} \bibinfo{year}{2024}\natexlab{}.
\newblock \showarticletitle{REASONS: A benchmark for REtrieval and Automated
  citationS Of scieNtific Sentences using Public and Proprietary LLMs}.
\newblock \bibinfo{journal}{\emph{arXiv:2405.02228}} (\bibinfo{year}{2024}).
\newblock


\bibitem[Todd et~al\mbox{.}(2024)]%
        {DBLP:conf/iclr/ToddLSMWB24}
\bibfield{author}{\bibinfo{person}{Eric Todd}, \bibinfo{person}{Millicent~L.
  Li}, \bibinfo{person}{Arnab~Sen Sharma}, \bibinfo{person}{Aaron Mueller},
  \bibinfo{person}{Byron~C. Wallace}, {and} \bibinfo{person}{David Bau}.}
  \bibinfo{year}{2024}\natexlab{}.
\newblock \showarticletitle{Function Vectors in Large Language Models}. In
  \bibinfo{booktitle}{\emph{{ICLR} 2024, Vienna, Austria, May 7-11, 2024}}.
\newblock


\bibitem[Venditti et~al\mbox{.}(2024)]%
        {venditti2024enhancing}
\bibfield{author}{\bibinfo{person}{Davide Venditti},
  \bibinfo{person}{Elena~Sofia Ruzzetti}, \bibinfo{person}{Giancarlo~A
  Xompero}, \bibinfo{person}{Cristina Giannone}, \bibinfo{person}{Andrea
  Favalli}, {et~al\mbox{.}}} \bibinfo{year}{2024}\natexlab{}.
\newblock \showarticletitle{Enhancing Data Privacy in Large Language Models
  through Private Association Editing}.
\newblock \bibinfo{journal}{\emph{arXiv preprint arXiv:2406.18221}}
  (\bibinfo{year}{2024}).
\newblock


\bibitem[Verma et~al\mbox{.}(2023)]%
        {verma2023ghostbuster}
\bibfield{author}{\bibinfo{person}{Vivek Verma}, \bibinfo{person}{Eve Fleisig},
  \bibinfo{person}{Nicholas Tomlin}, {and} \bibinfo{person}{Dan Klein}.}
  \bibinfo{year}{2023}\natexlab{}.
\newblock \bibinfo{title}{Ghostbuster: Detecting Text Ghostwritten by Large
  Language Models}.
\newblock
\showeprint[arxiv]{2305.15047}~[cs.CL]


\bibitem[Von~Oswald et~al\mbox{.}(2023)]%
        {von2023transformers}
\bibfield{author}{\bibinfo{person}{Johannes Von~Oswald},
  \bibinfo{person}{Eyvind Niklasson}, \bibinfo{person}{Ettore Randazzo},
  \bibinfo{person}{Jo{\~a}o Sacramento}, {et~al\mbox{.}}}
  \bibinfo{year}{2023}\natexlab{}.
\newblock \showarticletitle{Transformers learn in-context by gradient descent}.
  In \bibinfo{booktitle}{\emph{International Conference on Machine Learning}}.
  PMLR, \bibinfo{pages}{35151--35174}.
\newblock


\bibitem[Wang et~al\mbox{.}(2023b)]%
        {wang2023wasa}
\bibfield{author}{\bibinfo{person}{Jingtan Wang}, \bibinfo{person}{Xinyang Lu},
  \bibinfo{person}{Zitong Zhao}, \bibinfo{person}{Zhongxiang Dai},
  \bibinfo{person}{Chuan-Sheng Foo}, {et~al\mbox{.}}}
  \bibinfo{year}{2023}\natexlab{b}.
\newblock \showarticletitle{WASA: Watermark-based source attribution for large
  language model-generated data}.
\newblock \bibinfo{journal}{\emph{arXiv preprint arXiv:2310.00646}}
  (\bibinfo{year}{2023}).
\newblock


\bibitem[Wang et~al\mbox{.}(2023c)]%
        {wang2023source}
\bibfield{author}{\bibinfo{person}{Jingtan Wang}, \bibinfo{person}{Xinyang Lu},
  \bibinfo{person}{Zitong Zhao}, \bibinfo{person}{Zhongxiang Dai},
  \bibinfo{person}{Chuan-Sheng Foo}, \bibinfo{person}{See-Kiong Ng}, {and}
  \bibinfo{person}{Bryan Kian~Hsiang Low}.} \bibinfo{year}{2023}\natexlab{c}.
\newblock \showarticletitle{Source Attribution for Large Language
  Model-Generated Data}.
\newblock \bibinfo{journal}{\emph{arXiv preprint arXiv:2310.00646}}
  (\bibinfo{year}{2023}).
\newblock


\bibitem[Wang et~al\mbox{.}(2023e)]%
        {DBLP:conf/iclr/WangVCSS23}
\bibfield{author}{\bibinfo{person}{Kevin~Ro Wang}, \bibinfo{person}{Alexandre
  Variengien}, \bibinfo{person}{Arthur Conmy}, \bibinfo{person}{Buck
  Shlegeris}, {and} \bibinfo{person}{Jacob Steinhardt}.}
  \bibinfo{year}{2023}\natexlab{e}.
\newblock \showarticletitle{Interpretability in the Wild: a Circuit for
  Indirect Object Identification in {GPT-2} Small}. In
  \bibinfo{booktitle}{\emph{{ICLR} 2023, Kigali, Rwanda, May 1-5, 2023}}.
\newblock


\bibitem[Wang et~al\mbox{.}(2024a)]%
        {DBLP:conf/emnlp/WangYXQD00GJX0C24}
\bibfield{author}{\bibinfo{person}{Mengru Wang}, \bibinfo{person}{Yunzhi Yao},
  \bibinfo{person}{Ziwen Xu}, \bibinfo{person}{Shuofei Qiao}, {et~al\mbox{.}}}
  \bibinfo{year}{2024}\natexlab{a}.
\newblock \showarticletitle{Knowledge Mechanisms in Large Language Models: {A}
  Survey and Perspective}. In \bibinfo{booktitle}{\emph{Findings of the
  Association for Computational Linguistics: {EMNLP} 2024}}.
\newblock


\bibitem[Wang et~al\mbox{.}(2024b)]%
        {wang-etal-2024-detoxifying}
\bibfield{author}{\bibinfo{person}{Mengru Wang}, \bibinfo{person}{Ningyu
  Zhang}, \bibinfo{person}{Ziwen Xu}, \bibinfo{person}{Zekun Xi},
  {et~al\mbox{.}}} \bibinfo{year}{2024}\natexlab{b}.
\newblock \showarticletitle{Detoxifying Large Language Models via Knowledge
  Editing}. In \bibinfo{booktitle}{\emph{Proceedings of the 62nd Annual Meeting
  of the Association for Computational Linguistics}}.
\newblock


\bibitem[Wang et~al\mbox{.}(2023a)]%
        {wang2023detectgptsc}
\bibfield{author}{\bibinfo{person}{Rongsheng Wang}, \bibinfo{person}{Qi Li},
  {and} \bibinfo{person}{Sihong Xie}.} \bibinfo{year}{2023}\natexlab{a}.
\newblock \bibinfo{title}{DetectGPT-SC: Improving Detection of Text Generated
  by Large Language Models through Self-Consistency with Masked Predictions}.
\newblock
\showeprint[arxiv]{2310.14479}~[cs.CL]


\bibitem[Wang et~al\mbox{.}(2025c)]%
        {wang2025medcite}
\bibfield{author}{\bibinfo{person}{Xiao Wang}, \bibinfo{person}{Mengjue Tan},
  \bibinfo{person}{Qiao Jin}, \bibinfo{person}{Guangzhi Xiong},
  \bibinfo{person}{Yu Hu}, \bibinfo{person}{Aidong Zhang},
  \bibinfo{person}{Zhiyong Lu}, {and} \bibinfo{person}{Minjia Zhang}.}
  \bibinfo{year}{2025}\natexlab{c}.
\newblock \showarticletitle{MedCite: Can Language Models Generate Verifiable
  Text for Medicine?}
\newblock \bibinfo{journal}{\emph{arXiv preprint arXiv:2506.06605}}
  (\bibinfo{year}{2025}).
\newblock


\bibitem[Wang et~al\mbox{.}(2023d)]%
        {wang2023m4}
\bibfield{author}{\bibinfo{person}{Yuxia Wang}, \bibinfo{person}{Jonibek
  Mansurov}, \bibinfo{person}{Petar Ivanov}, \bibinfo{person}{Jinyan Su},
  {et~al\mbox{.}}} \bibinfo{year}{2023}\natexlab{d}.
\newblock \bibinfo{title}{M4: Multi-generator, Multi-domain, and Multi-lingual
  Black-Box Machine-Generated Text Detection}.
\newblock
\showeprint[arxiv]{2305.14902}~[cs.CL]


\bibitem[Wang et~al\mbox{.}(2025b)]%
        {wang2025tradeoffsynergyversatilesymbiotic}
\bibfield{author}{\bibinfo{person}{Yidan Wang}, \bibinfo{person}{Yubing Ren},
  \bibinfo{person}{Yanan Cao}, {and} \bibinfo{person}{Binxing Fang}.}
  \bibinfo{year}{2025}\natexlab{b}.
\newblock \bibinfo{title}{From Trade-off to Synergy: A Versatile Symbiotic
  Watermarking Framework for Large Language Models}.
\newblock
\showeprint[arxiv]{2505.09924}~[cs.CL]
\urldef\tempurl%
\url{https://arxiv.org/abs/2505.09924}
\showURL{%
\tempurl}


\bibitem[Wang et~al\mbox{.}(2025a)]%
        {wang2025morphmarkflexibleadaptivewatermarking}
\bibfield{author}{\bibinfo{person}{Zongqi Wang}, \bibinfo{person}{Tianle Gu},
  \bibinfo{person}{Baoyuan Wu}, {and} \bibinfo{person}{Yujiu Yang}.}
  \bibinfo{year}{2025}\natexlab{a}.
\newblock \bibinfo{title}{MorphMark: Flexible Adaptive Watermarking for Large
  Language Models}.
\newblock
\showeprint[arxiv]{2505.11541}~[cs.CR]
\urldef\tempurl%
\url{https://arxiv.org/abs/2505.11541}
\showURL{%
\tempurl}


\bibitem[Wei et~al\mbox{.}(2022)]%
        {wei2022emergent}
\bibfield{author}{\bibinfo{person}{Jason Wei}, \bibinfo{person}{Yi Tay},
  \bibinfo{person}{Rishi Bommasani}, \bibinfo{person}{Colin Raffel},
  \bibinfo{person}{Barret Zoph}, \bibinfo{person}{Sebastian Borgeaud},
  \bibinfo{person}{Dani Yogatama}, \bibinfo{person}{Maarten Bosma},
  \bibinfo{person}{Denny Zhou}, {et~al\mbox{.}}}
  \bibinfo{year}{2022}\natexlab{}.
\newblock \showarticletitle{Emergent abilities of large language models}.
\newblock \bibinfo{journal}{\emph{arXiv preprint arXiv:2206.07682}}
  (\bibinfo{year}{2022}).
\newblock


\bibitem[Wei et~al\mbox{.}(2025)]%
        {DBLP:conf/coling/WeiDPDSC25}
\bibfield{author}{\bibinfo{person}{Zihao Wei}, \bibinfo{person}{Jingcheng
  Deng}, \bibinfo{person}{Liang Pang}, \bibinfo{person}{Hanxing Ding},
  \bibinfo{person}{Huawei Shen}, {and} \bibinfo{person}{Xueqi Cheng}.}
  \bibinfo{year}{2025}\natexlab{}.
\newblock \showarticletitle{MLaKE: Multilingual Knowledge Editing Benchmark for
  Large Language Models}. In \bibinfo{booktitle}{\emph{{COLING} 2025, Abu
  Dhabi, UAE, January 19-24, 2025}}.
\newblock


\bibitem[Wei et~al\mbox{.}(2024)]%
        {wei2024stable}
\bibfield{author}{\bibinfo{person}{Zihao Wei}, \bibinfo{person}{Liang Pang},
  \bibinfo{person}{Hanxing Ding}, \bibinfo{person}{Jingcheng Deng},
  \bibinfo{person}{Huawei Shen}, {and} \bibinfo{person}{Xueqi Cheng}.}
  \bibinfo{year}{2024}\natexlab{}.
\newblock \bibinfo{title}{Stable Knowledge Editing in Large Language Models}.
\newblock
\showeprint[arxiv]{2402.13048}~[cs.CL]


\bibitem[Wu et~al\mbox{.}(2025)]%
        {wu2025wrotethiskeyzeroshot}
\bibfield{author}{\bibinfo{person}{Junchao Wu}, \bibinfo{person}{Runzhe Zhan},
  \bibinfo{person}{Derek~F. Wong}, \bibinfo{person}{Shu Yang},
  \bibinfo{person}{Xuebo Liu}, \bibinfo{person}{Lidia~S. Chao}, {and}
  \bibinfo{person}{Min Zhang}.} \bibinfo{year}{2025}\natexlab{}.
\newblock \bibinfo{title}{Who Wrote This? The Key to Zero-Shot LLM-Generated
  Text Detection Is GECScore}.
\newblock
\showeprint[arxiv]{2405.04286}~[cs.CL]


\bibitem[Wu et~al\mbox{.}(2024)]%
        {wu2024enhancing}
\bibfield{author}{\bibinfo{person}{Kangxi Wu}, \bibinfo{person}{Liang Pang},
  \bibinfo{person}{Huawei Shen}, {and} \bibinfo{person}{Xueqi Cheng}.}
  \bibinfo{year}{2024}\natexlab{}.
\newblock \showarticletitle{Enhancing Training Data Attribution for Large
  Language Models with Fitting Error Consideration}. In
  \bibinfo{booktitle}{\emph{Proceedings of the 2024 Conference on Empirical
  Methods in Natural Language Processing}}. \bibinfo{pages}{14131--14143}.
\newblock


\bibitem[Wu et~al\mbox{.}(2023b)]%
        {wu2023llmdetpartylargelanguage}
\bibfield{author}{\bibinfo{person}{Kangxi Wu}, \bibinfo{person}{Liang Pang},
  \bibinfo{person}{Huawei Shen}, \bibinfo{person}{Xueqi Cheng}, {and}
  \bibinfo{person}{Tat-Seng Chua}.} \bibinfo{year}{2023}\natexlab{b}.
\newblock \bibinfo{title}{LLMDet: A Third Party Large Language Models Generated
  Text Detection Tool}.
\newblock
\showeprint[arxiv]{2305.15004}~[cs.CL]
\urldef\tempurl%
\url{https://arxiv.org/abs/2305.15004}
\showURL{%
\tempurl}


\bibitem[Wu et~al\mbox{.}(2023a)]%
        {wu-etal-2023-depn}
\bibfield{author}{\bibinfo{person}{Xinwei Wu}, \bibinfo{person}{Junzhuo Li},
  \bibinfo{person}{Minghui Xu}, {et~al\mbox{.}}}
  \bibinfo{year}{2023}\natexlab{a}.
\newblock \showarticletitle{{DEPN}: Detecting and Editing Privacy Neurons in
  Pretrained Language Models}. In \bibinfo{booktitle}{\emph{Proceedings of the
  2023 Conference on Empirical Methods in Natural Language Processing}}.
\newblock


\bibitem[Xia et~al\mbox{.}(2024)]%
        {xia2024ground}
\bibfield{author}{\bibinfo{person}{Sirui Xia}, \bibinfo{person}{Xintao Wang},
  \bibinfo{person}{Jiaqing Liang}, \bibinfo{person}{Yifei Zhang},
  \bibinfo{person}{Weikang Zhou}, {et~al\mbox{.}}}
  \bibinfo{year}{2024}\natexlab{}.
\newblock \showarticletitle{Ground Every Sentence: Improving
  Retrieval-Augmented LLMs with Interleaved Reference-Claim Generation}.
\newblock \bibinfo{journal}{\emph{arXiv preprint arXiv:2407.01796}}
  (\bibinfo{year}{2024}).
\newblock


\bibitem[Xu et~al\mbox{.}(2025a)]%
        {xu2025reversephysicianairelationshipfullprocess}
\bibfield{author}{\bibinfo{person}{Shicheng Xu}, \bibinfo{person}{Xin Huang},
  \bibinfo{person}{Zihao Wei}, \bibinfo{person}{Liang Pang},
  \bibinfo{person}{Huawei Shen}, {and} \bibinfo{person}{Xueqi Cheng}.}
  \bibinfo{year}{2025}\natexlab{a}.
\newblock \bibinfo{title}{Reverse Physician-AI Relationship: Full-process
  Clinical Diagnosis Driven by a Large Language Model}.
\newblock
\showeprint[arxiv]{2508.10492}~[cs.AI]


\bibitem[Xu et~al\mbox{.}(2024b)]%
        {xu2024search}
\bibfield{author}{\bibinfo{person}{Shicheng Xu}, \bibinfo{person}{Liang Pang},
  \bibinfo{person}{Huawei Shen}, \bibinfo{person}{Xueqi Cheng},
  {et~al\mbox{.}}} \bibinfo{year}{2024}\natexlab{b}.
\newblock \showarticletitle{Search-in-the-chain: Interactively enhancing large
  language models with search for knowledge-intensive tasks}. In
  \bibinfo{booktitle}{\emph{Proceedings of the ACM Web Conference 2024}}.
\newblock


\bibitem[Xu et~al\mbox{.}(2024a)]%
        {xu2024aliice}
\bibfield{author}{\bibinfo{person}{Yilong Xu}, \bibinfo{person}{Jinhua Gao},
  \bibinfo{person}{Xiaoming Yu}, \bibinfo{person}{Baolong Bi},
  \bibinfo{person}{Huawei Shen}, {and} \bibinfo{person}{Xueqi Cheng}.}
  \bibinfo{year}{2024}\natexlab{a}.
\newblock \showarticletitle{ALiiCE: Evaluating Positional Fine-grained Citation
  Generation}.
\newblock \bibinfo{journal}{\emph{arXiv preprint arXiv:2406.13375}}
  (\bibinfo{year}{2024}).
\newblock


\bibitem[Xu et~al\mbox{.}(2025c)]%
        {xu2025markyourllm}
\bibfield{author}{\bibinfo{person}{Yijie Xu}, \bibinfo{person}{Hui Xiong},
  \bibinfo{person}{Aiwei Liu}, \bibinfo{person}{Lijie Wen}, {and}
  \bibinfo{person}{Xuming Hu}.} \bibinfo{year}{2025}\natexlab{c}.
\newblock \bibinfo{title}{Mark Your LLM: Detecting the Misuse of Open-Source
  Large Language Models via Watermarking}.
\newblock
\showeprint[arxiv]{2503.04636}~[cs.CL]
\urldef\tempurl%
\url{https://arxiv.org/abs/2503.04636}
\showURL{%
\tempurl}


\bibitem[Xu et~al\mbox{.}(2025b)]%
        {DBLP:journals/corr/abs-2504-15133}
\bibfield{author}{\bibinfo{person}{Ziwen Xu}, \bibinfo{person}{Shuxun Wang},
  \bibinfo{person}{Kewei Xu}, \bibinfo{person}{Haoming Xu},
  \bibinfo{person}{Mengru Wang}, \bibinfo{person}{Xinle Deng},
  \bibinfo{person}{Yunzhi Yao}, {et~al\mbox{.}}}
  \bibinfo{year}{2025}\natexlab{b}.
\newblock \showarticletitle{EasyEdit2: An Easy-to-use Steering Framework for
  Editing Large Language Models}.
\newblock \bibinfo{journal}{\emph{CoRR}}  \bibinfo{volume}{abs/2504.15133}
  (\bibinfo{year}{2025}).
\newblock
\showeprint[arXiv]{2504.15133}


\bibitem[Yaggel et~al\mbox{.}(2025)]%
        {Yaggel2025MoEMLoRA}
\bibfield{author}{\bibinfo{person}{Ken Yaggel}, \bibinfo{person}{Eyal German},
  {and} \bibinfo{person}{Aviel Ben Siman~Tov}.}
  \bibinfo{year}{2025}\natexlab{}.
\newblock \showarticletitle{MoE-MLoRA for Multi-Domain CTR Prediction:
  Efficient Adaptation with Expert Specialization}.
\newblock \bibinfo{journal}{\emph{arXiv preprint arXiv:2506.07563}}
  (\bibinfo{year}{2025}).
\newblock
\href{https://doi.org/10.48550/arXiv.2506.07563}{doi:\nolinkurl{10.48550/arXiv.2506.07563}}


\bibitem[Yang et~al\mbox{.}(2024)]%
        {yang2024qwen2}
\bibfield{author}{\bibinfo{person}{An Yang}, \bibinfo{person}{Baosong Yang},
  \bibinfo{person}{Beichen Zhang}, \bibinfo{person}{Binyuan Hui},
  \bibinfo{person}{Bo Zheng}, \bibinfo{person}{Bowen Yu},
  \bibinfo{person}{Chengyuan Li}, \bibinfo{person}{Dayiheng Liu},
  \bibinfo{person}{Fei Huang}, \bibinfo{person}{Haoran Wei}, {et~al\mbox{.}}}
  \bibinfo{year}{2024}\natexlab{}.
\newblock \showarticletitle{Qwen2. 5 technical report}.
\newblock \bibinfo{journal}{\emph{arXiv preprint arXiv:2412.15115}}
  (\bibinfo{year}{2024}).
\newblock


\bibitem[Yang et~al\mbox{.}(2025a)]%
        {yang2025gmvaluatorsimilaritybaseddatavaluation}
\bibfield{author}{\bibinfo{person}{Jiaxi Yang}, \bibinfo{person}{Wenglong
  Deng}, \bibinfo{person}{Benlin Liu}, \bibinfo{person}{Yangsibo Huang},
  \bibinfo{person}{James Zou}, {and} \bibinfo{person}{Xiaoxiao Li}.}
  \bibinfo{year}{2025}\natexlab{a}.
\newblock \bibinfo{title}{GMValuator: Similarity-based Data Valuation for
  Generative Models}.
\newblock
\showeprint[arxiv]{2304.10701}~[cs.CV]
\urldef\tempurl%
\url{https://arxiv.org/abs/2304.10701}
\showURL{%
\tempurl}


\bibitem[Yang et~al\mbox{.}(2023a)]%
        {yang2023watermarkingtextgeneratedblackbox}
\bibfield{author}{\bibinfo{person}{Xi Yang}, \bibinfo{person}{Kejiang Chen},
  \bibinfo{person}{Weiming Zhang}, \bibinfo{person}{Chang Liu},
  \bibinfo{person}{Yuang Qi}, \bibinfo{person}{Jie Zhang}, \bibinfo{person}{Han
  Fang}, {and} \bibinfo{person}{Nenghai Yu}.} \bibinfo{year}{2023}\natexlab{a}.
\newblock \bibinfo{title}{Watermarking Text Generated by Black-Box Language
  Models}.
\newblock
\showeprint[arxiv]{2305.08883}~[cs.CL]
\urldef\tempurl%
\url{https://arxiv.org/abs/2305.08883}
\showURL{%
\tempurl}


\bibitem[Yang et~al\mbox{.}(2023b)]%
        {yang2023zeroshot}
\bibfield{author}{\bibinfo{person}{Xianjun Yang}, \bibinfo{person}{Kexun
  Zhang}, \bibinfo{person}{Haifeng Chen}, \bibinfo{person}{Linda Petzold},
  \bibinfo{person}{William~Yang Wang}, {and} \bibinfo{person}{Wei Cheng}.}
  \bibinfo{year}{2023}\natexlab{b}.
\newblock \bibinfo{title}{Zero-Shot Detection of Machine-Generated Codes}.
\newblock
\showeprint[arxiv]{2310.05103}~[cs.CL]


\bibitem[Yang et~al\mbox{.}(2025b)]%
        {yang2025agentguidesimple}
\bibfield{author}{\bibinfo{person}{Zhongliang Yang}, \bibinfo{person}{Linna
  Zhou}, \bibinfo{person}{Zipei Zhang}, {and} \bibinfo{person}{Kaibo Huang}.}
  \bibinfo{year}{2025}\natexlab{b}.
\newblock \bibinfo{title}{Agent Guide: A Simple Agent Behavioral Watermarking
  Framework}.
\newblock
\showeprint[arxiv]{2504.05871}~[cs.AI]
\urldef\tempurl%
\url{https://arxiv.org/abs/2504.05871}
\showURL{%
\tempurl}


\bibitem[Yao et~al\mbox{.}(2024)]%
        {DBLP:conf/nips/Yao0XWXDC24}
\bibfield{author}{\bibinfo{person}{Yunzhi Yao}, \bibinfo{person}{Ningyu Zhang},
  \bibinfo{person}{Zekun Xi}, \bibinfo{person}{Mengru Wang},
  \bibinfo{person}{Ziwen Xu}, \bibinfo{person}{Shumin Deng}, {and}
  \bibinfo{person}{Huajun Chen}.} \bibinfo{year}{2024}\natexlab{}.
\newblock \showarticletitle{Knowledge Circuits in Pretrained Transformers}. In
  \bibinfo{booktitle}{\emph{NeurIPS 2024, Vancouver, BC, Canada, December 10 -
  15, 2024}}.
\newblock


\bibitem[Ye et~al\mbox{.}(2023)]%
        {ye2023effective}
\bibfield{author}{\bibinfo{person}{Xi Ye}, \bibinfo{person}{Ruoxi Sun},
  \bibinfo{person}{Sercan~{\"O} Arik}, {and} \bibinfo{person}{Tomas Pfister}.}
  \bibinfo{year}{2023}\natexlab{}.
\newblock \showarticletitle{Effective large language model adaptation for
  improved grounding}.
\newblock \bibinfo{journal}{\emph{arXiv preprint arXiv:2311.09533}}
  (\bibinfo{year}{2023}).
\newblock


\bibitem[Yu et~al\mbox{.}(2023b)]%
        {yu2023unlearning}
\bibfield{author}{\bibinfo{person}{Charles Yu}, \bibinfo{person}{Sullam
  Jeoung}, \bibinfo{person}{Anish Kasi}, \bibinfo{person}{Pengfei Yu}, {and}
  \bibinfo{person}{Heng Ji}.} \bibinfo{year}{2023}\natexlab{b}.
\newblock \showarticletitle{Unlearning bias in language models by partitioning
  gradients}. In \bibinfo{booktitle}{\emph{Findings of the Association for
  Computational Linguistics: ACL 2023}}. \bibinfo{pages}{6032--6048}.
\newblock


\bibitem[Yu et~al\mbox{.}(2023a)]%
        {yu2023cheat}
\bibfield{author}{\bibinfo{person}{Peipeng Yu}, \bibinfo{person}{Jiahan Chen},
  \bibinfo{person}{Xuan Feng}, {and} \bibinfo{person}{Zhihua Xia}.}
  \bibinfo{year}{2023}\natexlab{a}.
\newblock \bibinfo{title}{CHEAT: A Large-scale Dataset for Detecting
  ChatGPT-writtEn AbsTracts}.
\newblock
\showeprint[arxiv]{2304.12008}~[cs.CL]


\bibitem[Yu et~al\mbox{.}(2024b)]%
        {yu2024waterseekerpioneeringefficient}
\bibfield{author}{\bibinfo{person}{Philip~S. Yu}, \bibinfo{person}{Aiwei Liu},
  \bibinfo{person}{Lijie Wen}, \bibinfo{person}{Leyi Pan},
  \bibinfo{person}{Yijian Lu}, \bibinfo{person}{Irwin King},
  \bibinfo{person}{Shiyu Huang}, \bibinfo{person}{Zitian Gao}, {and}
  \bibinfo{person}{Yichen Di}.} \bibinfo{year}{2024}\natexlab{b}.
\newblock \bibinfo{title}{WaterSeeker: Pioneering Efficient Detection of
  Watermarked Segments in Large Documents}.
\newblock
\showeprint[arxiv]{2409.05112}~[cs.CL]


\bibitem[Yu et~al\mbox{.}(2024a)]%
        {yu2024mates}
\bibfield{author}{\bibinfo{person}{Zichun Yu}, \bibinfo{person}{Spandan Das},
  {and} \bibinfo{person}{Chenyan Xiong}.} \bibinfo{year}{2024}\natexlab{a}.
\newblock \showarticletitle{Mates: Model-aware data selection for efficient
  pretraining with data influence models}.
\newblock \bibinfo{journal}{\emph{Advances in Neural Information Processing
  Systems}}  \bibinfo{volume}{37} (\bibinfo{year}{2024}),
  \bibinfo{pages}{108735--108759}.
\newblock


\bibitem[Yu et~al\mbox{.}(2025)]%
        {yu2025data}
\bibfield{author}{\bibinfo{person}{Zichun Yu}, \bibinfo{person}{Fei Peng},
  \bibinfo{person}{Jie Lei}, \bibinfo{person}{Arnold Overwijk},
  \bibinfo{person}{Wen-tau Yih}, {and} \bibinfo{person}{Chenyan Xiong}.}
  \bibinfo{year}{2025}\natexlab{}.
\newblock \showarticletitle{Data-efficient pretraining with group-level data
  influence modeling}.
\newblock \bibinfo{journal}{\emph{arXiv preprint arXiv:2502.14709}}
  (\bibinfo{year}{2025}).
\newblock


\bibitem[Yue et~al\mbox{.}(2023)]%
        {yue2023automatic}
\bibfield{author}{\bibinfo{person}{Xiang Yue}, \bibinfo{person}{Boshi Wang},
  \bibinfo{person}{Ziru Chen}, \bibinfo{person}{Kai Zhang}, \bibinfo{person}{Yu
  Su}, {and} \bibinfo{person}{Huan Sun}.} \bibinfo{year}{2023}\natexlab{}.
\newblock \showarticletitle{Automatic evaluation of attribution by large
  language models}.
\newblock \bibinfo{journal}{\emph{arXiv preprint arXiv:2305.06311}}
  (\bibinfo{year}{2023}).
\newblock


\bibitem[Zeng et~al\mbox{.}(2025)]%
        {zeng2025cite}
\bibfield{author}{\bibinfo{person}{Jingying Zeng}, \bibinfo{person}{Hui Liu},
  \bibinfo{person}{Zhenwei Dai}, \bibinfo{person}{Xianfeng Tang},
  \bibinfo{person}{Chen Luo}, {et~al\mbox{.}}} \bibinfo{year}{2025}\natexlab{}.
\newblock \showarticletitle{Cite before you speak: Enhancing context-response
  grounding in e-commerce conversational llm-agents}.
\newblock \bibinfo{journal}{\emph{arXiv preprint arXiv:2503.04830}}
  (\bibinfo{year}{2025}).
\newblock


\bibitem[Zhang et~al\mbox{.}(2024b)]%
        {zhang2024longcite}
\bibfield{author}{\bibinfo{person}{Jiajie Zhang}, \bibinfo{person}{Yushi Bai},
  \bibinfo{person}{Xin Lv}, \bibinfo{person}{Wanjun Gu},
  \bibinfo{person}{Danqing Liu}, \bibinfo{person}{Minhao Zou},
  \bibinfo{person}{Shulin Cao}, \bibinfo{person}{Lei Hou}, {et~al\mbox{.}}}
  \bibinfo{year}{2024}\natexlab{b}.
\newblock \showarticletitle{Longcite: Enabling llms to generate fine-grained
  citations in long-context qa}.
\newblock \bibinfo{journal}{\emph{arXiv preprint arXiv:2409.02897}}
  (\bibinfo{year}{2024}).
\newblock


\bibitem[Zhang et~al\mbox{.}(2025)]%
        {zhang2024citalaw}
\bibfield{author}{\bibinfo{person}{Kepu Zhang}, \bibinfo{person}{Weijie Yu},
  \bibinfo{person}{Sunhao Dai}, {and} \bibinfo{person}{Jun Xu}.}
  \bibinfo{year}{2025}\natexlab{}.
\newblock \showarticletitle{CitaLaw: Enhancing LLM with Citations in Legal
  Domain}.
\newblock \bibinfo{journal}{\emph{Findings of the Association for Computational
  Linguistics: ACL 2025}} (\bibinfo{year}{2025}).
\newblock


\bibitem[Zhang et~al\mbox{.}(2024c)]%
        {zhang2024truthx}
\bibfield{author}{\bibinfo{person}{Shaolei Zhang}, \bibinfo{person}{Tian Yu},
  {and} \bibinfo{person}{Yang Feng}.} \bibinfo{year}{2024}\natexlab{c}.
\newblock \showarticletitle{Truthx: Alleviating hallucinations by editing large
  language models in truthful space}.
\newblock \bibinfo{journal}{\emph{arXiv preprint arXiv:2402.17811}}
  (\bibinfo{year}{2024}).
\newblock


\bibitem[Zhang et~al\mbox{.}(2024a)]%
        {zhang2024towards}
\bibfield{author}{\bibinfo{person}{Weijia Zhang}, \bibinfo{person}{Mohammad
  Aliannejadi}, \bibinfo{person}{Yifei Yuan}, \bibinfo{person}{Jiahuan Pei},
  {et~al\mbox{.}}} \bibinfo{year}{2024}\natexlab{a}.
\newblock \showarticletitle{Towards fine-grained citation evaluation in
  generated text: A comparative analysis of faithfulness metrics}.
\newblock \bibinfo{journal}{\emph{arXiv preprint arXiv:2406.15264}}
  (\bibinfo{year}{2024}).
\newblock


\bibitem[Zhao et~al\mbox{.}(2019)]%
        {zhao2019inferringtrainingdataattributes}
\bibfield{author}{\bibinfo{person}{Benjamin Zi~Hao Zhao},
  \bibinfo{person}{Hassan~Jameel Asghar}, \bibinfo{person}{Raghav Bhaskar},
  {and} \bibinfo{person}{Mohamed~Ali Kaafar}.} \bibinfo{year}{2019}\natexlab{}.
\newblock \bibinfo{title}{On Inferring Training Data Attributes in Machine
  Learning Models}.
\newblock
\showeprint[arxiv]{1908.10558}~[cs.CR]
\urldef\tempurl%
\url{https://arxiv.org/abs/1908.10558}
\showURL{%
\tempurl}


\bibitem[Zhao et~al\mbox{.}(2023b)]%
        {zhao2023explainability}
\bibfield{author}{\bibinfo{person}{Haiyan Zhao}, \bibinfo{person}{Hanjie Chen},
  \bibinfo{person}{Fan Yang}, \bibinfo{person}{Ninghao Liu},
  \bibinfo{person}{Huiqi Deng}, \bibinfo{person}{Hengyi Cai},
  \bibinfo{person}{Shuaiqiang Wang}, \bibinfo{person}{Dawei Yin}, {and}
  \bibinfo{person}{Mengnan Du}.} \bibinfo{year}{2023}\natexlab{b}.
\newblock \bibinfo{title}{Explainability for Large Language Models: A Survey}.
\newblock
\showeprint[arxiv]{2309.01029}~[cs.CL]


\bibitem[Zhao et~al\mbox{.}(2024)]%
        {zhao-etal-2024-defending-large}
\bibfield{author}{\bibinfo{person}{Wei Zhao}, \bibinfo{person}{Zhe Li},
  \bibinfo{person}{Yige Li}, \bibinfo{person}{Ye Zhang}, {and}
  \bibinfo{person}{Jun Sun}.} \bibinfo{year}{2024}\natexlab{}.
\newblock \showarticletitle{Defending Large Language Models Against Jailbreak
  Attacks via Layer-specific Editing}. In \bibinfo{booktitle}{\emph{Findings of
  the Association for Computational Linguistics: EMNLP 2024}}.
\newblock


\bibitem[Zhao et~al\mbox{.}(2023c)]%
        {zhao2023survey}
\bibfield{author}{\bibinfo{person}{Wayne~Xin Zhao}, \bibinfo{person}{Kun Zhou},
  \bibinfo{person}{Junyi Li}, \bibinfo{person}{Tianyi Tang},
  \bibinfo{person}{Xiaolei Wang}, \bibinfo{person}{Yupeng Hou},
  {et~al\mbox{.}}} \bibinfo{year}{2023}\natexlab{c}.
\newblock \bibinfo{title}{A Survey of Large Language Models}.
\newblock
\showeprint[arxiv]{2303.18223}~[cs.CL]


\bibitem[Zhao et~al\mbox{.}(2023a)]%
        {zhao2023provable}
\bibfield{author}{\bibinfo{person}{Xuandong Zhao}, \bibinfo{person}{Prabhanjan
  Ananth}, \bibinfo{person}{Lei Li}, {and} \bibinfo{person}{Yu-Xiang Wang}.}
  \bibinfo{year}{2023}\natexlab{a}.
\newblock \showarticletitle{Provable robust watermarking for ai-generated
  text}.
\newblock \bibinfo{journal}{\emph{arXiv preprint arXiv:2306.17439}}
  (\bibinfo{year}{2023}).
\newblock


\bibitem[Zhao et~al\mbox{.}(2025)]%
        {zhao2025sokwatermarkingaigeneratedcontent}
\bibfield{author}{\bibinfo{person}{Xuandong Zhao}, \bibinfo{person}{Sam Gunn},
  \bibinfo{person}{Miranda Christ}, \bibinfo{person}{Jaiden Fairoze},
  \bibinfo{person}{Andres Fabrega}, \bibinfo{person}{Nicholas Carlini},
  \bibinfo{person}{Sanjam Garg}, \bibinfo{person}{Sanghyun Hong},
  {et~al\mbox{.}}} \bibinfo{year}{2025}\natexlab{}.
\newblock \bibinfo{title}{SoK: Watermarking for AI-Generated Content}.
\newblock
\showeprint[arxiv]{2411.18479}~[cs.CR]


\bibitem[Zhong et~al\mbox{.}(2020)]%
        {zhong2020neural}
\bibfield{author}{\bibinfo{person}{Wanjun Zhong}, \bibinfo{person}{Duyu Tang},
  \bibinfo{person}{Zenan Xu}, \bibinfo{person}{Ruize Wang},
  \bibinfo{person}{Nan Duan}, \bibinfo{person}{Ming Zhou},
  \bibinfo{person}{Jiahai Wang}, {and} \bibinfo{person}{Jian Yin}.}
  \bibinfo{year}{2020}\natexlab{}.
\newblock \showarticletitle{Neural deepfake detection with factual structure of
  text}.
\newblock \bibinfo{journal}{\emph{arXiv preprint arXiv:2010.07475}}
  (\bibinfo{year}{2020}).
\newblock


\bibitem[Zhong et~al\mbox{.}(2023)]%
        {zhong2023mquake}
\bibfield{author}{\bibinfo{person}{Zexuan Zhong}, \bibinfo{person}{Zhengxuan
  Wu}, \bibinfo{person}{Christopher~D Manning}, \bibinfo{person}{Christopher
  Potts}, {and} \bibinfo{person}{Danqi Chen}.} \bibinfo{year}{2023}\natexlab{}.
\newblock \showarticletitle{Mquake: Assessing knowledge editing in language
  models via multi-hop questions}.
\newblock \bibinfo{journal}{\emph{arXiv preprint arXiv:2305.14795}}
  (\bibinfo{year}{2023}).
\newblock


\bibitem[Zhou et~al\mbox{.}(2025)]%
        {zhou2025dont}
\bibfield{author}{\bibinfo{person}{Yukai Zhou}, \bibinfo{person}{Jian Lou},
  {and} \bibinfo{person}{et al.}} \bibinfo{year}{2025}\natexlab{}.
\newblock \showarticletitle{Don’t Say No: Jailbreaking LLM by Suppressing
  Refusal}. In \bibinfo{booktitle}{\emph{Findings of the Association for
  Computational Linguistics: ACL 2025}}. \bibinfo{publisher}{Association for
  Computational Linguistics}, \bibinfo{pages}{25224--25249}.
\newblock


\bibitem[Zhu et~al\mbox{.}(2024)]%
        {zhu2024duwakdualwatermarkslarge}
\bibfield{author}{\bibinfo{person}{Chaoyi Zhu}, \bibinfo{person}{Jeroen
  Galjaard}, \bibinfo{person}{Pin-Yu Chen}, {and} \bibinfo{person}{Lydia~Y.
  Chen}.} \bibinfo{year}{2024}\natexlab{}.
\newblock \bibinfo{title}{Duwak: Dual Watermarks in Large Language Models}.
\newblock
\showeprint[arxiv]{2403.13000}~[cs.LG]
\urldef\tempurl%
\url{https://arxiv.org/abs/2403.13000}
\showURL{%
\tempurl}


\end{thebibliography}
